\documentclass[10pt,journal,compsoc]{IEEEtran}

\usepackage{amsmath}
\usepackage{amsthm,amssymb}
\usepackage{graphicx}
\usepackage{algorithm}
\usepackage{algorithmic}
\usepackage{multirow}
\usepackage{makecell}
\usepackage{threeparttable}
\usepackage{xcolor}
\usepackage{color}
\usepackage{colortbl}

\usepackage{caption}
\usepackage{booktabs}
\usepackage{subcaption}
\usepackage{longtable}
\usepackage{fontawesome}
\usepackage{utfsym}
\definecolor{forestgreen}{HTML}{228B22}
\definecolor{darkgreen}{HTML}{006400}
\definecolor{grey}{HTML}{808080}

\ifCLASSINFOpdf
\else
\fi

\begin{document}
%
\title{Are Large Language Models Really Good Logical Reasoners? A Comprehensive Evaluation and Beyond}
%
%
%
%

\author{Fangzhi Xu*, Qika Lin*,~\IEEEmembership{Member,~IEEE}, Jiawei Han, Tianzhe Zhao,\\Jun Liu,~\IEEEmembership{Senior Member,~IEEE}, Erik Cambria,~\IEEEmembership{Fellow,~IEEE}
\IEEEcompsocitemizethanks{\IEEEcompsocthanksitem Fangzhi Xu, Qika Lin, Jiawei Han and Tianzhe Zhao are with the School of Computer Science and Technology, Xi’an Jiaotong University, Xi’an, Shaanxi 710049, China (email: \{Leo981106, tara1208260223\}@stu.xjtu.edu.cn, \{qikalin, ztz8758\}@foxmail.com).
\IEEEcompsocthanksitem Jun Liu is with the Shaanxi Provincial Key Laboratory of Big Data Knowledge Engineering, and National Engineering Lab for Big Data Analytics, Xi’an, Shaanxi 710049, China (e-mail: liukeen@xjtu.edu.cn).
\IEEEcompsocthanksitem Erik Cambria is with the School of Computer Science and Engineering, Nanyang Technological University (e-mail: cambria@ntu.edu.sg).
\IEEEcompsocthanksitem Fangzhi Xu and Qika Lin contribute equally to this paper.
\IEEEcompsocthanksitem Corresponding author: Qika Lin.
}
}


\IEEEtitleabstractindextext{%
\begin{abstract}
Logical reasoning consistently plays a fundamental and significant role in the domains of knowledge engineering and artificial intelligence.
Recently, Large Language Models (LLMs) have emerged as a noteworthy innovation in natural language processing (NLP).
However, the question of whether LLMs can effectively address the task of logical reasoning, which requires gradual cognitive inference similar to human intelligence, remains unanswered.
To this end, we aim to bridge this gap and provide comprehensive evaluations in this paper. Firstly, to offer systematic evaluations, we select fifteen typical logical reasoning datasets and organize them into deductive, inductive, abductive and mixed-form reasoning settings. 
Considering the comprehensiveness of evaluations, 
we include 3 early-era representative LLMs and 4 trending LLMs.
Secondly, different from previous evaluations relying only on simple metrics (e.g., \emph{accuracy}), we propose fine-level evaluations in objective and subjective manners, covering both answers and explanations, including \emph{answer correctness}, \emph{explain correctness}, \emph{explain completeness} and \emph{explain redundancy}.
Additionally, to uncover the logical flaws of LLMs, problematic cases will be attributed to five error types from two dimensions, i.e., \emph{evidence selection process} and \emph{reasoning process}.
Thirdly, to avoid the influences of knowledge bias and concentrate purely on benchmarking the logical reasoning capability of LLMs, we propose a new dataset with neutral content.
Based on the in-depth evaluations, this paper finally forms a general evaluation scheme of logical reasoning capability from six dimensions (i.e., \emph{Correct}, \emph{Rigorous}, \emph{Self-aware}, \emph{Active}, \emph{Oriented} and \emph{No hallucination}). It reflects the pros and cons of LLMs and gives guiding directions for future works.
\end{abstract}

\begin{IEEEkeywords}
Logical reasoning, large language model, deductive reasoning, inductive reasoning, abductive reasoning.
\end{IEEEkeywords}}

\maketitle

\IEEEdisplaynontitleabstractindextext

\IEEEpeerreviewmaketitle

\section{Introduction}

\IEEEPARstart{A}{s} a fundamental and significant topic in the domains of knowledge engineering and artificial intelligence,
logical reasoning has consistently remained a subject of intense research interest~\cite{DBLP:journals/tkde/AntoniouB07,DBLP:journals/tkde/Lukasiewicz10,DBLP:journals/tkde/LinLZPHXZ23}.
Through the integration of logical reasoning, a wide range of intelligent applications can be developed (e.g., recommendation systems~\cite{DBLP:conf/icdm/WuZZ22}, relation prediction~\cite{DBLP:conf/sigir/LinLXPZZZ22} and question generation~\cite{DBLP:journals/tkde/YuSQY23}),
which not only offer powerful capabilities but also ensure natural interpretability.
However, developing efficient and robust logical reasoning systems~\cite{DBLP:conf/sigir/XuLLPZ22} 
remains a challenging academic pursuit due to complexities in handling intricate semantic and syntactic structures, managing symbolic knowledge, executing high-level abstractions and inferences, and navigating uncertainty and ambiguity~\cite{bronkhorst2020logical,DBLP:journals/corr/abs-2303-12023}.

With the aid of large-scale pre-training, instruction fine-tuning, and human feedback reinforcement learning strategies \cite{DBLP:journals/corr/abs-2304-03277}, Large Language Models (LLMs) have made unparalleled strides in the artificial intelligence community, particularly in the field of natural language processing (NLP).
They have achieved remarkable performance in numerous traditional NLP tasks~\cite{DBLP:journals/corr/abs-2303-18223,DBLP:journals/expert/AminCS23}, such as question answering, information retrieval, and affective computing.
Under this circumstance, researchers have begun to question the efficacy of LLMs in addressing complex logical reasoning tasks~\cite{DBLP:conf/acl/JiaoGSN22, xu-etal-2024-pathreasoner}and their ability to apply this knowledge to downstream intelligent applications.
As such, a natural question arises: \emph{are large language models really good logical reasoners}?

\begin{figure}[t]
	\centering
		\includegraphics[scale=0.65]{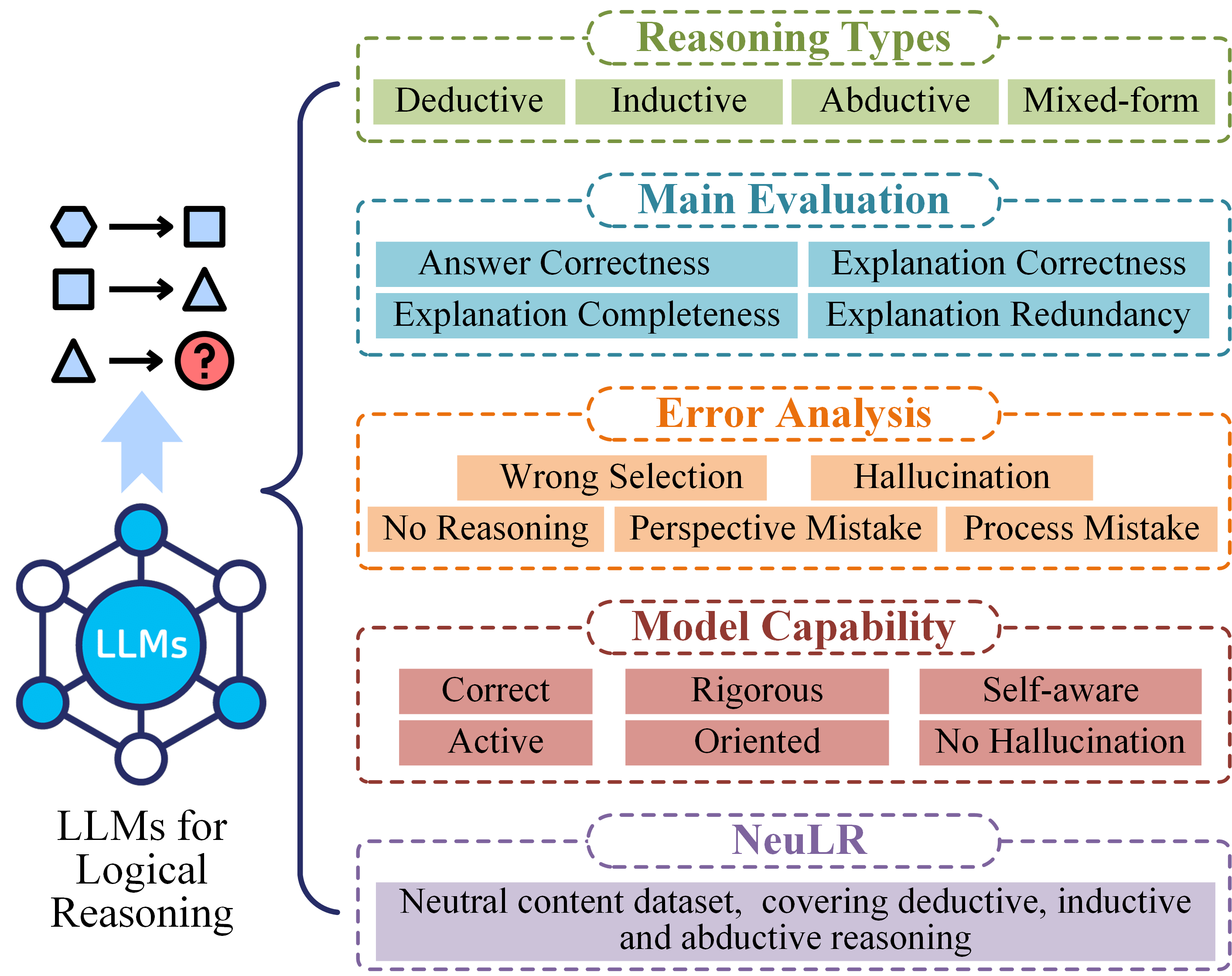}
		\caption{The overall architecture of the evaluation.}
		\label{figintro}
\end{figure}

There are already several studies to evaluate the capability of LLMs from various reasoning perspectives,
e.g., multilingual reasoning~\cite{DBLP:journals/corr/abs-2302-04023}, commonsense reasoning~\cite{DBLP:journals/corr/abs-2303-16421}, and mathematical reasoning~\cite{DBLP:journals/corr/abs-2303-05398}. 
These efforts are meaningful and inspiring.
Nevertheless, all of them are confronted with one or more of the following defects:
(1) There are no comprehensive evaluations, which are hampered by the issue of lacking systematic category, limited LLMs for comparison, and limited data samples for evaluation; (2) The majority of evaluation works purely report the accuracy of answers, which limits fine-level analysis on the reasoning process and fails to explore the causes of mistakes; (3) Current logical reasoning benchmarks may fail to purely evaluate logical reasoning ability since the reasoning of LLMs can be affected by the content; (4) There lack a complete evaluation system or well-defined dimensions to comprehensively conclude the logical reasoning capability of LLMs. Therefore, our work aims to fill these gaps and provide a comprehensive evaluation, where the overall architecture encompasses the five main points as illustrated in Fig.~\ref{figintro}: category with reasoning settings, main metrics for evaluation, perspectives of error analysis, dimensions for measuring logical reasoning capability, and new benchmark NeuLR with neutral content. In detail, we address the above limitations from the following aspects.

Firstly,
our work starts from systematic views and provides comprehensive evaluations. According to the classical definition~\cite{flach2000abductive}, logical reasoning can be mainly categorized into three fundamental types, i.e., deductive, inductive and abductive reasoning. They together form a complete chain of reasoning, thus it is meaningful to evaluate LLMs from these views. 
Based on it, our comprehensive evaluation is reflected in three orthogonal dimensions. For the reasoning type view, all the evaluated datasets are categorized into four reasoning settings, i.e., deductive, inductive, abductive and mixed-forms. The former three involve the independent reasoning manner.
For the dataset view, we include fifteen typical logical reasoning datasets according to the above categories. 
{\color{black}For the model view, we evaluate three previous LLMs (i.e., text-davinci-003~\cite{DBLP:conf/nips/Ouyang0JAWMZASR22}, ChatGPT~\cite{ouyang2022training} and BARD~\cite{anil2023palm}) as well as four up-to-date LLMs (i.e., LLaMA3.1-Chat~\footnote{https://llama.meta.com/}, Mistral-Instruct-v0.3~\footnote{https://huggingface.co/mistralai/Mistral-7B-Instruct-v0.3}, Claude-3~\footnote{https://claude.ai/} and GPT-4~\footnote{https://openai.com/index/gpt-4/}).}

Secondly, our work fills the blank in fine-level evaluations of logical reasoning tasks. Current benchmarks only rely on a few objective metrics (e.g., accuracy) to measure the model capability. It may not be sufficient in the case of generative LLMs, since the role of LLMs is not only limited to correctly answer questions but also serves as practical tools, which are required to provide reasoning chains or explanations. Previous works~\cite{DBLP:journals/corr/abs-2302-06476, DBLP:journals/corr/abs-2304-14106} conduct extensive experiments on popular NLP datasets, but they purely report the performance results. Since some LLMs (e.g., ChatGPT) function as the interactive tools for human use, it is necessary to introduce subjective metrics to do fine-grained evaluations. In this paper, we employ four dimensions of metrics, covering \emph{answer correctness}, \emph{explanation correctness}, \emph{explanation completeness} and \emph{explanation redundancy}. It can provide more meaningful and complete evaluations from both objective and subjective views. Considering problematic cases (i.e., wrong answer or wrong explanation) can reflect obvious logical flaws of LLMs, we further attribute them to several error types from two dimensions of \emph{evidence selection process} and \emph{reasoning process} and give in-depth analysis. 
{\color{black} Notably, the whole evaluation system can be applied under both human-annotation and automatic modes.}

Thirdly, our work focuses on the issue of content neutrality and provides new solutions. The current benchmark for evaluating logical reasoning ability is strongly coupled with text comprehension; in other words, the rule reasoning process of LLMs may be affected by the content in the inputs, which limits the test of real logical reasoning ability. Also, LLMs are highly powerful due to their massive training data, which may overlap with popular benchmarks. As a result, testing LLMs on these benchmarks may not be entirely fair, as it can only demonstrate the fitting ability of LLMs rather than their real logical reasoning capability. Therefore, language models may be trained to learn a biased pattern from text, rather than really capture the logical reasoning capability. Some previous works~\cite{DBLP:journals/corr/abs-2305-08322, DBLP:journals/corr/abs-2301-13848} propose to establish complete benchmarks for LLMs. But few works focus on logical reasoning and fail to attend to the content-neutral problem. To narrow this gap, we propose a new dataset named NeuLR, which contains 3,000 content-neutral samples and covers the deductive, inductive and abductive reasoning types. It is expected to offer a novel perspective for benchmarking the logical reasoning ability of LLMs.

Finally, we conclude the extensive performance results of LLMs and form an evaluation scheme with six key properties, i.e., \emph{Correct}, \emph{Rigorous}, \emph{Self-aware}, \emph{Active}, \emph{Oriented} and \emph{No Hallucination}. Among them, \emph{Correct} purely measures the accuracy of the answer. \emph{Rigorous} measures whether LLMs give both correct answers and complete and correct explanations. \emph{Self-aware} is reflected by the redundancy of the generated content. \emph{Active} is measured by the proportion of reasoning. \emph{Oriented} illustrates whether LLMs can reason from the right perspectives. \emph{No hallucination} measures whether LLMs are more prone to produce hallucinations. 
The above dimensions can all be quantified from existing evaluation experiments. 
For deductive, inductive, abductive, and mixed reasoning settings respectively, we visualize the ability maps for each LLM. It is meaningful to identify the strengths and weaknesses of LLMs under the four reasoning settings, thus guiding future directions. 

The main contributions of the paper are listed as follows:

(1) In view of the great success of LLMs in massive NLP tasks, our work is targeted at answering \emph{are LLMs really good logical reasoners?}. In this paper, we provide a comprehensive evaluation and give potential directions for future researches.

(2) For a comprehensive evaluation of logical reasoning, this paper classifies datasets into four reasoning manners, i.e., deductive, inductive, abductive, and mixed-form. 
{\color{black} We include fifteen typical logical reasoning datasets and evaluate seven representative LLMs under diverse prompting strategies, including 3 early-era LLMs (text-davinci-003, ChatGPT, and BARD) and 4 trending LLMs (LLaMA3.1-Chat, Mistral-Instruct-v0.3, Claude-3 and GPT-4).
}

(3) Considering the drawbacks in current objective metrics, this paper gives fine-level evaluations including four dimensions i.e., answer correctness, explanation correctness, explanation completeness, and explanation redundancy. To explore the value of failure cases, we attribute them to several error types and explore the logical flaws of LLMs.

(4) To provide fair evaluations with neutral content and decouple logical reasoning from text understanding, this paper proposes a new dataset named NeuLR\footnote{Available at https://github.com/DeepReasoning/NeuLR}. It contains 3,000 content-neutral samples and covers deductive, inductive and abductive reasoning manners. 

(5) In view of the evaluation results,
this paper forms a general evaluation scheme for the logical reasoning capability of LLMs for the first time, which concludes six key properties, i.e, \emph{Correct}, \emph{Rigorous}, \emph{Self-aware}, \emph{Active}, \emph{Oriented} and \emph{No hallucination}.
Furthermore, we derive the ability maps for each LLM under four reasoning settings respectively and propose future directions.

\section{Preliminary}

Logical reasoning aims to generate logical implications that contain new facts using one- or multi-step inference based on given premises~\cite{DBLP:conf/emnlp/PanLZZLHW22,DBLP:conf/kdd/HuangCL22,DBLP:conf/kdd/ChengL0S22}, i.e.,
\emph{premise}$\Rightarrow$\emph{conclusion}.
Elements of logical reasoning typically include knowledge facts or premises and logical rules, for example:
\begin{itemize}
\setlength{\itemindent}{-1.2em}
\item rule: All eight-year-old children are in primary school.
\item fact1/premise1: Jordan is a child of eight years old.
\item fact2/premise2: Jordan is in primary school.
\end{itemize}

According to the reasoning classification system of classical logic~\cite{flach2000abductive}, there are three major types of logical reasoning:
deductive, inductive and abductive.
Drawing upon the aforementioned rule and facts, we can depict the objective of these three types of reasoning as forecasting the third item based on the two provided ones.

\begin{figure}[t]
	\centering
		\includegraphics[scale=0.5]{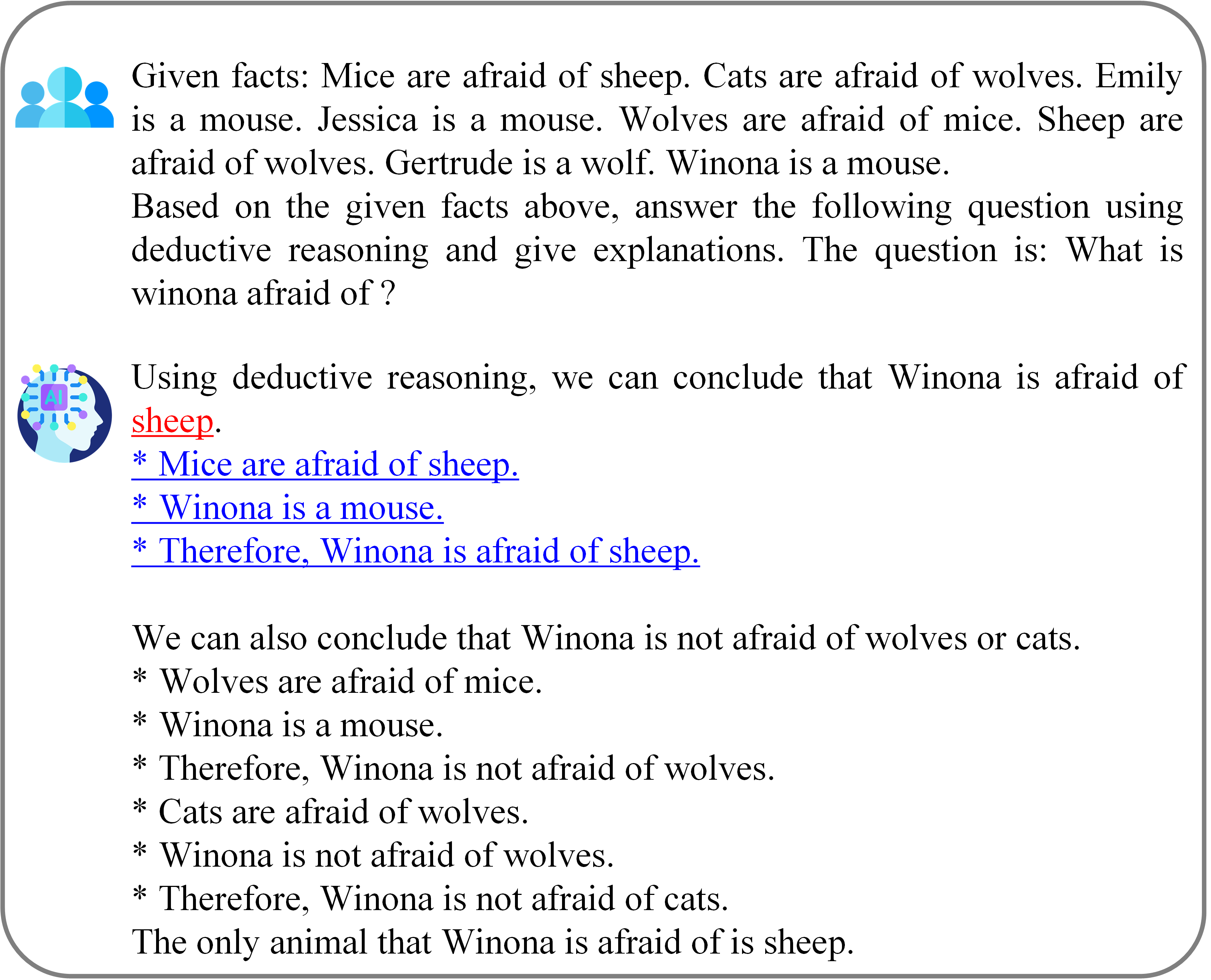}
		\caption{An example of using an LLM to answer reasoning questions. The \textcolor{red}{\underline{red}} words represent the generated results, while the \textcolor{blue}{\underline{blue}} ones represent the generated explanations.}
		\label{figintroexample}
\end{figure}

\textbf{Deductive Reasoning}. Deductive reasoning is the psychological process of drawing deductive inferences that start from the given premises
and reason with logical rules or commonsense to obtain \emph{certain} conclusions~\cite{johnson1999deductive,goel2007anatomy}.
It can be \emph{premise1}+\emph{rule}$\Rightarrow$\emph{premise2}. Fig.~\ref{figintroexample} presents an example of deductive reasoning.
Its reasoning progress generates specific knowledge facts from general counterparts,
e.g., \emph{premise2} and \emph{rule} are specific and general knowledge, respectively.
Therefore, deductive reasoning is actually a top-down way.

\textbf{Inductive Reasoning}. Distinct from deductive reasoning,
inductive reasoning derives general principles from a body of observations
which means making broad generalizations based on specific observations~\cite{heit2010relations,yu2023nature}.
For example, an example of inductive reasoning can be \emph{premise1}+\emph{premise2}$\Rightarrow$\emph{rule},
concluding generalized knowledge \emph{rule} that is independent with specific item \emph{Jordan}.
Generally, the truth of the conclusion of an inductive argument is \emph{probable} rather than \emph{certain} in inductive reasoning.
Thus, inductive reasoning is a bottom-up approach and is contrasted with deductive reasoning.

\textbf{Abductive Reasoning}. Formally, abductive reasoning is similar to deductive reasoning which seeks conclusions from a set of observations.
But differently, its target is to generate the simplest and most likely explanation for the given observations~\cite{josephson1996abductive,walton2001abductive}.
So the result is \emph{probable} like in inductive reasoning.
An example of abductive reasoning can be \emph{premise2}+\emph{rule}$\Rightarrow$\emph{premise1},
which means \emph{premise1} is the most likely cause of \emph{premise2}.



{\color{black}
\textbf{Mixed-form Reasoning.} Numerous real-life reasoning scenarios may require multiple steps and the incorporation of at least two of these three types.
In this paper, they are vieweds as a more intricate type of reasoning, namely, \emph{mixed-form}.
For example, given the following fact:
1) James is a crow and is black; 
2) Sophia is a crow and is black; 
3) Emily is a crow.
To answer the question \emph{what color is Emily? }, we should first get the conclusion that \emph{all crows are black} using inductive reasoning and then conduct deductive
reasoning on this conclusion and fact 3) to obtain the final answer \emph{black}.
}

\vspace{-0.5cm}
\section{Evaluation Details}
This section presents the comprehensive experimental settings, including the models, datasets and metrics.
More detailed prompts for zero- or few-shot are shown in Table 3 of the Appendix~\footnote{Appendix is at: https://github.com/DeepReasoning/NeuLR}.

\subsection{Evaluated Models}
Because of the rapid emergence of LLMs, it is not realistic to include all LLMs in this paper. Thus, we mainly select three early-era representative LLMs (i.e., text-davinci-003,
ChatGPT,
and BARD)
{\color{black}and four up-to-date LLMs (i.e., LLaMA3.1-Chat, Mistral-Instruct-v0.3, Claude-3 and GPT-4).}
The details of these models are listed in the Appendix.


\subsection{Evaluated Datasets}
According to the previous discussion, the evaluation is conducted systematically from deductive, inductive, abductive and mixed-form views. Therefore, this paper selects fifteen popular datasets in logical reasoning and divides them into the above four folds. Table 2 in the Appendix presents detailed information of these datasets.
The selected datasets contain both generation and classification ones and there exist diverse forms of tasks, which illustrate the comprehensiveness of our evaluation.
Diverging from prior works that only employ a limited number of samples, this paper significantly expands the evaluation size.
Since ChatGPT is one of the most popular LLM for the public, we give much focus on it.
For parts of the datasets, we keep all the test examples for ChatGPT evaluation (i.e., EntailmentBank~\cite{DBLP:conf/emnlp/DalviJTXSPC21}, FOLIO~\cite{DBLP:journals/corr/abs-2209-00840}, Leap-Of-Thought~\cite{DBLP:conf/nips/TalmorTCGB20}, CLUTRR~\cite{DBLP:conf/emnlp/SinhaSDPH19}, ReClor~\cite{DBLP:conf/iclr/YuJDF20}, LogiQA~\cite{DBLP:conf/ijcai/LiuCLHWZ20}, LogiQA 2.0~\cite{DBLP:journals/corr/abs-2304-03439} and LogiQA2NLI~\cite{DBLP:journals/corr/abs-2304-03439}), while other large datasets (i.e., bAbI-15~\cite{DBLP:journals/corr/WestonBCM15}, RuleTaker~\cite{DBLP:conf/ijcai/ClarkTR20}, bAbI-16~\cite{DBLP:journals/corr/WestonBCM15}, $\alpha$-NLI~\cite{DBLP:conf/iclr/BhagavatulaBMSH20}, $\alpha$-NLG~\cite{DBLP:conf/iclr/BhagavatulaBMSH20}, AbductiveRules~\cite{DBLP:conf/acl/Young0BW22} and D*-Ab~\cite{DBLP:conf/acl/TafjordDC21}) are sampled to 1,000 examples.
As for the evaluation of text-davinci-003 and BARD,
we sample 100 test examples for each dataset.

\subsection{Selected Metrics}
The majority of prior evaluation studies solely report the accuracy metric, which may not be comprehensive enough for evaluating the effectiveness of LLMs. Consequently, we assert that a more nuanced assessment is necessary. To this end, we propose to evaluate LLMs from both objective and subjective perspectives.
Specifically, we introduce four evaluation metrics to reflect the intermediate reasoning process of LLMs:
\emph{answer correctness}, \emph{explanation correctness}, \emph{explanation completeness} and \emph{explanation redundancy}. 

\begin{itemize}
\item \textbf{Answer Correctness}.
{\color{black}
This metric measures the alignment between a generated answer and the true label, emphasizing semantic coherence over exact token matching. For example, "the car hit the tree" and "the tree was struck by the car" demonstrate this alignment despite differing expression style.
}
\item \textbf{Explanation Correctness}.
{\color{black} To reason towards the answer, explanations usually provide a logical and
coherent narrative that connects the available evidence or premises to the ultimate conclusion.
Thus this metric indicates whether the generated explanation is logically correct for the true answer.}

\item \textbf{Explanation Completeness}. 
{\color{black}
It indicates whether there are any missing explanations to reason towards the true answer.
An explanation is deemed complete if it encompasses all elements present in the true explanation, which means that in the reasoning process,
the correct answer can be inferred through the selected known facts and the generated intermediate facts.
But this does not necessarily cause answer correctness or explanation correctness.
}
\item \textbf{Explanation Redundancy}.
{\color{black}
This suggests that the combination of selected established facts and additional intermediate facts surpasses the necessary information for arriving at a precise conclusion, leading to the inclusion of unnecessary details in the reasoning process. For instance, only three specific facts are essential for drawing a conclusion; any extra facts beyond this set would be redundant.}
\end{itemize}

Notably, the metric values of explanation correctness, completeness, and redundancy are independent.
To identify prevalent logical flaws in LLMs, we establish error types to categorize problematic cases. This paper classifies errors along two primary dimensions:
(1) \emph{evidence selection process} and (2) \emph{reasoning process}.
The first dimension centers on evaluating the evidence selected by LLMs, whereas the second dimension emphasizes the logical reasoning process using the selected evidence.
Detailedly, \emph{evidence selection process} category can be further divided into \emph{wrong selection} and \emph{hallucination}. The former denotes that LLMs select the wrong facts or ignore the necessary facts from the beginning of the reasoning. The latter denotes that LLMs select the evidence which contradicts the given context or can not be verified by the context.
\emph{Reasoning process} category can be further divided into \emph{no reasoning},
\emph{perspective mistake} and \emph{process mistake}.
The first error type signifies instances where LLMs fail to conduct reasoning, instead merely listing the given facts and the final answer.
The second denotes LLMs starting from an irrelevant point or focusing on an improper perspective for the correct answer.
The last refers to LLMs commencing from a proper viewpoint, but making mistakes during the reasoning process.

\section{Overall Experiments}


In this section, we conduct evaluation experiments 
{\color{black} on three early-era LLMs (i.e., text-davinci-003, ChatGPT and BARD) and four trending LLMs (i.e., LLaMA3.1-Chat, Mistral-Instruct-v0.3, Claude-3 and GPT-4) under various prompting methods.}
Table~\ref{main} presents the overall answer correctness of these three early-stage LLMs on fifteen logical reasoning datasets.
{\color{black} Table~\ref{main_new} supplements the performances of four popular LLMs.}
Generally, LLMs' performances on logical reasoning tasks still have significant room for improvement in comparison to the state-of-the-art (SOTA) metric. Most of the results fall short of those achieved by smaller-sized SOTA models. We provide a detailed analysis of the results from the following perspectives. 
{\color{black} For simplicity, we only include text-davinci-003, ChatGPT and BARD for discussion.}

\begin{table*}[t]
	\centering
	\small
	\caption{Overall results of LLMs' answer correctness across the zero-shot, one-shot and three-shot logical reasoning settings. The notations \emph{De.}, \emph{In.}, \emph{Ab.} and \emph{Mix} correspond to deductive, inductive, abductive and mixed-form reasoning, respectively (as in the following tables and figures). \emph{Gen.} indicates whether the task is a generation one. The percentage signs (\%) of performance values are omitted for simplicity in the paper.}
	\begin{tabular}{c|c|c|ccc|ccc|ccc|c}
		\toprule
            & \multicolumn{1}{c|}{\multirow{2}{*}{\textbf{Dataset}}} & \multicolumn{1}{c|}{\multirow{2}{*}{\textbf{Gen.}}} & \multicolumn{3}{c|}{\textbf{text-davinci-003}} & \multicolumn{3}{c|}{\textbf{ChatGPT}} & \multicolumn{3}{c|}{\textbf{BARD}} & \multicolumn{1}{c}{\multirow{2}{*}{\textbf{SOTA}}}\\
            & \multicolumn{1}{c|}{} & \multicolumn{1}{c|}{} & \textbf{0-shot} & \textbf{1-shot} & \textbf{3-shot} & \textbf{0-shot} & \textbf{1-shot} & \textbf{3-shot} & \textbf{0-shot} & \textbf{1-shot} & \textbf{3-shot} \\
            \midrule 
            \multicolumn{1}{c|}{\multirow{5}{*}{\rotatebox{90}{\textbf{De.}}}} &\textbf{bAbI-15} &\checkmark &85.00 &76.00 &75.00 &38.40 &46.40 &39.70 &79.00 &80.00 &88.00 &100~\cite{DBLP:journals/corr/WestonBCM15}\\
            & \textbf{EntailmentBank} &\checkmark &93.00 &88.00 &89.00 &83.82 &82.06 &77.94 &96.00 &97.00 &97.00 &100~\cite{DBLP:conf/emnlp/DalviJTXSPC21}\\
            & \textbf{RuleTaker} & &64.00 &60.00 &62.00 &42.00 &38.00 &40.20 &64.00 &57.00 &70.00 &$\approx$100~\cite{DBLP:conf/ijcai/ClarkTR20}\\
            & \textbf{FOLIO} & &48.00 &53.00 &52.00 &50.00 &50.98 &54.41 &52.00 &43.00 &49.00 &62.11~\cite{DBLP:journals/corr/abs-2209-00840}\\
            & \textbf{Leap-Of-Thought} & &82.00 &90.00 &87.00 &72.61 &74.01 &61.21 &79.00 &72.00 &79.00 &99.7~\cite{DBLP:conf/nips/TalmorTCGB20}\\
            \midrule 
            \multicolumn{1}{c|}{\multirow{2}{*}{\rotatebox{90}{\textbf{In.}}}} & \textbf{bAbI-16} &\checkmark &84.00 &81.00 &74.00 &17.10 &24.70 &12.90 &73.00 &44.00 &52.00 &100~\cite{DBLP:journals/corr/WestonBCM15} \\
            & \textbf{CLUTRR} &\checkmark &6.00 &23.00 &20.00 &21.99 &19.55 &12.83 &23.00 &26.00 &24.00 &95.0~\cite{DBLP:conf/icml/Minervini0SGR20} \\
            \midrule 
            \multicolumn{1}{c|}{\multirow{4}{*}{\rotatebox{90}{\textbf{Ab.}}}} & \textbf{$\alpha$-NLI} & & 74.00 &70.00 &74.00 &80.90 &80.00 &79.10 &75.00 &74.00 &77.00 &68.90~\cite{DBLP:conf/iclr/BhagavatulaBMSH20} \\
            & \textbf{$\alpha$-NLG} & \checkmark & 9.00 &10.00 &12.00 &21.90 &23.40 &25.90 &10.00 &12.00 &15.00 &45.00~\cite{DBLP:conf/iclr/BhagavatulaBMSH20} \\
            & \textbf{AbductiveRules} & \checkmark &75.00 &42.00 &35.00 &23.30 &35.10 &29.80 &71.00 &49.00 &22.00 &100~\cite{DBLP:conf/acl/Young0BW22}\\
            & \textbf{D*-Ab} & \checkmark & 8.00 &21.00 &23.00 &11.60 &2.50 &1.80 &11.00 &0.00 &0.00 &$\geq$95~\cite{DBLP:conf/acl/TafjordDC21} \\
		\midrule 
            \multicolumn{1}{c|}{\multirow{4}{*}{\rotatebox{90}{\textbf{Mix}}}} & \textbf{ReClor} & &53.00 &53.00 &55.00 &58.80 &56.00 &58.80 &56.00 &55.00 &56.00 &75.00~\cite{DBLP:conf/acl/JiaoGSN22} \\
            & \textbf{LogiQA} & &41.00 &35.00 &39.00 &40.25 &39.48 &40.86 &48.00 &46.00 &47.00 &46.10~\cite{DBLP:conf/acl/JiaoGSN22} \\
            & \textbf{LogiQA 2.0} & &43.00 &42.00 &41.00 &54.60 &50.80 &54.80 &53.00 &46.00 &47.00 &72.25~\cite{DBLP:journals/corr/abs-2304-03439} \\
            & \textbf{LogiQA2NLI} & &59.00 &55.00 &58.00 &57.83 &53.83 &57.00 &48.00 &50.00 &47.00 &$\approx$80~\cite{DBLP:journals/corr/abs-2304-03439} \\
		\bottomrule
	\end{tabular}
	\label{main}
\end{table*}

\begin{table*}[t]
	\centering
	\small
	\caption{{\color{black}Supplementary results on the four trending LLMs across diverse prompt selections. In the implementation, \emph{Direct} utilizes the zero-shot prompting to output both explanations and answers, but not prompted with chain-of-thought. The values in the table represent the answer correctness.}}
	\begin{tabular}{c|c|c|cc|cc|cc|cc}
		\toprule
            & \multicolumn{1}{c|}{\multirow{2}{*}{\textbf{Dataset}}} & \multicolumn{1}{c|}{\multirow{2}{*}{\textbf{Gen.}}} & \multicolumn{2}{c|}{\textbf{LLaMA3.1-Chat}} & \multicolumn{2}{c|}{\textbf{Mistral-Ins-v0.3}} & \multicolumn{2}{c|}{\textbf{Claude-3.5}} & \multicolumn{2}{c}{\textbf{GPT-4}} \\
            & \multicolumn{1}{c|}{} & \multicolumn{1}{c|}{} & \textbf{Direct} & \textbf{COT} & \textbf{Direct} & \textbf{COT} & \textbf{Direct} & \textbf{COT} & \textbf{Direct} & \textbf{COT} \\
            \midrule
            \multicolumn{1}{c|}{\multirow{5}{*}{\rotatebox{90}{\textbf{De.}}}} &\textbf{bAbI-15} &\checkmark &89.00 &90.00 &59.00 &51.00 &97.00&100.00&98.00&98.00 \\
            & \textbf{EntailmentBank} &\checkmark &72.00 &74.00 &82.00 &86.00 &85.00 &73.00 &84.00 &77.00 \\
            & \textbf{RuleTaker} & &62.00 &64.00 &45.00 &64.00&61.00&64.00&59.00&67.00 \\
            & \textbf{FOLIO} & &53.00 &56.00 &48.00 &45.00&81.00&84.00&70.00&69.00 \\
            & \textbf{Leap-Of-Thought} & &81.00 &80.00 &78.00 &73.00&55.00&53.00&75.00&80.00 \\
            \midrule
            \multicolumn{1}{c|}{\multirow{2}{*}{\rotatebox{90}{\textbf{In.}}}} & \textbf{bAbI-16} &\checkmark &92.00 &88.00 &36.00 &23.00&80.00&86.00&91.00&92.00\\
            & \textbf{CLUTRR} &\checkmark &39.00 &40.00 &15.00 &20.00 &29.00 &21.00 &33.00 &26.00 \\
            \midrule
            \multicolumn{1}{c|}{\multirow{4}{*}{\rotatebox{90}{\textbf{Ab.}}}} & \textbf{$\alpha$-NLI} & &39.00 &37.00 &31.00 &28.00&36.00&38.00&37.00&38.00 \\
            & \textbf{$\alpha$-NLG} & \checkmark &17.00 &17.00 &16.00 &15.00 &26.00 &14.00 &29.00 &23.00 \\
            & \textbf{AbductiveRules} & \checkmark &42.00 &22.00 &35.00 &25.00 &49.00 &41.00 &40.00 &34.18 \\
            & \textbf{D*-Ab} & \checkmark &10.00 &24.00 &6.00 &3.00 &38.00 &27.00 &35.00 &33.00 \\
		\midrule
            \multicolumn{1}{c|}{\multirow{4}{*}{\rotatebox{90}{\textbf{Mix}}}} & \textbf{ReClor} & &67.00 &63.00 &55.00 &52.00&88.00&90.00&88.00&85.00 \\
            & \textbf{LogiQA} & &46.54 &44.55 &50.23 &42.70&64.00&68.00&64.00&65.00 \\
            & \textbf{LogiQA 2.0} & &63.00 &55.00 &45.00 &47.00&74.00&79.00&81.00&82.00 \\
            & \textbf{LogiQA2NLI} & &61.00 &51.00 &54.00 &54.00 &60.00&61.00&59.00&48.00\\
		\bottomrule
	\end{tabular}
	\label{main_new}
\end{table*}

\begin{figure}[t]
	\centering
		\includegraphics[scale=1.0]{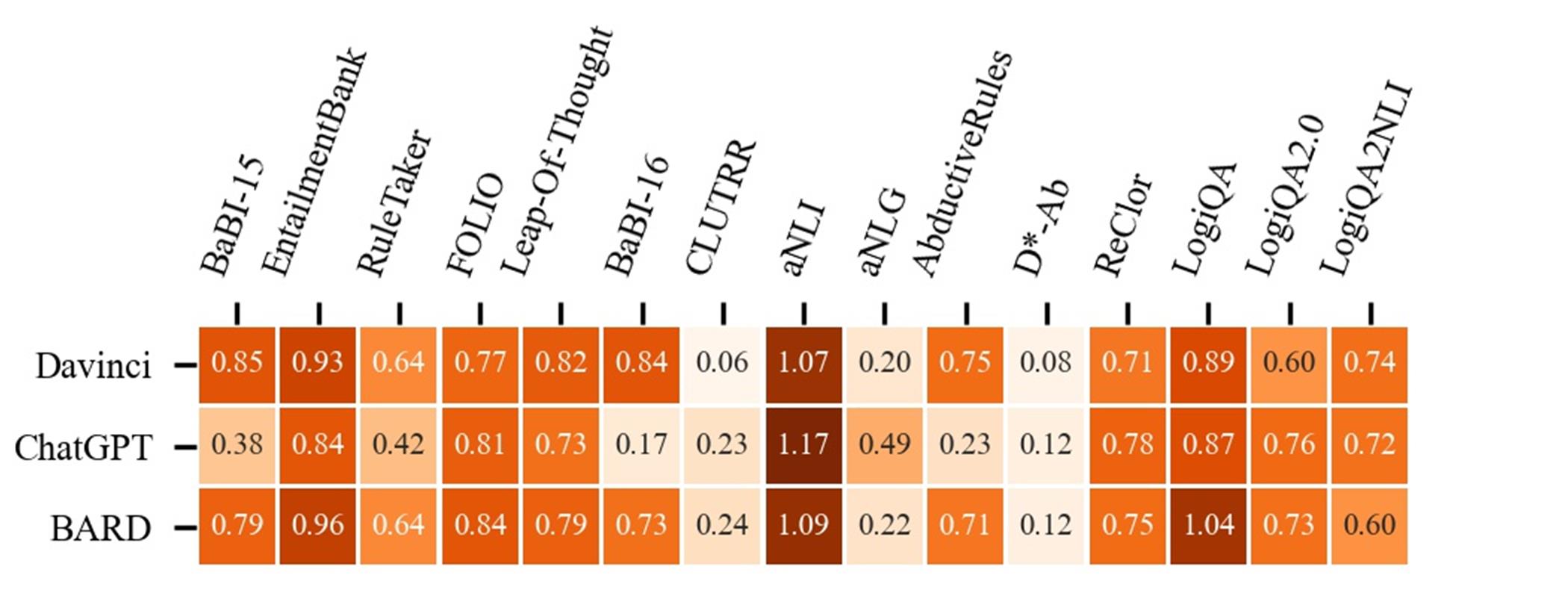}
		\caption{LLM performances on different datasets.}
		\label{heatmap_acc}
\end{figure}

\begin{figure}[t]
    \begin{minipage}[t]{0.49\linewidth}
        \large
        \centering
        \includegraphics[scale=0.30]{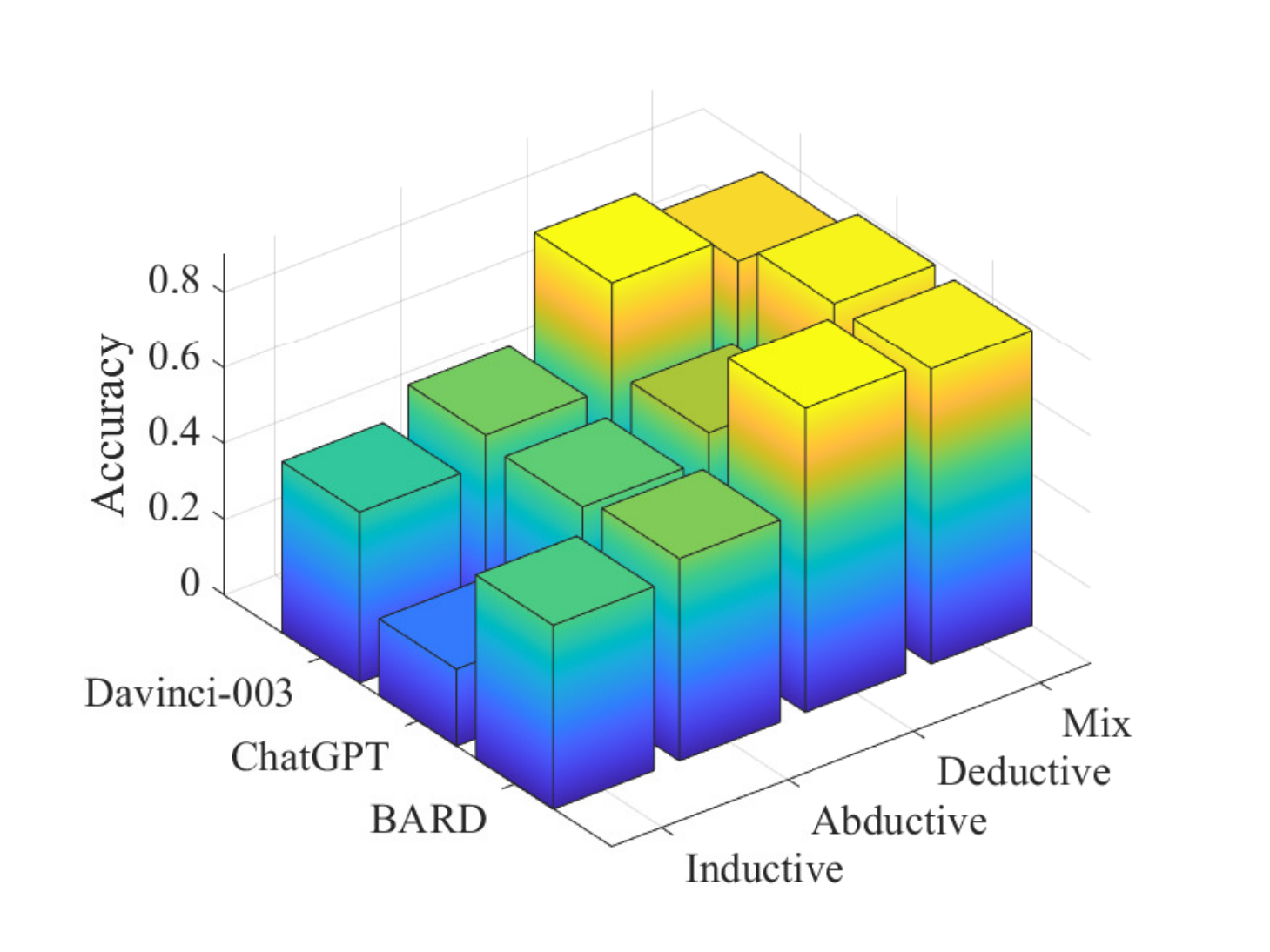}
        \vspace{-0.3cm}
        \subcaption{Different reasoning types.}
        \label{bar_correct}
    \end{minipage}
    \begin{minipage}[t]{0.49\linewidth}
        \large
        \centering
        \includegraphics[scale=0.30]{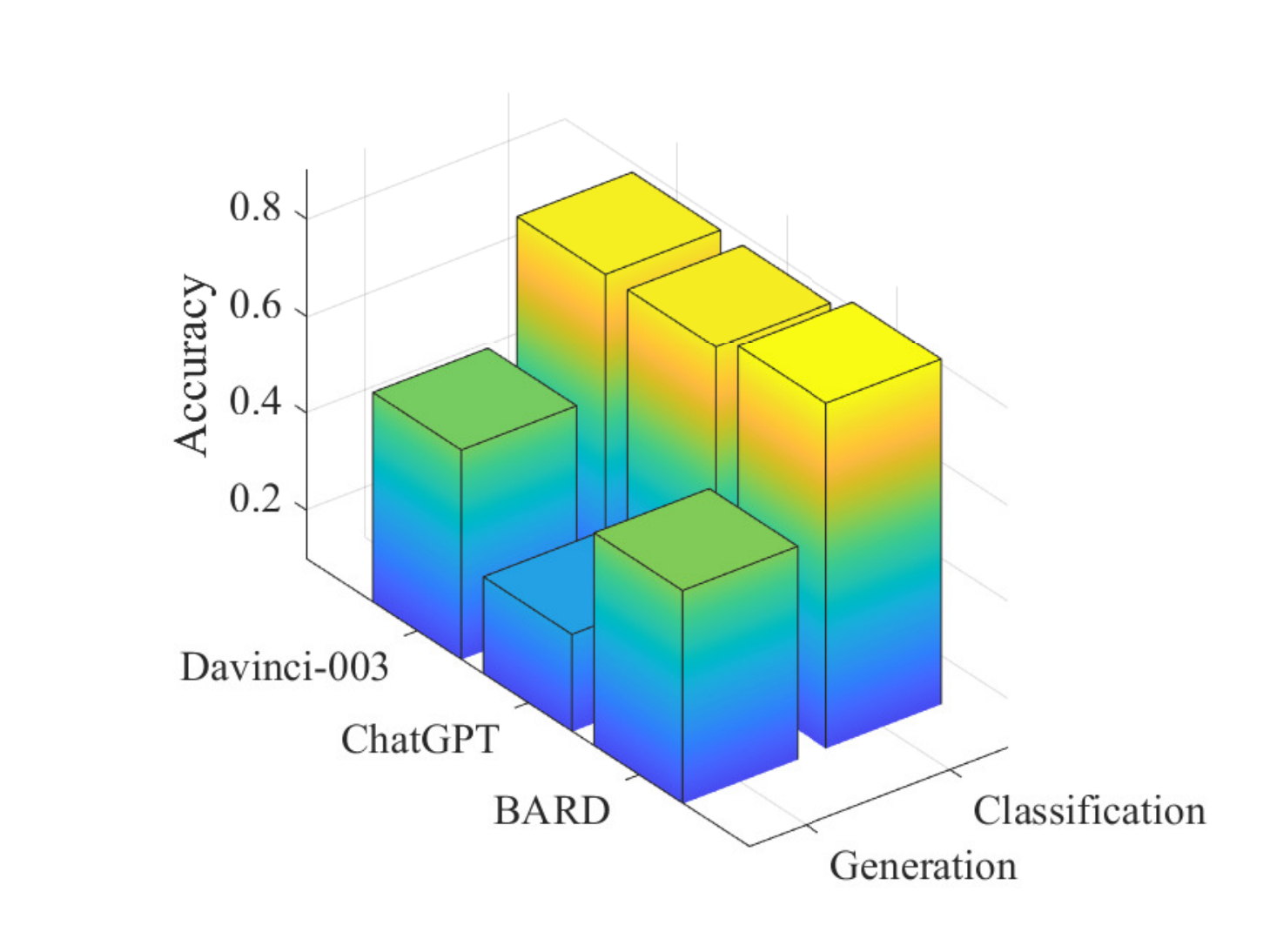}
        \vspace{-0.3cm}
        \subcaption{Generation/Classification.}
        \label{bar_type}
    \end{minipage}
    \caption{Visualization on the metric of answer correctness.}
\end{figure}

Firstly, we conduct an analysis of LLMs' performances across four reasoning manners, solely focusing on the zero-shot results to facilitate a clear comparison.
Furthermore, we introduce the relative performance metric (i.e., LLM accuracy/SOTA) to reflect the relative capability of LLMs in comparison to SOTA performances in Fig.~\ref{heatmap_acc}.
We also calculate the weighted results of four reasoning manners in Fig.~\ref{bar_correct}.
From the results, ChatGPT performs worse in deductive and inductive settings compared with text-davinci-003 and BARD. In the abductive setting, three LLMs show comparable performances and BARD wins with slight advantages. In the mixed-form setting, ChatGPT performs better and BARD ranks second. 
\textbf{Overall, BARD shows consistent
superiority among deductive, inductive and abductive settings, while text-davinci-003 also does relatively well.} 
It seems that ChatGPT struggles in the three settings, but is better at mixed-form reasoning.
Also, we compare the LLM performances between deductive, inductive and abductive settings. 
\textbf{LLMs do best in deductive setting, while they mostly struggle in inductive setting.}
We argue that deductive and abductive reasoning align with typical NLP scenarios, where LLMs have to provide missing facts. Conversely, inductive reasoning necessitates extracting high-level rules or knowledge from the given facts, which is more intricate and may not be readily available in the training corpus.

\begin{figure}[t]
    \begin{minipage}[t]{0.49\linewidth}
        \large
        \centering
        \includegraphics[scale=0.48]{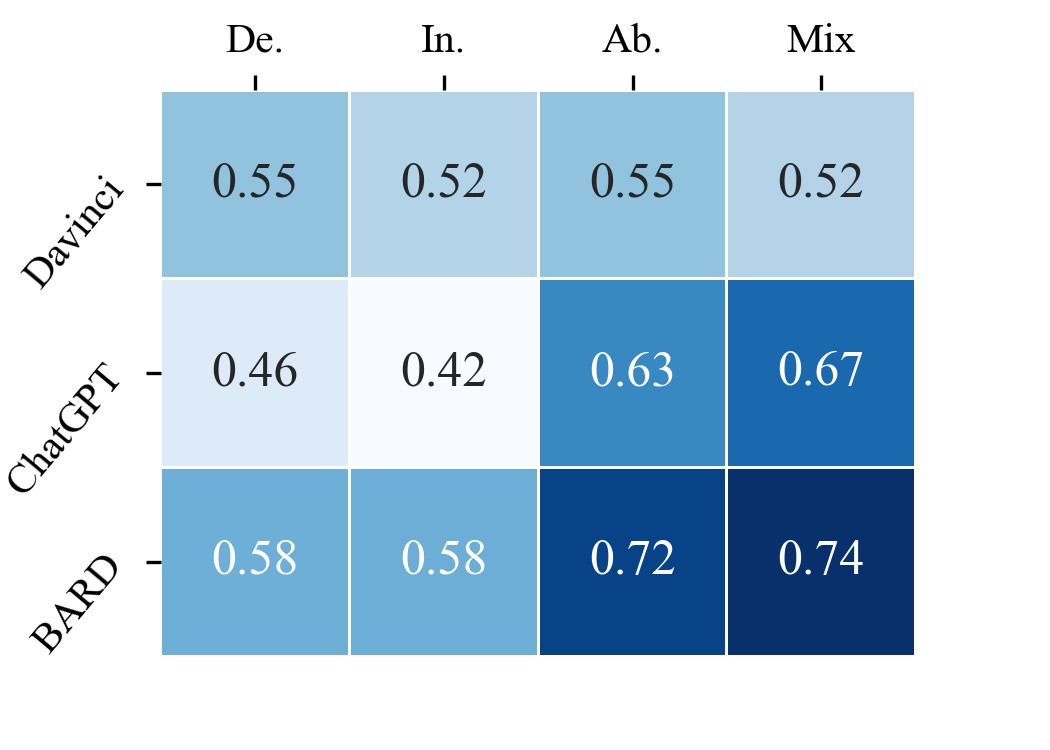}
        \vspace{-0.3cm}
        \subcaption{Rigorous evaluation.}
        \label{heatmap_rigorous}
    \end{minipage}
    \begin{minipage}[t]{0.49\linewidth}
        \large
        \centering
        \includegraphics[scale=0.48]{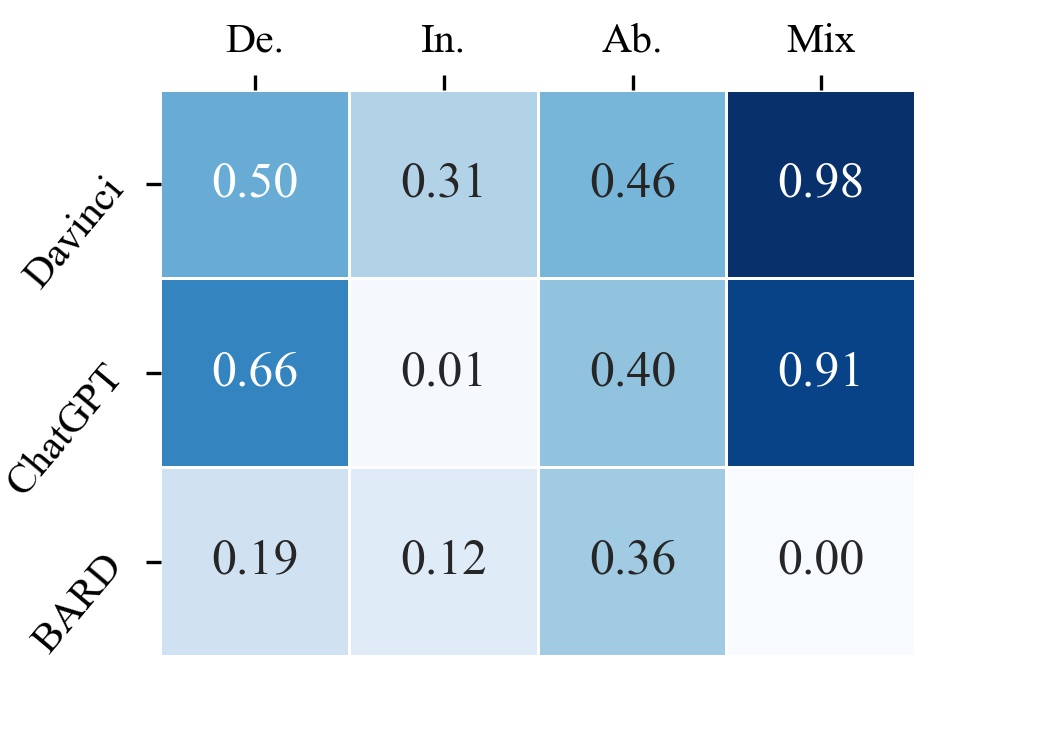}
        \vspace{-0.3cm}
        \subcaption{Self-aware evaluation.}
        \label{heatmap_redundant}
    \end{minipage}
    \caption{Heatmap visualization of rigor and self-awareness.}
\end{figure}
Secondly, we conduct a detailed analysis of LLMs' performances from the generation and classification perspectives in Fig.~\ref{bar_type}.
In general, classification scenarios tend to yield better performance than generation counterparts.
Notably, \textbf{ChatGPT exhibits particularly poor results in generation tasks}, such as bAbI-15, bAbI-16, CLUTRR, AbductiveRules and D*-Ab.
This observation may result from the fact that ChatGPT is designed to improve chatting capability rather than complex reasoning, which can lead to performance degradation in pure generative logic reasoning scenarios.

Thirdly, few-shot in-context learning (ICL) \cite{dong2022survey} does not necessarily bring improvements in logical reasoning tasks. It is quite inconsistent with the cases in other non-reasoning NLP tasks, such as topic classification and sentiment analysis \cite{wu2022self}. We count the cases where LLMs can continuously obtain the performance gains from few-shot ICL (i.e., 0-shot $\textless$ 1-shot $\textless$ 3-shot). For text-davinci-003, only two (out of four) abductive datasets continuously benefit from the few-shot ICL. ChatGPT witnesses performance improvements only in one (out of five) deductive dataset and one (out of four) abductive dataset. For BARD, few-shot ICL helps two (out of five) deductive datasets and one (out of four) abductive datasets.
\textbf{Remarkably, few-shot ICL fails to provide consistent benefits for LLMs under inductive reasoning and mixed-form reasoning manners.} We argue that inductive and mixed-form settings require more complex and high-order reasoning ability, which may be difficult to learn with a few question-answer samples without rationales.
But the task form of deductive and abductive reasoning is easy to follow, which provides the application potential for few-shot ICL.

\section{Fine-level Evaluations}

\begin{table*}[t!]
	\centering
	\small
	\caption{Evaluations on whether LLMs are rigorous reasoners. For each dataset, the first row of results represents the performances when LLMs give the correct answer, correct explanation as well as complete explanation simultaneously. The values in the subscripts denote the drops compared with only distinguishing the answer correctness. The second row of results represents the performances when LLMs give both correct answers and correct explanations, regardless of the explanation completeness. The third row represents the cases when LLMs give correct answers and list complete explanations, regardless of the explanation correctness.}
	\begin{tabular}{c|c|p{1.18cm}<{\centering}p{1.18cm}<{\centering}p{1.23cm}<{\centering}|p{1.18cm}<{\centering}p{1.18cm}<{\centering}p{1.23cm}<{\centering}|p{1.18cm}<{\centering}p{1.18cm}<{\centering}p{1.23cm}<{\centering}}
		\toprule
             & \multicolumn{1}{c|}{\multirow{2}{*}{\textbf{Dataset}}} &\multicolumn{3}{c|}{\textbf{text-davinci-003}} &\multicolumn{3}{c|}{\textbf{ChatGPT}} & \multicolumn{3}{c}{\textbf{BARD}} \\
            & \multicolumn{1}{c|}{} &\textbf{0-shot} & \textbf{1-shot} & \textbf{3-shot} &\textbf{0-shot} & \textbf{1-shot} & \textbf{3-shot} &\textbf{0-shot} & \textbf{1-shot} & \textbf{3-shot} \\
            \hline 
            \multicolumn{1}{c|}{\multirow{15}{*}{\textbf{\rotatebox{90}{Deductive}}}} & \multicolumn{1}{c|}{\multirow{3}{*}{\textbf{bAbI-15}}} &53.00$_{32.00 \downarrow}$ &60.00$_{16.00 \downarrow}$ &64.00$_{11.00 \downarrow}$ &25.50$_{12.90 \downarrow}$ &12.10$_{34.30 \downarrow}$ &14.10$_{25.60 \downarrow}$ &45.00$_{34.00 \downarrow}$ &25.00$_{55.00 \downarrow}$ &47.00$_{41.00 \downarrow}$ \\
            & &61.00 &66.00 &68.00 &32.10 &12.90 &16.40 &77.00 &74.00 &85.00 \\
            & &56.00 &60.00 &64.00 &27.90 &18.00 &17.60 &45.00 &25.00 &47.00 \\
            \cline{3-11}
            & \multicolumn{1}{c|}{\multirow{3}{*}{\textbf{EntailmentBank}}} &29.00$_{64.00 \downarrow}$ &37.00$_{51.00\downarrow}$ &30.00$_{59.00 \downarrow}$ &25.88$_{57.94 \downarrow}$ &20.00$_{62.06 \downarrow}$ &10.59$_{67.35 \downarrow}$ &26.00$_{38.00 \downarrow}$ &25.00$_{32.00 \downarrow}$ &33.00$_{37.00 \downarrow}$ \\
            & &29.00 &37.00 &30.00 &25.88 &20.00 &10.59 &54.00 &66.00 &71.00 \\
            & &72.00 &73.00 &75.00 &62.06 &57.65 &31.76 &94.00 &96.00 &97.00 \\
            \cline{3-11}
            & \multicolumn{1}{c|}{\multirow{3}{*}{\textbf{RuleTaker}}} &35.30$_{6.70 \downarrow}$ &22.50$_{15.50\downarrow}$ &24.80$_{15.40 \downarrow}$ &25.88$_{57.94 \downarrow}$ &20.00$_{62.06 \downarrow}$ &10.59$_{67.35 \downarrow}$ &54.00$_{42.00 \downarrow}$ &66.00$_{31.00 \downarrow}$ &71.00$_{26.00 \downarrow}$ \\
            & &36.20 &24.00 &26.00 &25.88 &20.00 &10.59 &26.00 &28.00 &33.00 \\
            & &36.00 &23.20 &26.00 &62.06 &57.65 &31.76 &26.00 &25.00 &34.00 \\
            \cline{3-11}
            & \multicolumn{1}{c|}{\multirow{3}{*}{\textbf{FOLIO}}} &27.00$_{21.00 \downarrow}$ &25.00$_{28.00\downarrow}$ &25.00$_{27.00 \downarrow}$ &28.92$_{21.08 \downarrow}$ &27.94$_{23.04 \downarrow}$ &27.94$_{26.47 \downarrow}$ &21.00$_{31.00 \downarrow}$ &19.00$_{24.00 \downarrow}$ &20.00$_{29.00 \downarrow}$ \\
            & &28.00 &25.00 &26.00 &28.92 &27.94 &27.94 &23.00 &19.00 &22.00 \\
            & &27.00 &28.00 &27.00 &33.33 &32.35 &35.29 &25.00 &22.00 &23.00 \\
            \cline{3-11}
            & \multicolumn{1}{c|}{\multirow{3}{*}{\textbf{Leap-of-Thought}}} &29.00$_{53.00 \downarrow}$ &43.00$_{47.00\downarrow}$ &38.00$_{49.00 \downarrow}$ &70.60$_{2.02 \downarrow}$ &24.36$_{49.65 \downarrow}$ &28.86$_{32.35 \downarrow}$ &76.00$_{3.00 \downarrow}$ &69.00$_{3.00 \downarrow}$ &77.00$_{2.00 \downarrow}$ \\
            & &63.00 &63.00 &58.00 &71.22 &24.83 &29.79 &76.00 &69.00 &77.00 \\
            & &32.00 &46.00 &42.00 &70.99 &48.33 &40.88 &79.00 &71.00 &7.00 \\
            \hline
            \multicolumn{1}{c|}{\multirow{6}{*}{\textbf{\rotatebox{90}{Inductive}}}}& \multicolumn{1}{c|}{\multirow{3}{*}{\textbf{bAbI-16}}} &59.00$_{25.00 \downarrow}$ &35.00$_{46.00\downarrow}$ &23.00$_{51.00 \downarrow}$ &8.30$_{8.80 \downarrow}$ &8.20$_{16.50 \downarrow}$ &2.60$_{10.30 \downarrow}$ &24.00$_{49.00 \downarrow}$ &15.00$_{29.00 \downarrow}$ &16.00$_{36.00 \downarrow}$ \\
            & &67.00 &50.00 &37.00 &10.20 &8.40 &3.10 &58.00 &32.00 &24.00 \\
            & &65.00 &44.00 &35.00 &10.00 &9.40 &3.50 &32.00 &22.00 &24.00 \\
            \cline{3-11}
            & \multicolumn{1}{c|}{\multirow{3}{*}{\textbf{CLUTRR}}} &2.00$_{4.00 \downarrow}$ &7.00$_{16.00\downarrow}$ &6.00$_{14.00 \downarrow}$ &7.85$_{14.14 \downarrow}$ &4.62$_{14.92 \downarrow}$ &2.71$_{10.12 \downarrow}$ &19.00$_{4.00 \downarrow}$ &25.00$_{1.00 \downarrow}$ &23.00$_{1.00 \downarrow}$ \\
            & &2.00 &7.00 &6.00 &7.94 &4.62 &2.71 &19.00 &25.00 &23.00 \\
            & &6.00 &18.00 &18.00 &10.56 &6.72 &3.75 &21.00 &26.00 &23.00 \\
            \hline
            \multicolumn{1}{c|}{\multirow{12}{*}{\textbf{\rotatebox{90}{Abductive}}}}& \multicolumn{1}{c|}{\multirow{3}{*}{\textbf{$\alpha$-NLI}}} &68.00$_{6.00 \downarrow}$ &69.00$_{1.00\downarrow}$ &68.00$_{6.00 \downarrow}$ &77.50$_{3.40 \downarrow}$ &64.50$_{15.50 \downarrow}$ &58.40$_{20.70 \downarrow}$ &75.00$_{0.00 -}$ &71.00$_{3.00 \downarrow}$ &77.00$_{0.00 -}$ \\
            & &70.00 &69.00 &68.00 &78.20 &66.70 &60.90 &75.00 &71.00 &77.00 \\
            & &68.00 &69.00 &68.00 &77.60 &64.50 &58.40 &75.00 &71.00 &77.00 \\
            \cline{3-11}
            & \multicolumn{1}{c|}{\multirow{3}{*}{\textbf{$\alpha$-NLG}}} &1.00$_{8.00 \downarrow}$ &0.00$_{10.00\downarrow}$ &2.00$_{10.00 \downarrow}$ &15.30$_{6.60 \downarrow}$ &16.00$_{7.40 \downarrow}$ &10.30$_{15.60 \downarrow}$ &7.00$_{3.00 \downarrow}$ &8.00$_{4.00 \downarrow}$ &9.00$_{6.00 \downarrow}$ \\
            & &1.00 &0.00 &2.00 &15.30 &16.00 &10.40 &7.00 &9.00 &10.00 \\
            & &8.00 &8.00 &9.00 &20.90 &22.40 &21.40 &10.00 &9.00 &14.00 \\
            \cline{3-11}
            & \multicolumn{1}{c|}{\multirow{3}{*}{\textbf{AbductiveRules}}} &50.00$_{25.00 \downarrow}$ &5.00$_{37.00\downarrow}$ &0.00$_{35.00 \downarrow}$ &12.00$_{11.30 \downarrow}$ &18.30$_{16.80 \downarrow}$ &5.70$_{24.10 \downarrow}$ &57.00$_{14.00 \downarrow}$ &30.00$_{19.00 \downarrow}$ &10.00$_{12.00 \downarrow}$ \\
            & &75.00 &10.00 &0.00 &20.50 &29.60 &9.50 &65.00 &40.00 &18.00 \\
            & &50.00 &5.00 &0.00 &13.00 &19.90 &5.80 &57.00 &30.00 &10.00 \\
            \cline{3-11}
            & \multicolumn{1}{c|}{\multirow{3}{*}{\textbf{D*-Ab}}} &4.00$_{4.00 \downarrow}$ &5.00$_{16.00\downarrow}$ &4.00$_{19.00 \downarrow}$ &4.10$_{7.50 \downarrow}$ &1.20$_{1.30 \downarrow}$ &1.10$_{0.70 \downarrow}$ &4.00$_{7.00 \downarrow}$ &0.00$_{0.00 -}$ &0.00$_{0.00 -}$ \\
            & &7.00 &7.00 &7.00 &5.50 &1.50 &1.20 &4.00 &0.00 &0.00 \\
            & &4.00 &5.00 &4.00 &4.50 &1.20 &1.10 &5.00 &0.00 &0.00 \\
            \hline
            \multicolumn{1}{c|}{\multirow{12}{*}{\textbf{\rotatebox{90}{Mixed-form}}}}& \multicolumn{1}{c|}{\multirow{3}{*}{\textbf{ReClor}}} &5.00$_{48.00 \downarrow}$ &0.00$_{53.00\downarrow}$ &0.00$_{55.00 \downarrow}$ &28.60$_{30.20 \downarrow}$ &25.20$_{30.80 \downarrow}$ &29.80$_{29.00 \downarrow}$ &38.00$_{18.00 \downarrow}$ &34.00$_{21.00 \downarrow}$ &33.00$_{23.00 \downarrow}$ \\
            & &5.00 &0.00 &0.00 &32.40 &28.00 &32.20 &48.00 &46.00 &42.00 \\
            & &42.00 &46.00 &42.00 &50.00 &32.80 &46.00 &38.00 &36.00 &38.00 \\
            \cline{3-11}
            & \multicolumn{1}{c|}{\multirow{3}{*}{\textbf{LogiQA}}} &5.00$_{36.00 \downarrow}$ &2.00$_{33.00\downarrow}$ &1.00$_{38.00 \downarrow}$ &25.96$_{14.29 \downarrow}$ &21.35$_{18.13 \downarrow}$ &20.43$_{20.43 \downarrow}$ &29.00$_{19.00 \downarrow}$ &26.00$_{20.00 \downarrow}$ &19.00$_{28.00 \downarrow}$ \\
            & &6.00 &2.00 &1.00 &28.42 &23.20 &23.81 &34.00 &27.00 &23.00 \\
            & &23.00 &26.00 &27.00 &29.19 &27.50 &26.42 &30.00 &35.00 &27.00 \\
            \cline{3-11}
            & \multicolumn{1}{c|}{\multirow{3}{*}{\textbf{LogiQA2.0}}} &38.00$_{5.00 \downarrow}$ &32.00$_{10.00\downarrow}$ &28.00$_{13.00 \downarrow}$ &43.40$_{11.20 \downarrow}$ &34.60$_{16.20 \downarrow}$ &38.80$_{16.00 \downarrow}$ &47.00$_{6.00 \downarrow}$ &39.00$_{7.00 \downarrow}$ &44.00$_{3.00 \downarrow}$ \\
            & &39.00 &32.00 &28.00 &43.80 &35.20 &38.80 &47.00 &39.00 &44.00 \\
            & &39.00 &32.00 &30.00 &44.20 &36.80 &40.60 &47.00 &39.00 &44.00 \\
            \cline{3-11}
            & \multicolumn{1}{c|}{\multirow{3}{*}{\textbf{LogiQA2NLI}}} &57.00$_{2.00 \downarrow}$ &51.00$_{4.00\downarrow}$ &56.00$_{2.00 \downarrow}$ &43.17$_{14.67 \downarrow}$ &36.33$_{17.50 \downarrow}$ &36.50$_{20.50 \downarrow}$ &38.00$_{10.00 \downarrow}$ &41.00$_{9.00 \downarrow}$ &37.00$_{10.00 \downarrow}$ \\
            & &57.00 &51.00 &56.00 &43.50 &36.50 &36.83 &40.00 &43.00 &40.00 \\
            & &57.00 &51.00 &56.00 &53.50 &49.00 &50.16 &38.00 &41.00 &37.00 \\
		\bottomrule
	\end{tabular}
	\vspace{-0.2cm}
	\label{explain_true_complete}
\end{table*}

This section presents a detailed analysis of the logical reasoning capabilities of LLMs from various perspectives.

\subsection{Are LLMs Rigorous Logical Reasoning?}

\begin{table*}[t!]
	\centering
	\small
	\caption{Evaluation results on the metric of explanation redundancy.}
	\begin{tabular}{c|c|c|ccc|ccc|ccc}
		\toprule
            & \multicolumn{1}{c|}{\multirow{2}{*}{\textbf{Dataset}}} & \multicolumn{1}{c|}{\multirow{2}{*}{\textbf{Gen.}}} & \multicolumn{3}{c|}{\textbf{text-davinci-003}} & \multicolumn{3}{c|}{\textbf{ChatGPT}} & \multicolumn{3}{c}{\textbf{BARD}}\\
            & \multicolumn{1}{c|}{} & & \textbf{0-shot} & \textbf{1-shot} & \textbf{3-shot} & \textbf{0-shot} & \textbf{1-shot} & \textbf{3-shot} & \textbf{0-shot} & \textbf{1-shot} & \textbf{3-shot}\\
            \midrule 
            \multicolumn{1}{c|}{\multirow{5}{*}{\rotatebox{90}{\textbf{De.}}}} &\textbf{bAbI-15} &\checkmark &63.00 &56.00 &43.00 &22.60 &39.40 &55.70 &99.00 &84.00 &62.00\\
            & \textbf{EntailmentBank} &\checkmark &8.00 &6.00 &7.00 &7.06 &5.88 &3.24 &26.00 &25.00 &28.00 \\
            & \textbf{RuleTaker} & &26.00 &29.00 &27.00 &21.30 &27.80 &34.80 &80.00 &84.00 &75.00 \\
            & \textbf{FOLIO} & &14.00 &23.00 &21.00 &31.86 &22.55 &19.61 &60.00 &63.00 &68.00\\
            & \textbf{Leap-Of-Thought} & &71.00 &55.00 &54.00 &32.74 &5.04 &4.73 &2.00 &2.00 &0.00\\
            \midrule 
            \multicolumn{1}{c|}{\multirow{2}{*}{\rotatebox{90}{\textbf{In.}}}} & \textbf{bAbI-16} &\checkmark &60.00 &77.00 &86.00 &93.60 &29.80 &41.20 &96.00 &98.00 &99.00\\
            & \textbf{CLUTRR} &\checkmark &2.00 &28.00 &31.00 &2.62 &1.57 &0.87 &2.00 &6.00 &14.00\\
            \midrule 
            \multicolumn{1}{c|}{\multirow{4}{*}{\rotatebox{90}{\textbf{Ab.}}}} & \textbf{$\alpha$-NLI} & &2.00 &2.00 &1.00 &1.00 &0.20 &0.10 &8.00 &16.00 &0.00\\
            & \textbf{$\alpha$-NLG} &\checkmark &63.00 &61.00 &72.00 &70.70 &69.70 &64.50 &24.00 &32.00 &31.00 \\
            & \textbf{AbductiveRules} &\checkmark &1.00 &0.00 &0.00 &42.40 &5.40 &0.50 &67.00 &48.00 &22.00 \\
            & \textbf{D*-Ab} &\checkmark &85.00 &27.00 &17.00 &55.30 &27.10 &16.70 &18.00 &16.00 &2.00\\
		\midrule 
            \multicolumn{1}{c|}{\multirow{4}{*}{\rotatebox{90}{\textbf{Mix}}}} & \textbf{ReClor} & &1.00 &1.00 &1.00 &2.00 &1.20 &1.40 &11.00 &16.00 &24.00\\
            & \textbf{LogiQA} & &0.00 &5.00 &0.00 &1.54 &0.77 &1.08 &32.00 &35.00 &43.00\\
            & \textbf{LogiQA 2.0} & &0.00 &0.00 &0.00 &0.80 &4.00 &0.8 &5.00 &5.00 &4.00\\
            & \textbf{LogiQA2NLI} & &0.00 &0.00 &0.00 &0.17 &0.50 &0.17 &11.00 &31.00 &4.00 \\
		\bottomrule
	\end{tabular}
	\vspace{-0.2cm}
	\label{self-aware}
\end{table*}

While LLMs may produce correct answers in some cases, it is unclear whether they perform the correct logical reasoning or simply arrive at the right answer by chance.
Therefore, we delve deeper into the reasoning process beyond the output answers.
We view cases where LLMs provide a correct answer along with a correct and complete explanation as \emph{Rigorous}. The detailed results are shown in Table~\ref{explain_true_complete}.
Compared with the simple judgment of answer correctness, all selected LLMs present obvious performance drops. 
For simplicity, we only take the zero-shot setting into consideration. In Fig.~\ref{heatmap_rigorous}, we calculate the ratio of rigorous performance and answer accuracy. The higher values (darker colors) mean better performance in rigorous reasoning. According to the results, BARD shows the best capability in rigorous reasoning, consistently under four reasoning manners. Meanwhile, ChatGPT still struggles in deductive and inductive settings, while text-davinci-003 comes last in both abductive and mixed-form manners.

Further, LLMs are best at keeping rigorous reasoning in the abductive setting, while they are weak in the deductive and inductive settings. The finding is a little different from the analysis of simple accuracy conditions in the previous section. 
{\color{black} We argue that the setting of abductive reasoning requires the LLMs to achieve the reasoning reversely, which can activate LLMs to provide sufficient reasoning process. 
In a deductive reasoning setting, the reasoning chain is sequential, which may cause LLMs to be in a lazy mode and harm rigorous reasoning.}

In Table~\ref{explain_true_complete}, we also include the two conditions (1) when correct answer and correct explanations are satisfied, and (2) when correct answer and complete explanations are satisfied. Results vary a lot with different datasets and different LLMs. Overall, ChatGPT performs relatively well in keeping correct explanations while it may fail to maintain complete explanations in most cases. In comparison, text-davinci-003 exhibits stronger characteristics in maintaining the completeness of explanations, compared with the correctness of explanations. These findings of the respective reasoning preferences are expected to guide the future utilization of LLMs.

\subsection{Are LLMs Self-aware Logical Reasoners?}
From an alternative perspective, the redundancy of the generated content by LLMs has been a frequently discussed topic, as it is deemed an important metric for assessing their practicality. In this paper, we consider LLMs with less redundant content as more \emph{self-aware}, as they can effectively express the necessary information without outputting all possible answers. Table~\ref{self-aware} presents the evaluation results of LLMs' self-awareness. Similar to our previous approach, we compute the weighted results for each reasoning setting and derive the self-awareness scores shown in Fig.~\ref{heatmap_redundant}. The darker color indicates a stronger self-awareness capability.
Results indicate that text-davinci-003 exhibits notable advantages, particularly in the inductive, abductive, and mixed-form reasoning settings.
Additionally, it ranks second in the deductive setting. Conversely, BARD performs poorly in deductive, abductive, and mixed-form reasoning settings.

In comparison to classification tasks, LLMs tend to generate redundant answers more frequently in generation tasks, such as $\alpha$-NLI vs. $\alpha$-NLG. This is because open-ended questions can prompt LLMs to generate content from various perspectives, which can lead to the inclusion of redundant information. Furthermore, the mixed-form reasoning setting observes significantly fewer instances of redundancy. The tasks in mixed-form reasoning are primarily based on question answering, which closely resembles real-life text, and LLMs tend to generate rational and specific content in such scenarios. However, in other settings, the input context is elaborately designed for logical reasoning and may provide sufficient background information. This can result in LLMs employing embodied commonsense knowledge to help reason and thus generate additional explanations.
{\color{black}Notably, considering that current LLMs do a lot of optimizations for preference alignment, 
it is inevitable that they would produce some redundant information to cover diverse conditions.}

\begin{figure*}[t!]
	\begin{minipage}[t]{0.49\linewidth}
		\large
		\centering
		\includegraphics[scale=0.52]{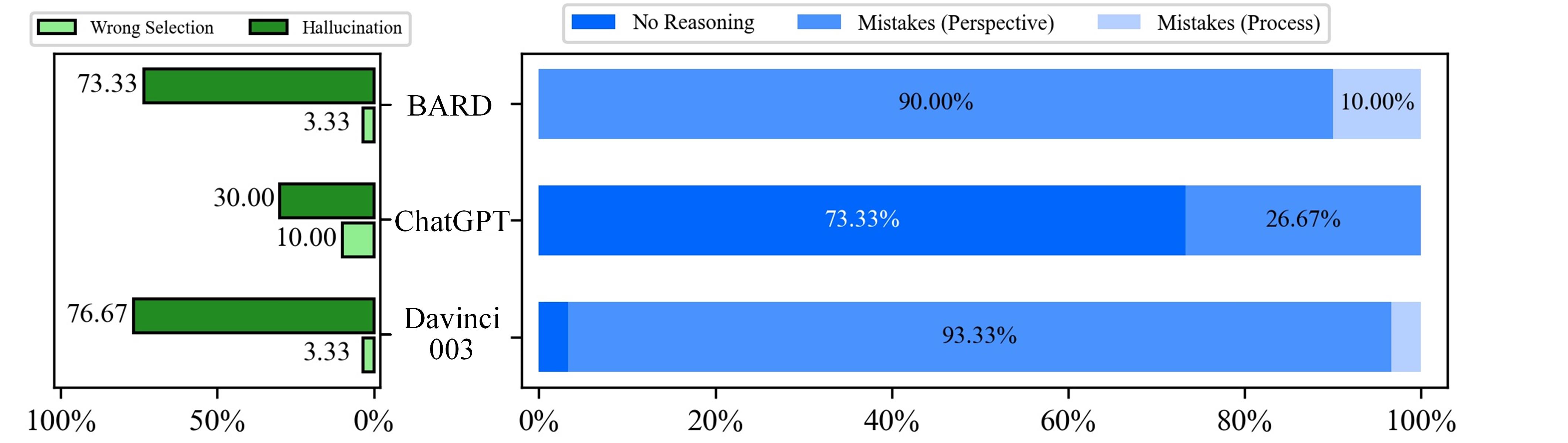}
            \vspace{-0.5cm}
		\subcaption{bAbI15 (Deductive).}
            \vspace{0.1cm}
		\label{AbductiveRules}
	\end{minipage}
        \begin{minipage}[t]{0.49\linewidth}
		\large
		\centering
		\includegraphics[scale=0.52]{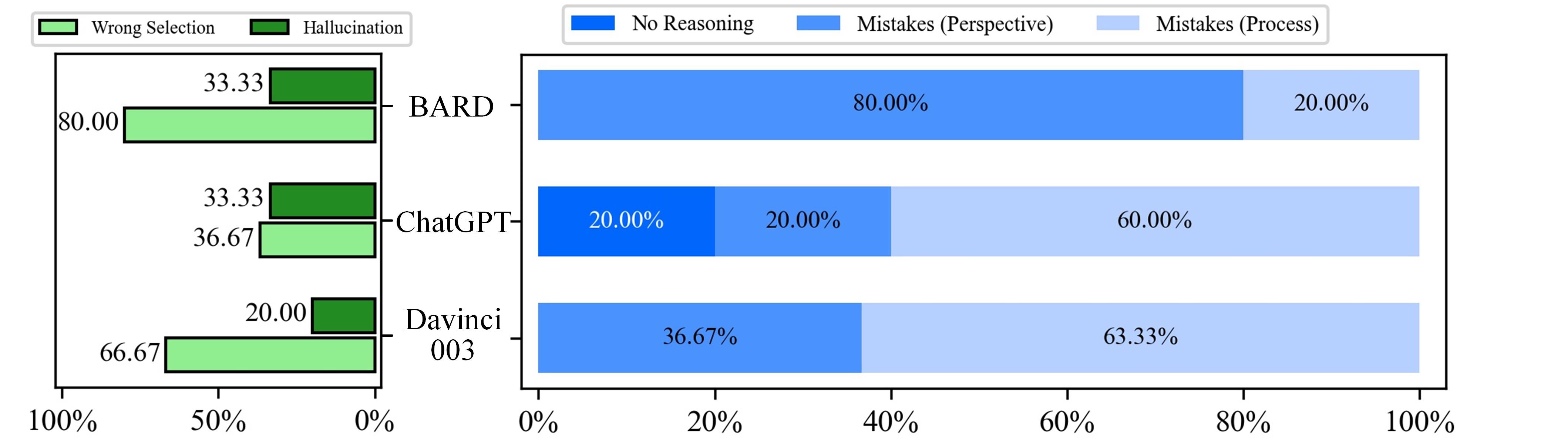}
            \vspace{-0.5cm}
		\subcaption{RuleTaker (Deductive).}
            \vspace{0.1cm}
		\label{AbductiveRules}
	\end{minipage}
        \\
	\begin{minipage}[t]{0.49\linewidth}
		\large
		\centering
		\includegraphics[scale=0.52]{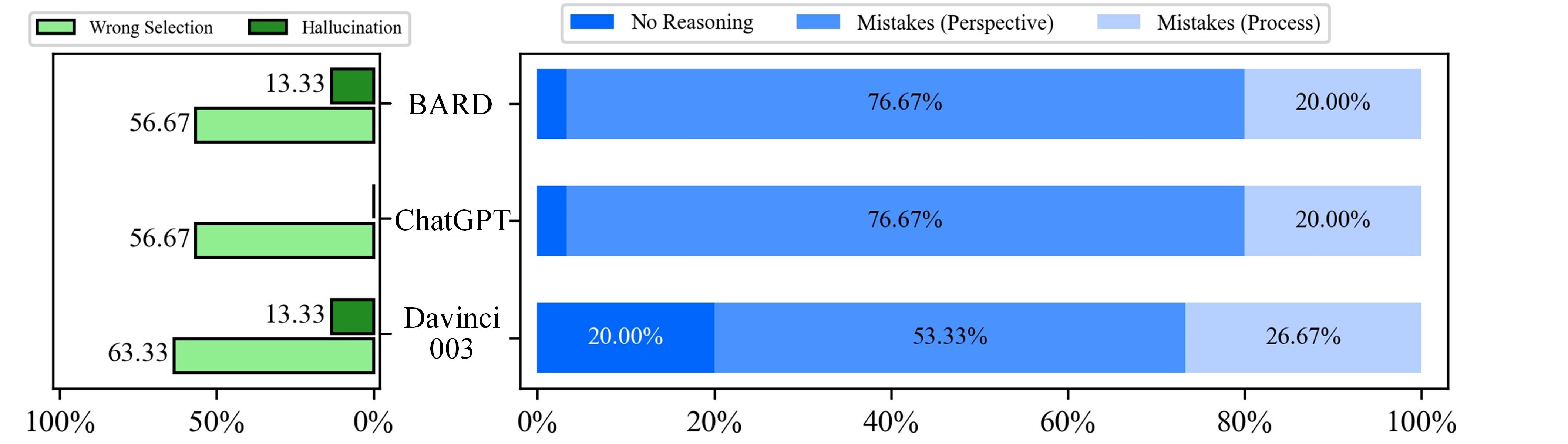}
            \vspace{-0.5cm}
		\subcaption{FOLIO (Deductive).}
            \vspace{0.1cm}
		\label{D-Ab}
	\end{minipage}
	\begin{minipage}[t]{0.49\linewidth}
		\large
		\centering
		\includegraphics[scale=0.52]{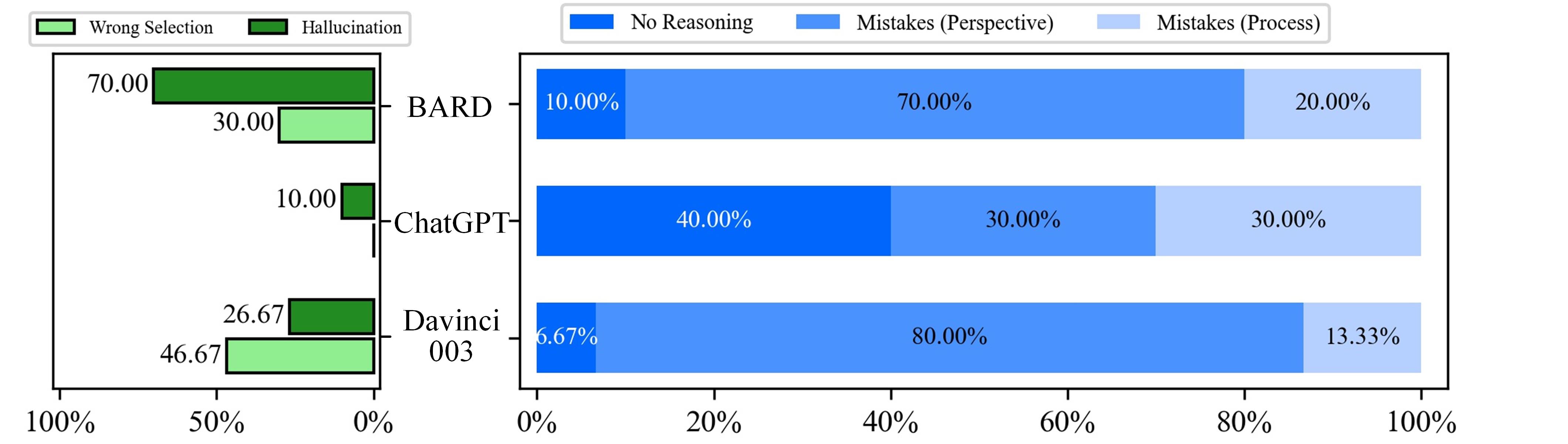}
            \vspace{-0.5cm}
		\subcaption{Leap-of-Thought (Deductive).}
            \vspace{0.1cm}		
            \label{D-Ab}
	\end{minipage}
        \\
        \begin{minipage}[t]{0.49\linewidth}
		\large
		\centering
		\includegraphics[scale=0.52]{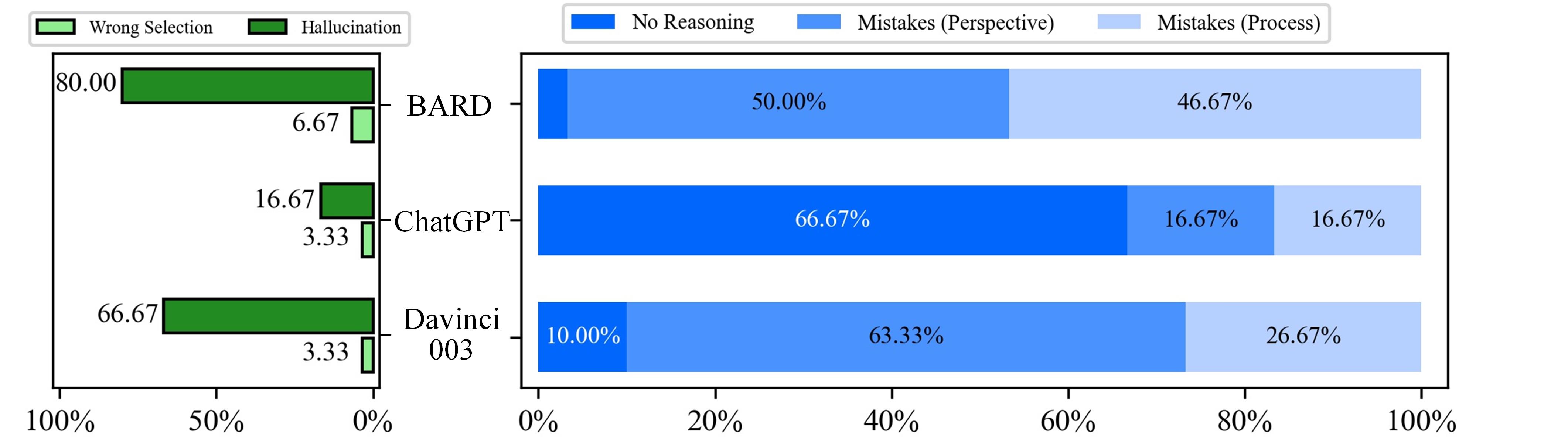}
            \vspace{-0.5cm}
		\subcaption{bAbI16 (Inductive).}
            \vspace{0.1cm}
		\label{AbductiveRules}
	\end{minipage}
	\begin{minipage}[t]{0.49\linewidth}
		\large
		\centering
		\includegraphics[scale=0.52]{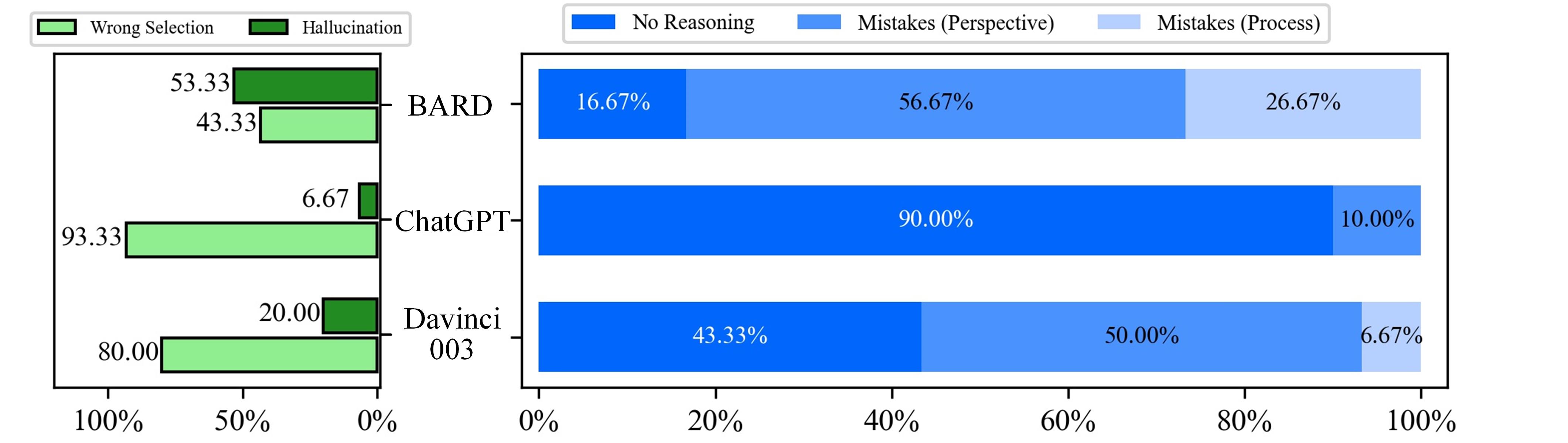}
            \vspace{-0.5cm}
		\subcaption{CLUTRR (Inductive).}
            \vspace{0.1cm}
		\label{D-Ab}
	\end{minipage}
        \\
	\begin{minipage}[t]{0.49\linewidth}
		\large
		\centering
		\includegraphics[scale=0.52]{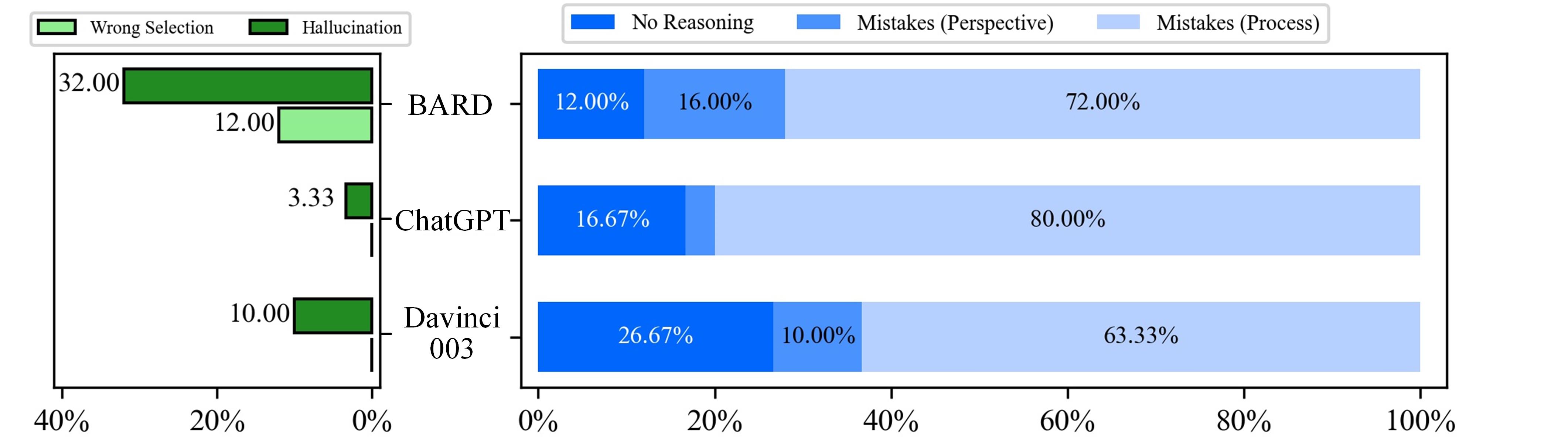}
            \vspace{-0.5cm}
		\subcaption{$\alpha$-NLI (Abductive).}
            \vspace{0.1cm}
		\label{AbductiveRules}
	\end{minipage}
	\begin{minipage}[t]{0.49\linewidth}
		\large
		\centering
		\includegraphics[scale=0.52]{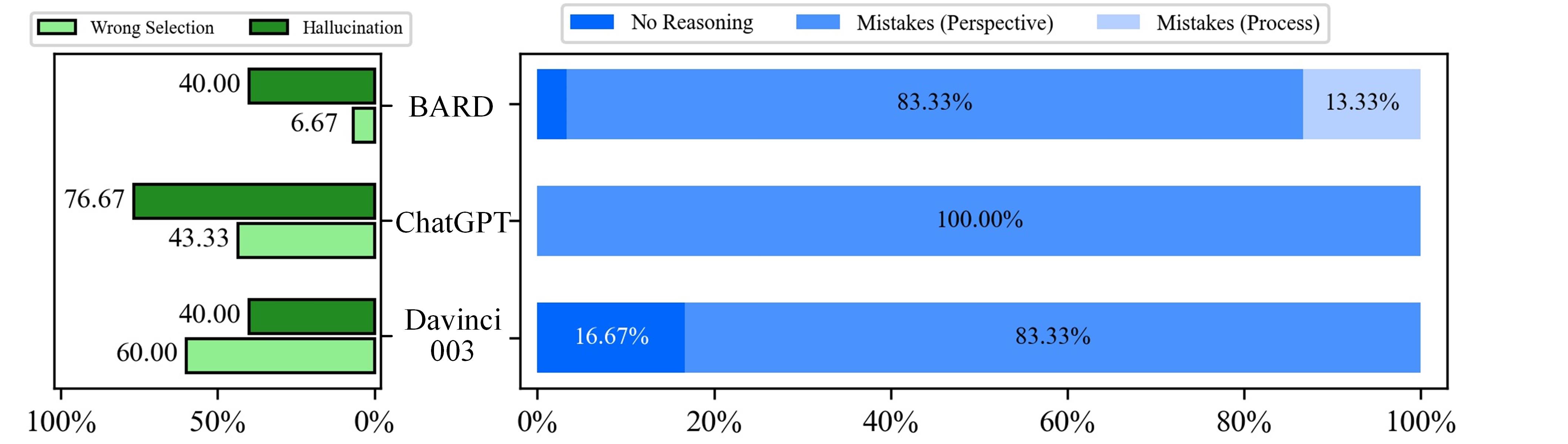}
            \vspace{-0.5cm}
		\subcaption{$\alpha$-NLG (Abductive).}
            \vspace{0.1cm}
		\label{D-Ab}
	\end{minipage}	
        \\
        \begin{minipage}[t]{0.49\linewidth}
		\large
		\centering
		\includegraphics[scale=0.52]{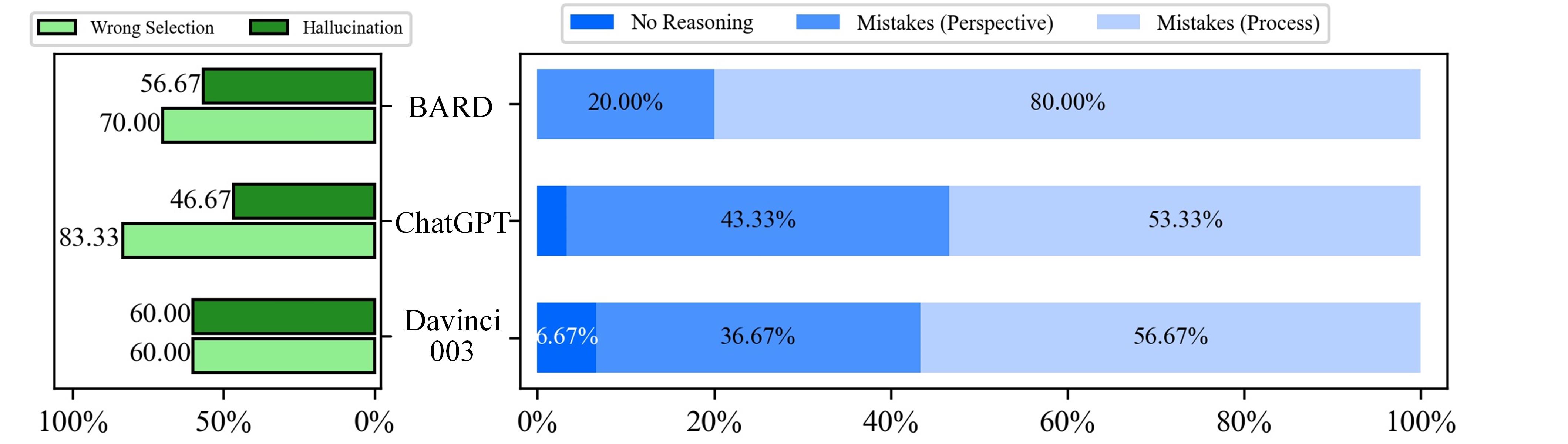}
            \vspace{-0.5cm}
		\subcaption{AbductiveRules (Abductive).}
            \vspace{0.1cm}		
            \label{AbductiveRules}
	\end{minipage}
	\begin{minipage}[t]{0.49\linewidth}
		\large
		\centering
		\includegraphics[scale=0.52]{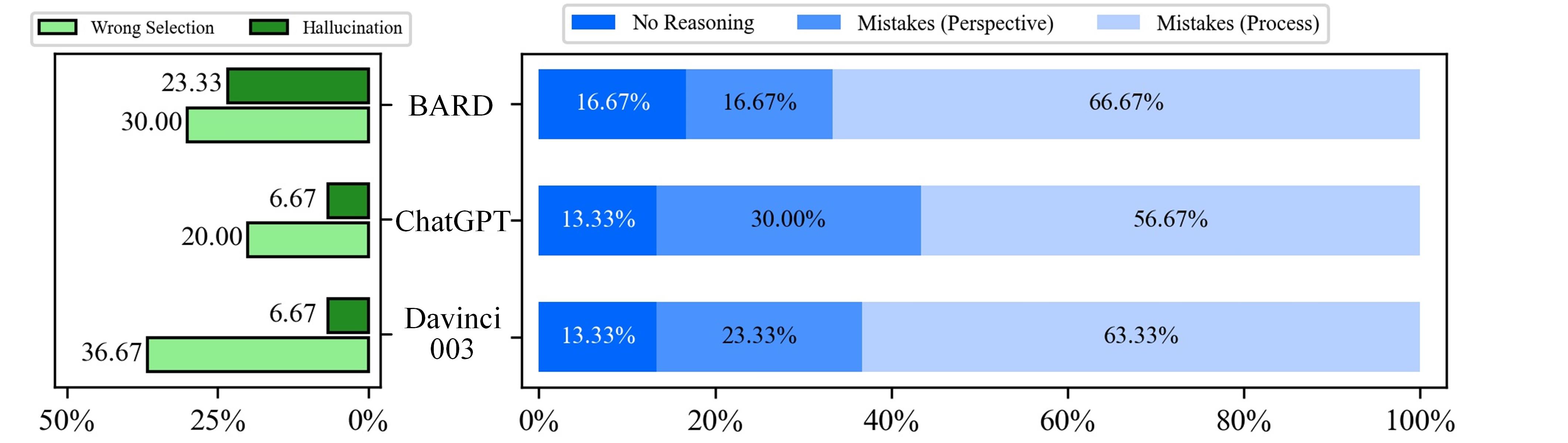}
            \vspace{-0.5cm}
		\subcaption{D*-Ab (Abductive).}
            \vspace{0.1cm}
		\label{D-Ab}
	\end{minipage}
        \\
	\begin{minipage}[t]{0.49\linewidth}
		\large
		\centering
		\includegraphics[scale=0.52]{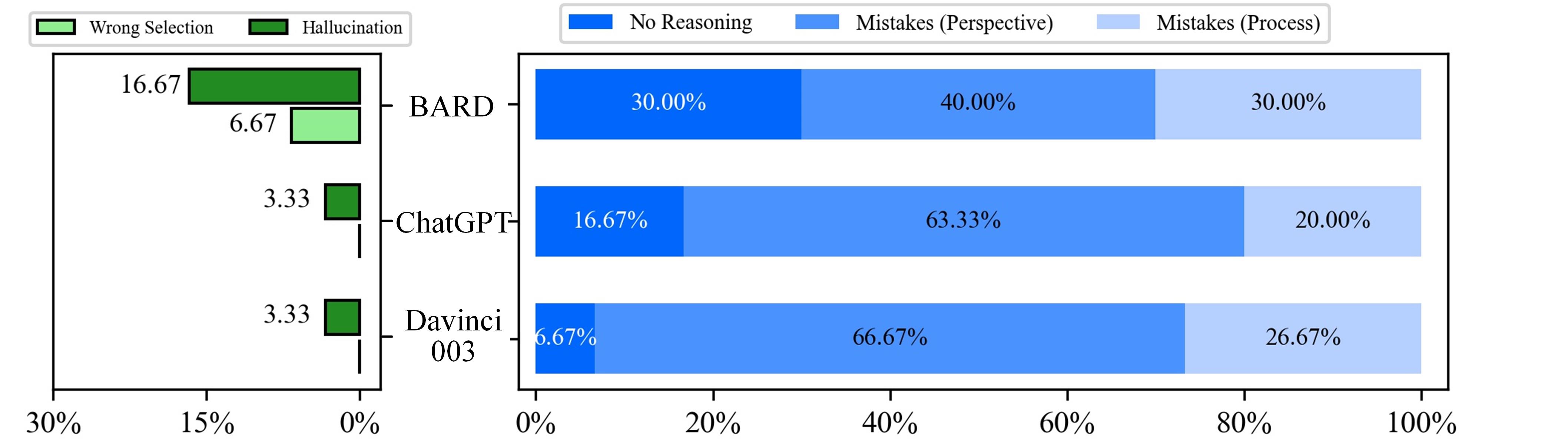}
            \vspace{-0.5cm}
		\subcaption{ReClor (Mixed-form).}
            \vspace{0.1cm}
		\label{AbductiveRules}
	\end{minipage}
	\begin{minipage}[t]{0.49\linewidth}
		\large
		\centering
		\includegraphics[scale=0.52]{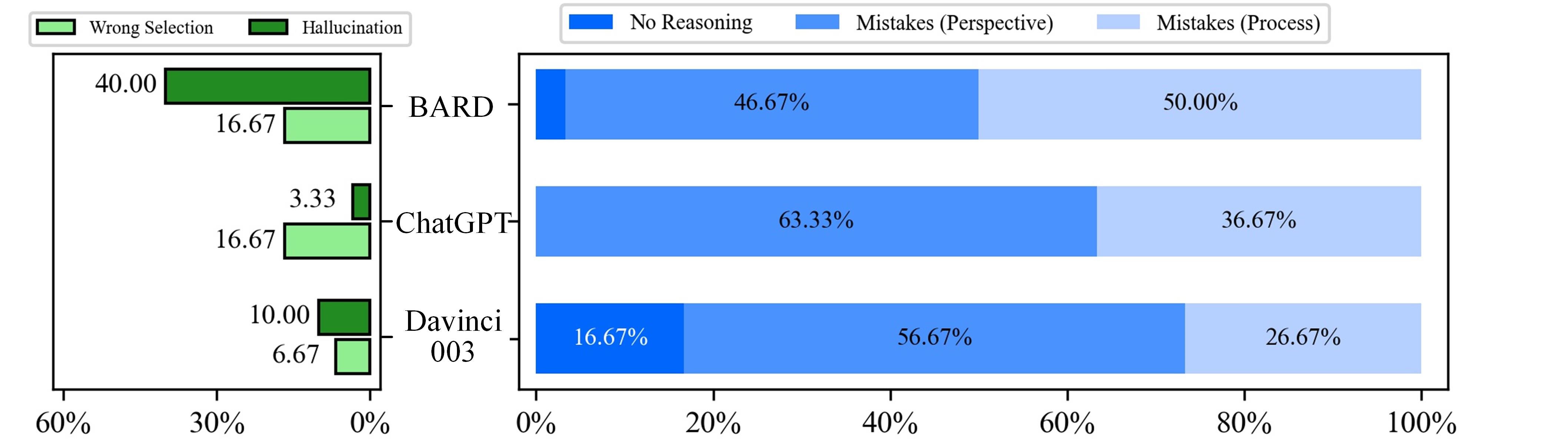}
            \vspace{-0.5cm}
		\subcaption{LogiQA (Mixed-form).}
            \vspace{0.1cm}
		\label{D-Ab}
	\end{minipage}
        \\
	\begin{minipage}[t]{0.49\linewidth}
		\large
		\centering
		\includegraphics[scale=0.52]{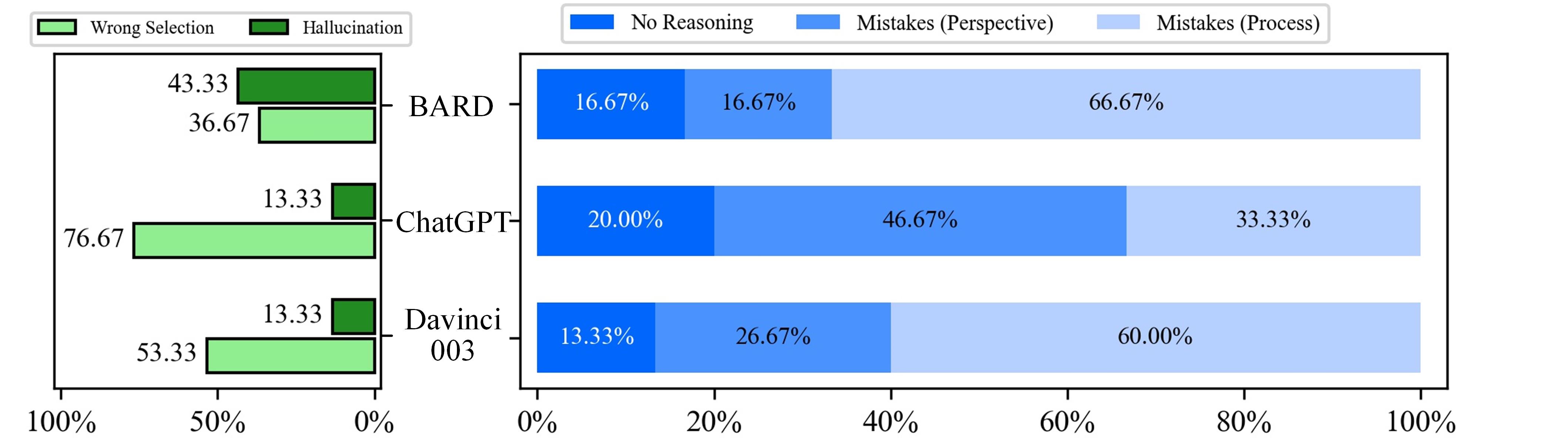}
            \vspace{-0.5cm}
		\subcaption{LogiQA2.0 (Mixed-form).}
            \vspace{0.1cm}
		\label{AbductiveRules}
	\end{minipage}
	\begin{minipage}[t]{0.49\linewidth}
		\large
		\centering
		\includegraphics[scale=0.52]{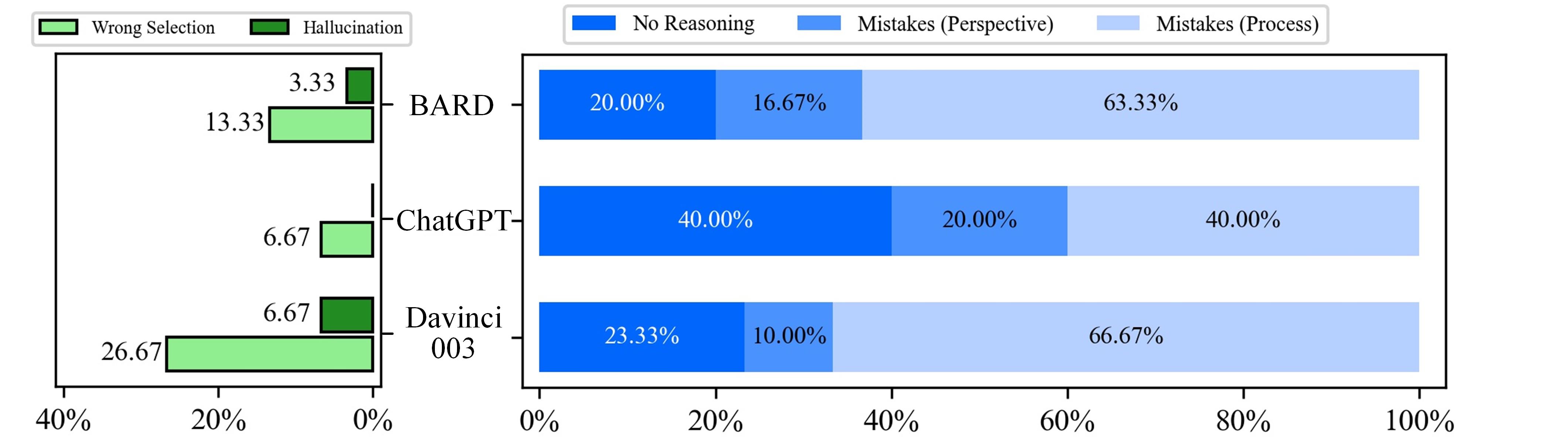}
            \vspace{-0.5cm}
		\subcaption{LogiQA2NLI (Mixed-form).}
            \vspace{0.1cm}
		\label{D-Ab}
	\end{minipage}
 
	\caption{Statistics of different error types from \emph{evidence selection process} and \emph{reasoning process} view. {\color{black} The light green bar is \emph{wrong selection} and the dark green bar means \emph{hallucination}. From dark to light blue, the bars represent \emph{No Reasoning}, \emph{Mistakes (Perspective)}, and \emph{Mistakes (Process)}.}}
	\label{logic_error}
\end{figure*}

\subsection{Do LLMs Have Obvious Logical Flaws?}

Based on the preceding statements, we establish error types for problematic cases (with incorrect explanations) from two dimensions: \emph{evidence selection process} and \emph{reasoning process}. The former dimension encompasses two error types: (i) \emph{wrong selection} and (ii) \emph{hallucination}, which are independent of each other. The latter dimension comprises three error types: (i) \emph{no reasoning}, (ii) \emph{perspective mistake} and (iii) \emph{process mistake}. Each problematic case can only be attributed to one of the three errors in the \emph{reasoning process} dimension.

We visualize the attribution results for fourteen datasets in Fig.~\ref{logic_error}, including four deductive, two inductive, four abductive and four mixed-form ones.
Overall speaking, the types of errors vary between datasets.
For the \emph{wrong selection}, 33.26\% of the problematic cases fail to select the right answers for reasoning. Also, 27.46\% suffers from the hallucination issue of LLMs. From the dimension of \emph{reasoning process}, \emph{no reasoning} error keeps a small portion in most of the cases, only covering 19.33\% of the selected cases in total. Meanwhile, \emph{perspective mistake} occupies 44.47\% of the cases and \emph{process mistake} covers 36.20\%.

\begin{figure*}[t]
    \begin{minipage}[t]{0.33\linewidth}
        \large
        \centering
        \includegraphics[scale=0.62]{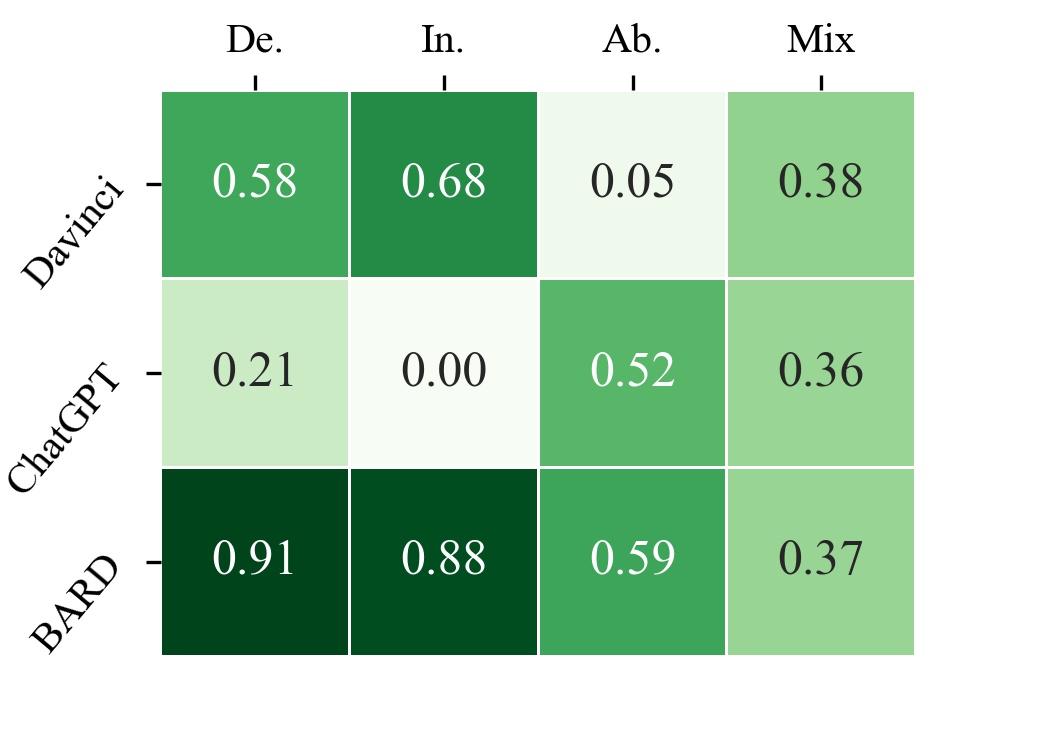}
        \vspace{-0.1cm}
        \subcaption{Visualization on the activity.}
        \label{heatmap_active}
    \end{minipage}
        \begin{minipage}[t]{0.33\linewidth}
        \large
        \centering
        \includegraphics[scale=0.62]{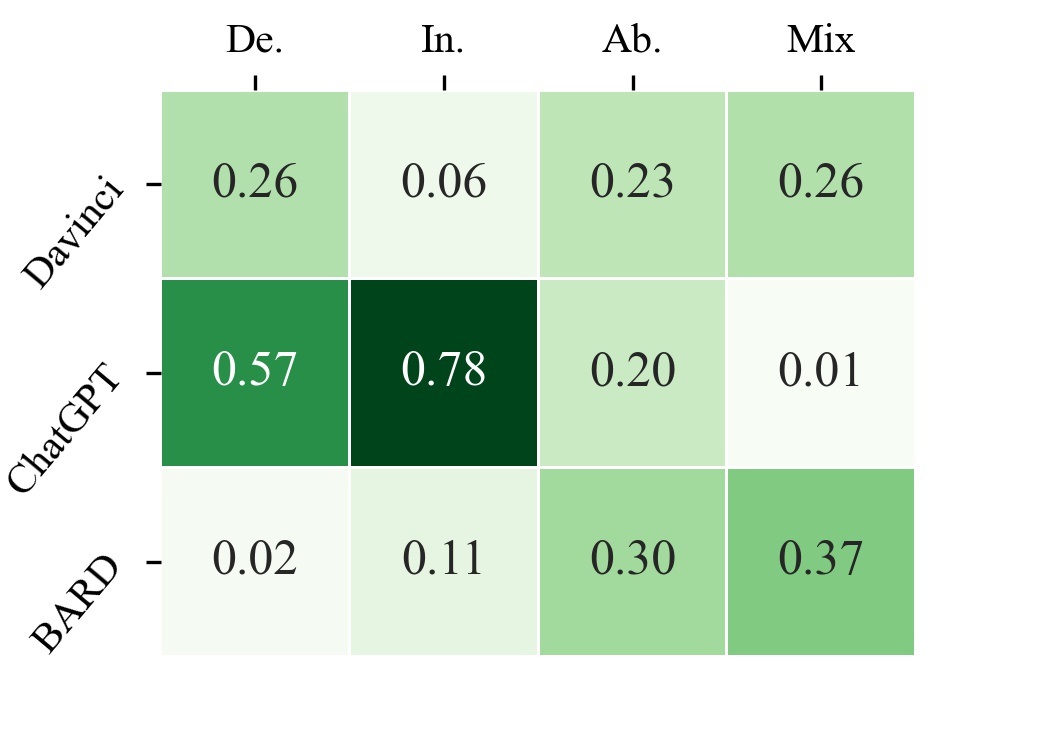}
        \vspace{-0.1cm}
        \subcaption{Visualization on the orientation.}
        \label{heatmap_oriented}
    \end{minipage}
    \begin{minipage}[t]{0.33\linewidth}
        \large
        \centering
        \includegraphics[scale=0.62]{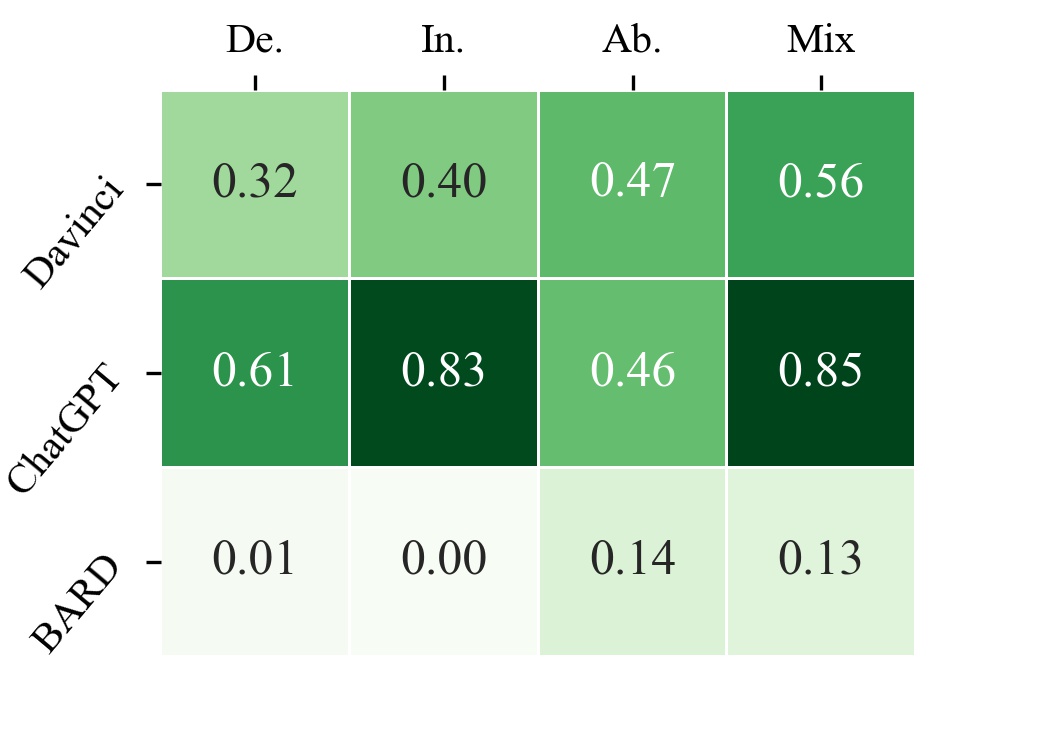}
        \vspace{-0.1cm}
        \subcaption{Visualization on the no-hallucination.}
        \label{heatmap_hallu}
    \end{minipage}
    \caption{Heatmap results for the activity, orientation and no-hallucination of LLMs.}
    \label{heatmap_error}
\end{figure*}

\begin{figure}[t]
	\begin{minipage}[t]{0.49\linewidth}
		\large
		\centering
		\includegraphics[scale=0.32]{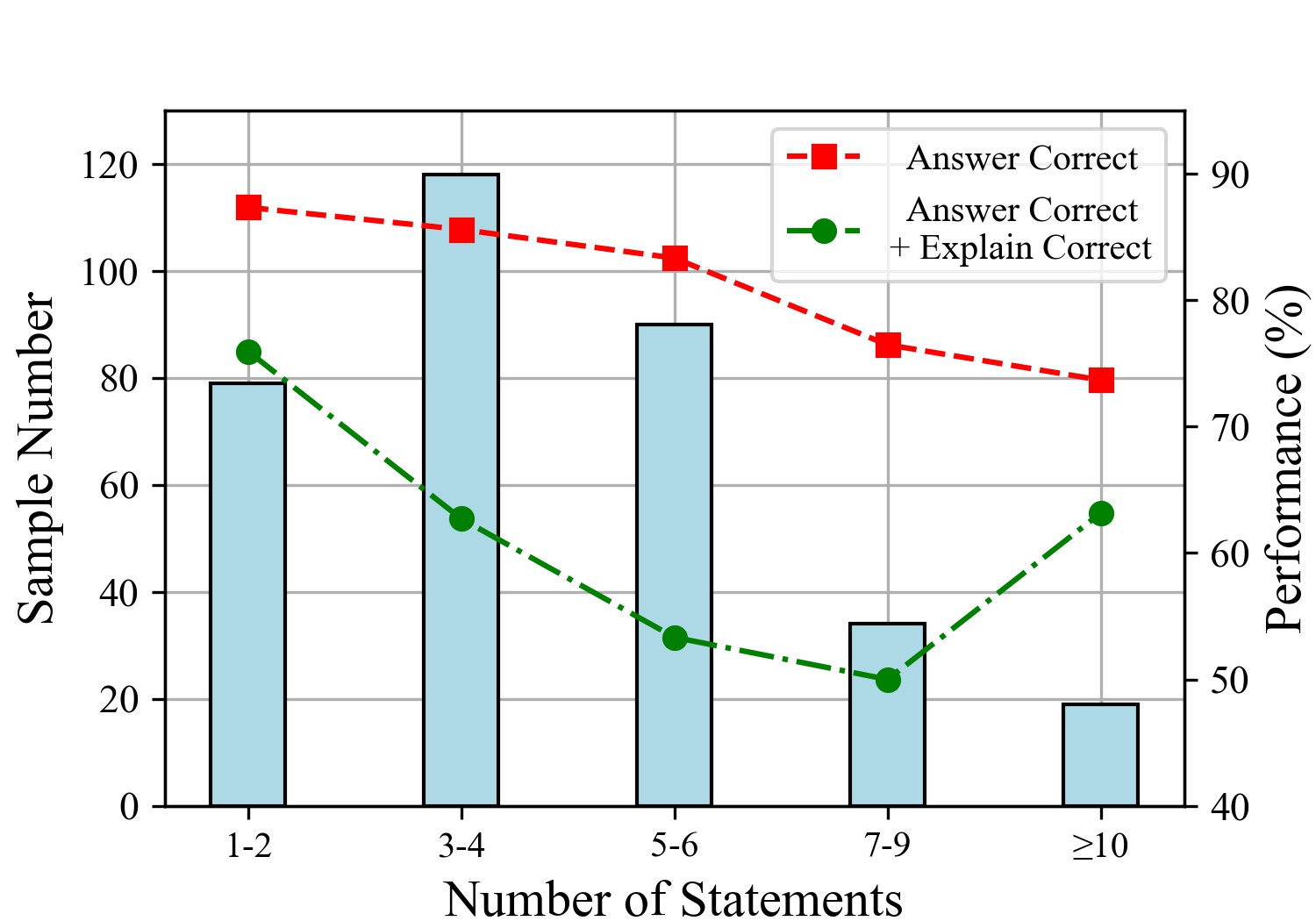}
            \vspace{-0.1cm}
		\subcaption{EntailmentBank (De.).}
	\end{minipage}
        \begin{minipage}[t]{0.49\linewidth}
		\large
		\centering
		\includegraphics[scale=0.32]{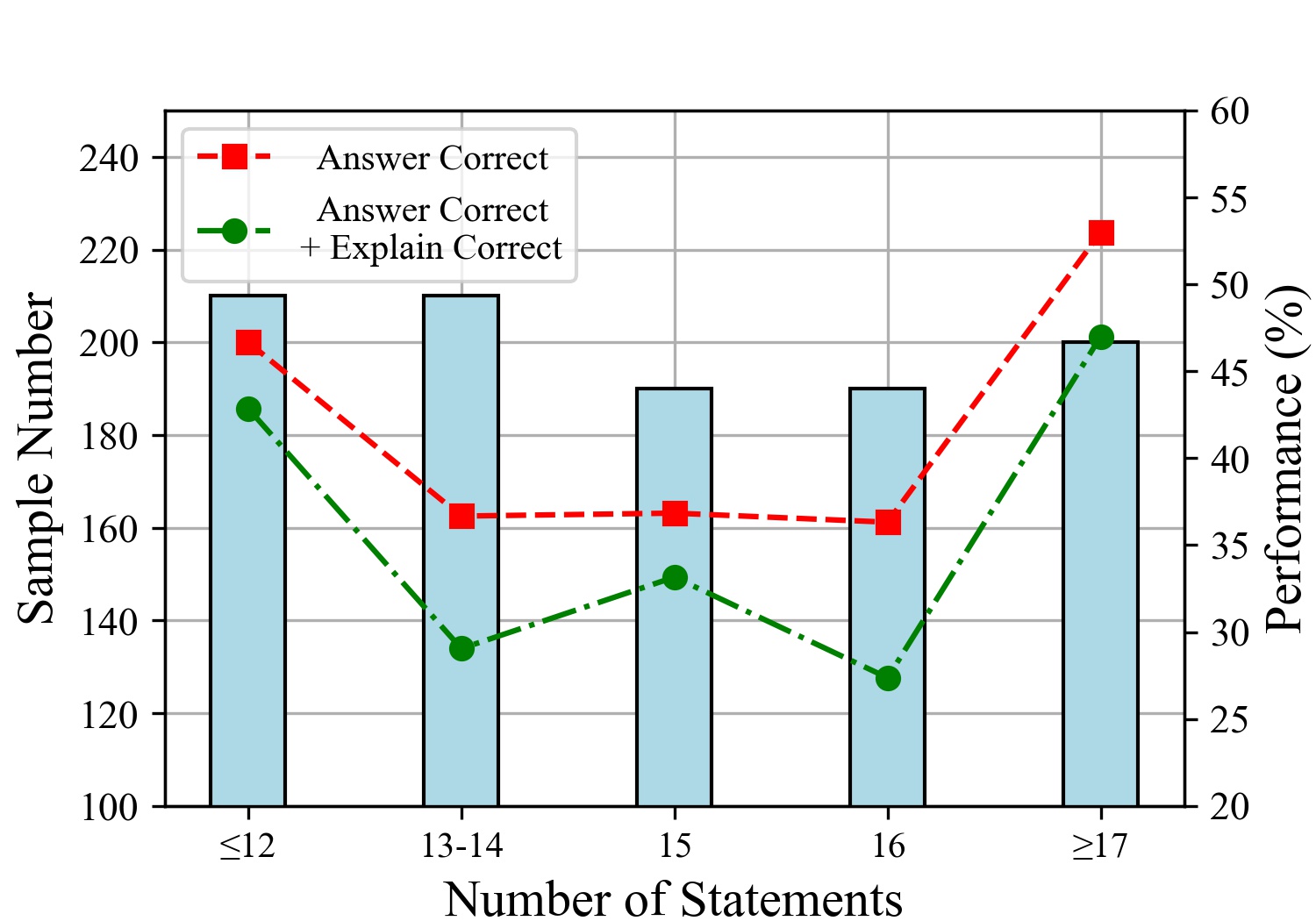}
            \vspace{-0.1cm}
		\subcaption{RuleTaker (De.).}
	\end{minipage}
        \\
	\begin{minipage}[t]{0.49\linewidth}
		\large
		\centering
		\includegraphics[scale=0.32]{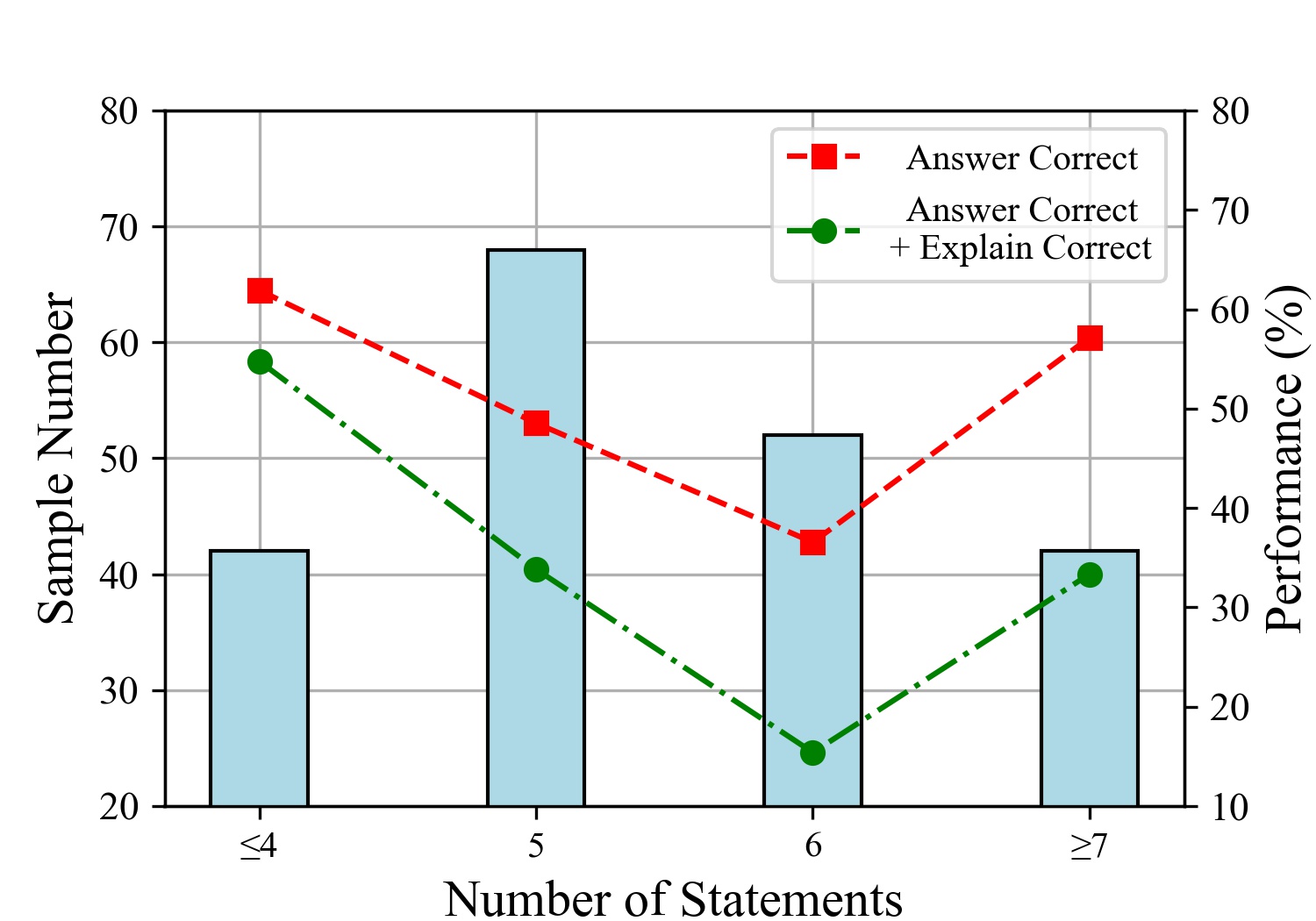}
            \vspace{-0.1cm}
		\subcaption{FOLIO (De.).}
	\end{minipage}
	\begin{minipage}[t]{0.49\linewidth}
		\large
		\centering
		\includegraphics[scale=0.32]{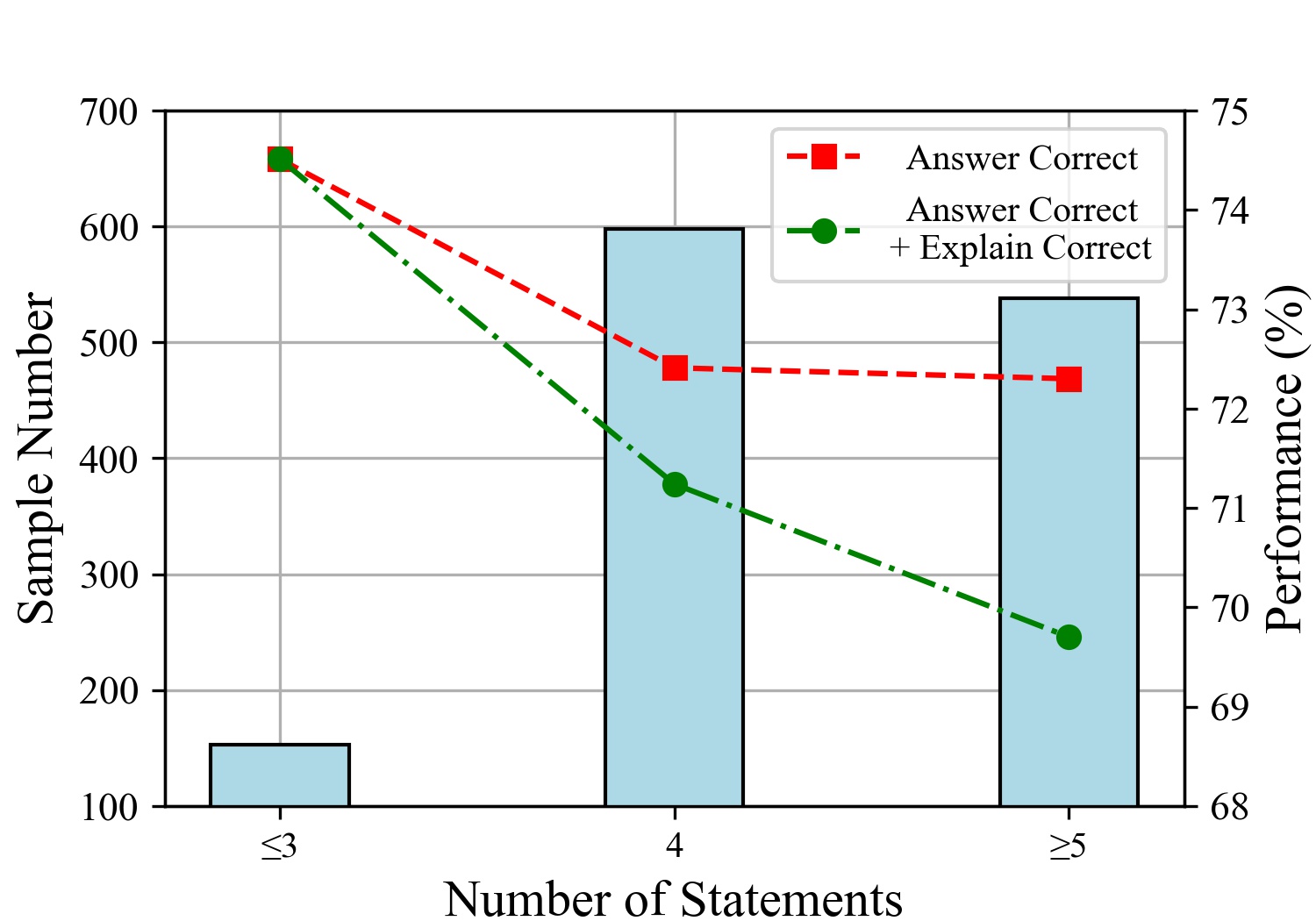}
            \vspace{-0.1cm}
		\subcaption{Leap (De.).}
	\end{minipage}
        \\
        \begin{minipage}[t]{0.49\linewidth}
		\large
		\centering
		\includegraphics[scale=0.32]{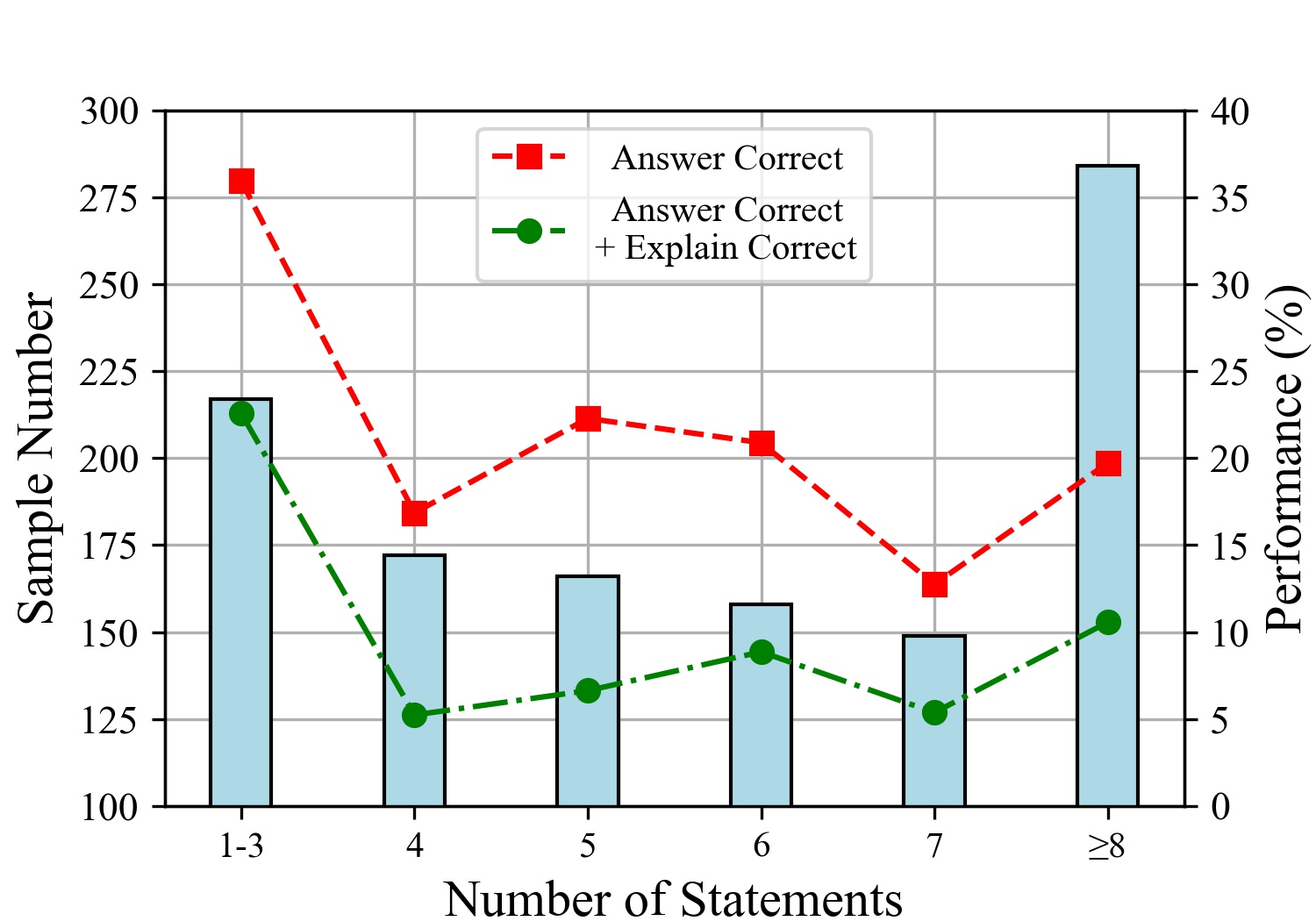}
            \vspace{-0.1cm}
		\subcaption{CLUTRR (In.).}
            \label{inductive_sta}
	\end{minipage}
	\begin{minipage}[t]{0.49\linewidth}
		\large
		\centering
		\includegraphics[scale=0.27]{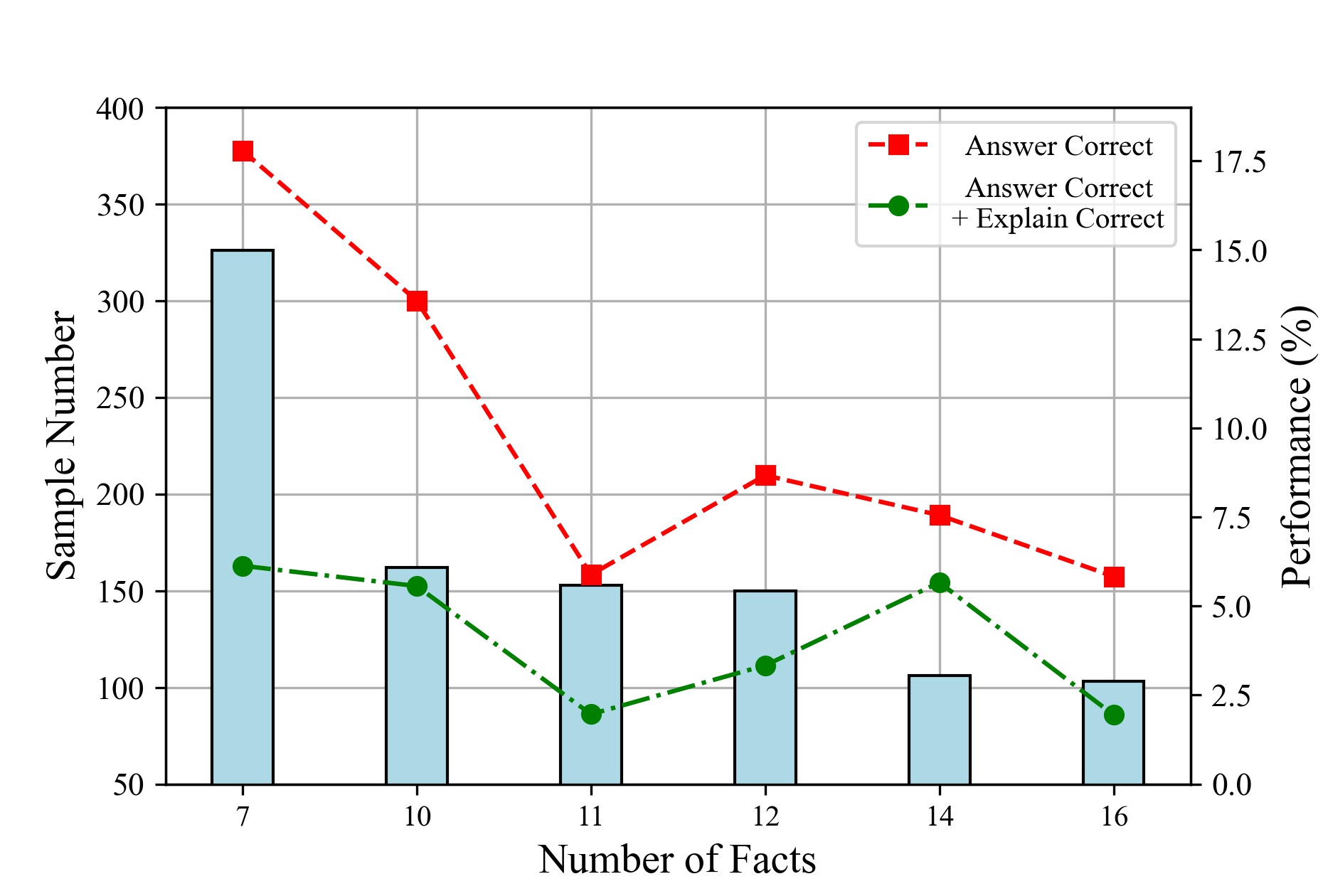}
            \vspace{-0.55cm}
		\subcaption{D*-Ab (Ab.).}
            \label{abductive_sta}
	\end{minipage}
        \caption{The LLM performances with different numbers of statements. {\color{black} The red line denotes \emph{Answer Correct} while the green line is \emph{Answer Correct+Explain Correct}.}}	
        \label{num_sta}
\end{figure}

{\color{black}The primary challenge evident from the results lies in the ability of LLMs to locate accurate evidence and perspectives for logical reasoning tasks.}
Due to the limited content of inputs, LLMs are also prone to generate hallucinatory facts to aid the reasoning process, which may affect the reliability of LLMs in real applications. In addition, LLMs abandon reasoning in a considerable number of cases. Such phenomena of laziness are worth noting, especially when we depend on LLMs to help reason in the downstream tasks. In the following section, we will provide a detailed analysis of specific LLM and specific reasoning settings. Considering the above-mentioned obvious logical flaws, we will especially focus on the detailed analysis of the activity in reasoning, the orientation selection of LLMs and model hallucination.

\subsection{Are LLMs Active Logical Reasoners?}
In this paper, we consider LLMs with fewer \emph{no reasoning} errors as more \emph{Active} logical reasoners. Fig.~\ref{heatmap_active} presents the weighted results for measuring active reasoning cases, where higher values indicate LLMs are more active in reasoning, while lower values represent lazier cases.
Among the three LLMs, BARD is the most active logical reasoner, excelling in deductive, inductive and abductive settings. ChatGPT, on the other hand, is deemed the lazier reasoner in deductive and inductive settings, while text-davinci-003 is lazier in abductive reasoning tasks.

Furthermore, we compare the performances of LLMs across different reasoning modes. In deductive reasoning tasks, LLMs exhibit more active reasoning, while in abductive settings, they tend to display lazier performances. Deductive tasks are in a forward reasoning mode, which is more natural for both generative LLMs and humans. This can inspire LLMs to generate effective reasoning chains. Conversely, abductive reasoning requires LLMs to provide explanations for the given inputs, which is in a backward reasoning mode. It is intuitive that LLMs may struggle to conduct reasoning in some cases.

\subsection{Are LLMs Oriented Logical Reasoners?}
At the outset of the reasoning process, it is crucial to identify the correct starting points and potential directions for reasoning. We consider LLMs with this capability as \emph{Oriented} logical reasoners and present evaluation results based on the error type of \emph{perspective mistake} in Fig.~\ref{heatmap_oriented}.
From the heatmap results, ChatGPT exhibits better oriented capability in deductive and inductive tasks compared to the other two LLMs. However, it frequently fails to identify the correct reasoning direction in abductive and mixed-form reasoning. Conversely, BARD performs well in identifying the right direction in abductive and mixed-form settings but struggles in deductive and inductive ones. Compared to the others, text-davinci-003 displays moderate performance in identifying reasoning perspectives.

In light of the preceding findings, ChatGPT is a lazier logical reasoner, but it excels at identifying the correct direction for reasoning. In other words, ChatGPT is prone to conduct confident reasoning. Conversely, text-davinci-003 and BARD are active in logical reasoning, but they tend to start from wrong directions, leading to reasoning mistakes.

{\color{black}The balance of active and oriented reasoning gives an insightful direction in exploring the knowledge and ability boundary of LLMs.
This implies that LLMs must understand their capabilities and limitations, avoiding tasks beyond their scope.
}

\subsection{Are LLMs Easy to Induce Hallucination in Logical Reasoning?}

\begin{figure*}[t]
    \begin{minipage}[t]{0.245\linewidth}
        \large
        \centering
        \includegraphics[scale=0.28]{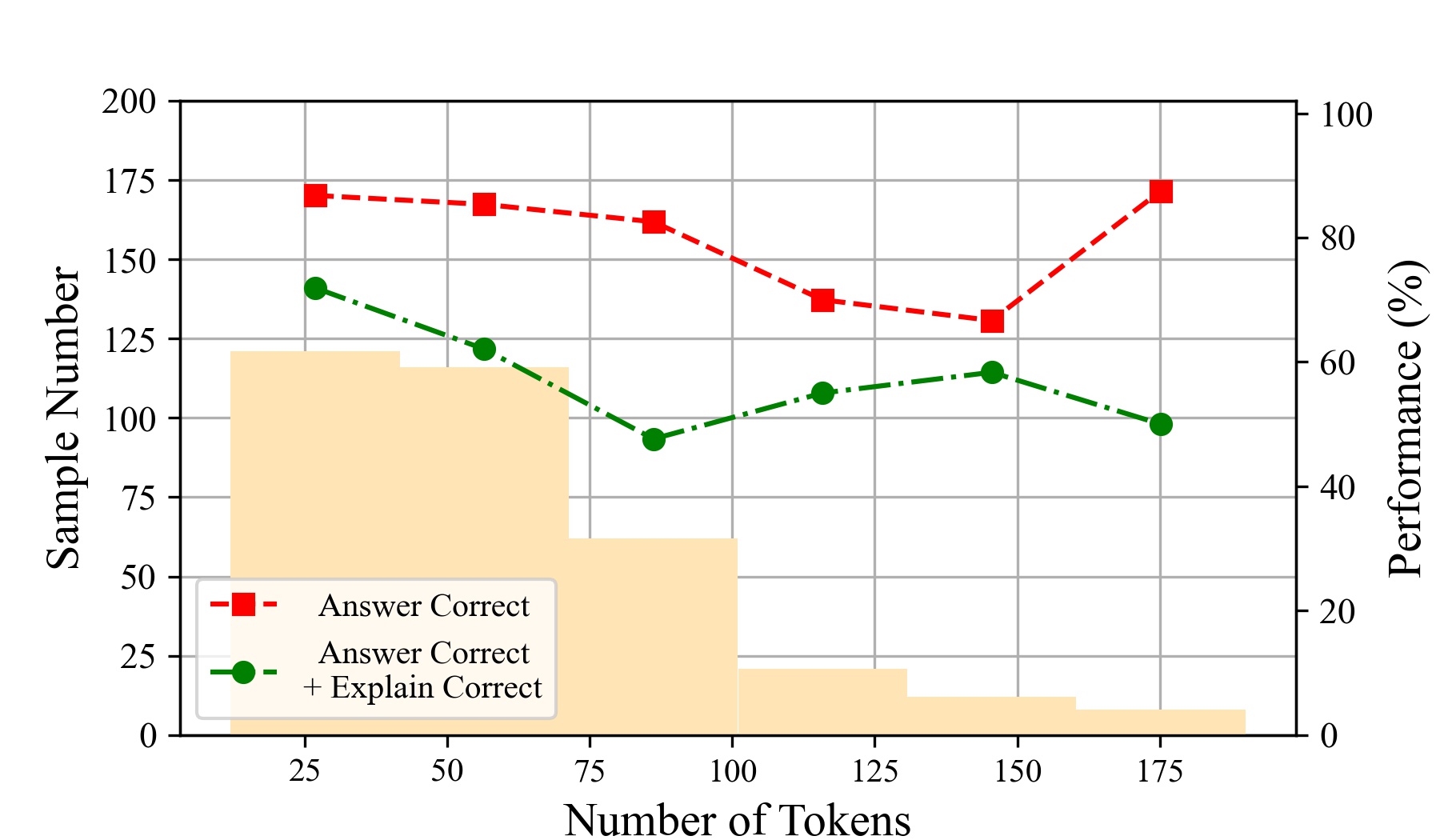}
        \vspace{-0.1cm}
        \subcaption{EntailmentBank (De.).}
    \end{minipage}
    \begin{minipage}[t]{0.245\linewidth}
        \large
        \centering
        \includegraphics[scale=0.28]{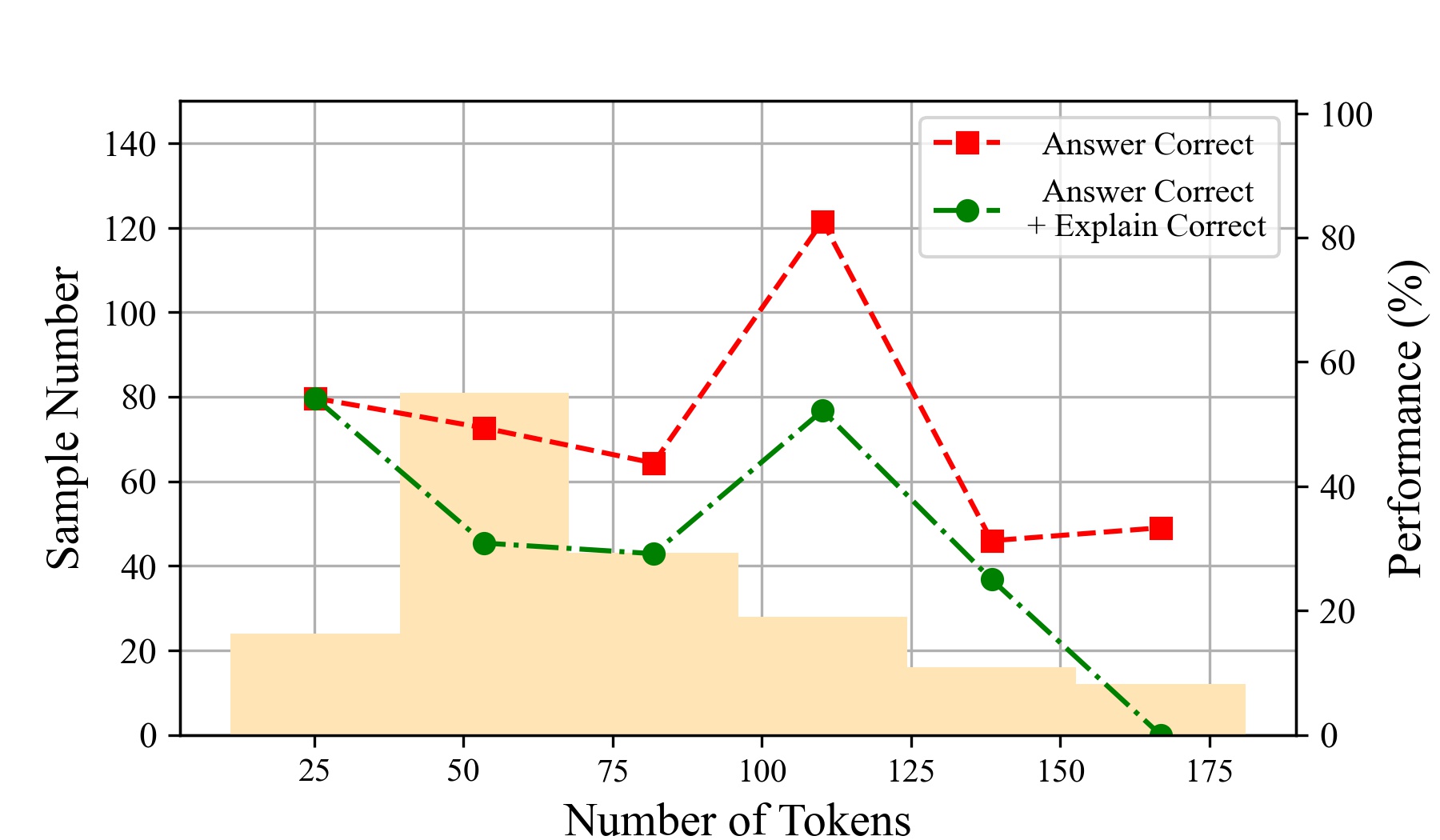}
        \vspace{-0.1cm}
        \subcaption{FOLIO (De.).}
    \end{minipage}
    \begin{minipage}[t]{0.245\linewidth}
        \large
        \centering
        \includegraphics[scale=0.28]{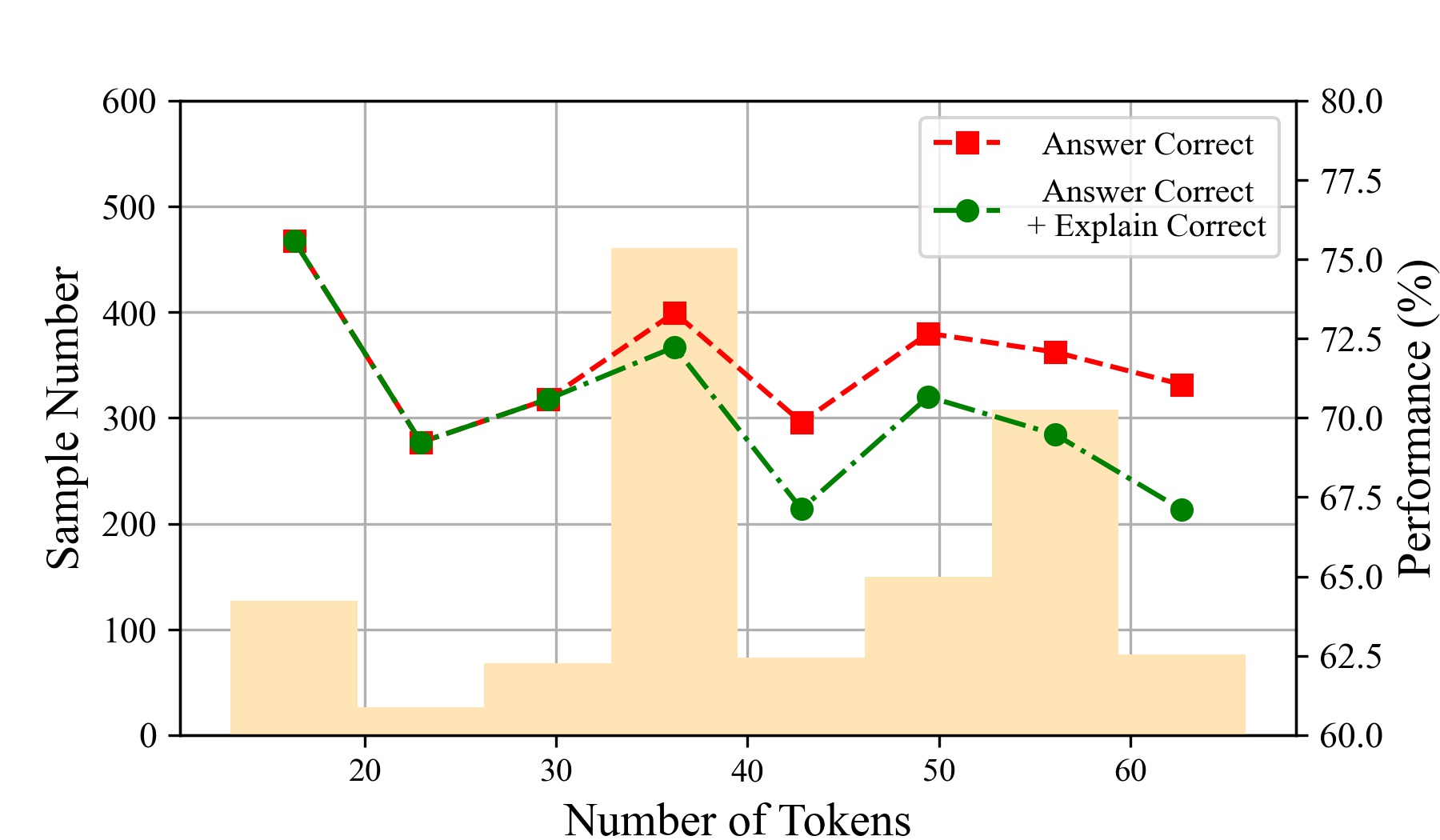}
        \vspace{-0.1cm}
        \subcaption{Leap-of-Thought (De.).}
    \end{minipage}
    \begin{minipage}[t]{0.245\linewidth}
        \large
        \centering
        \includegraphics[scale=0.28]{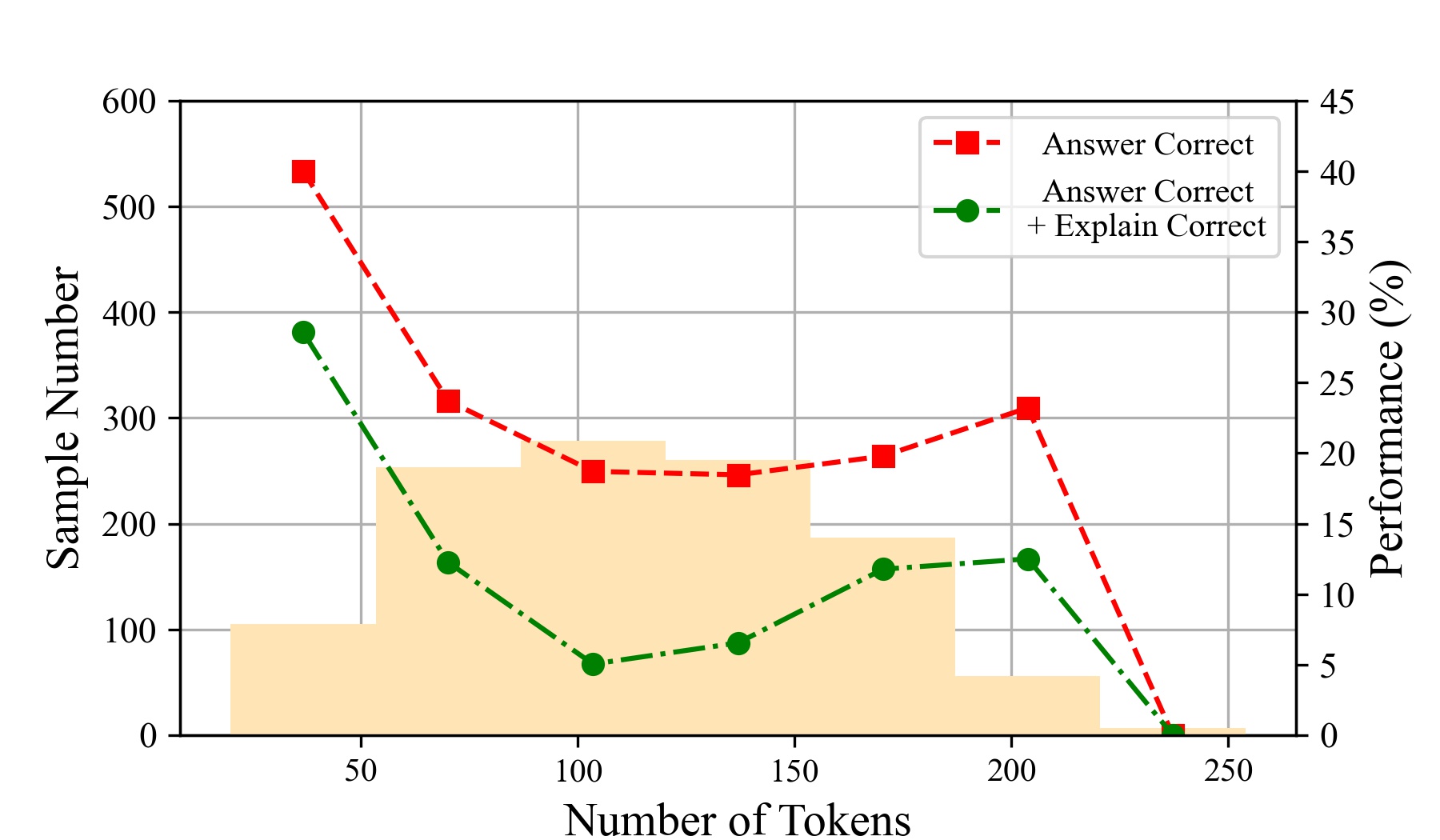}
        \vspace{-0.1cm}
        \subcaption{CLUTRR (In.).}
    \end{minipage}
    \\
    \begin{minipage}[t]{0.245\linewidth}
        \large
        \centering
        \includegraphics[scale=0.28]{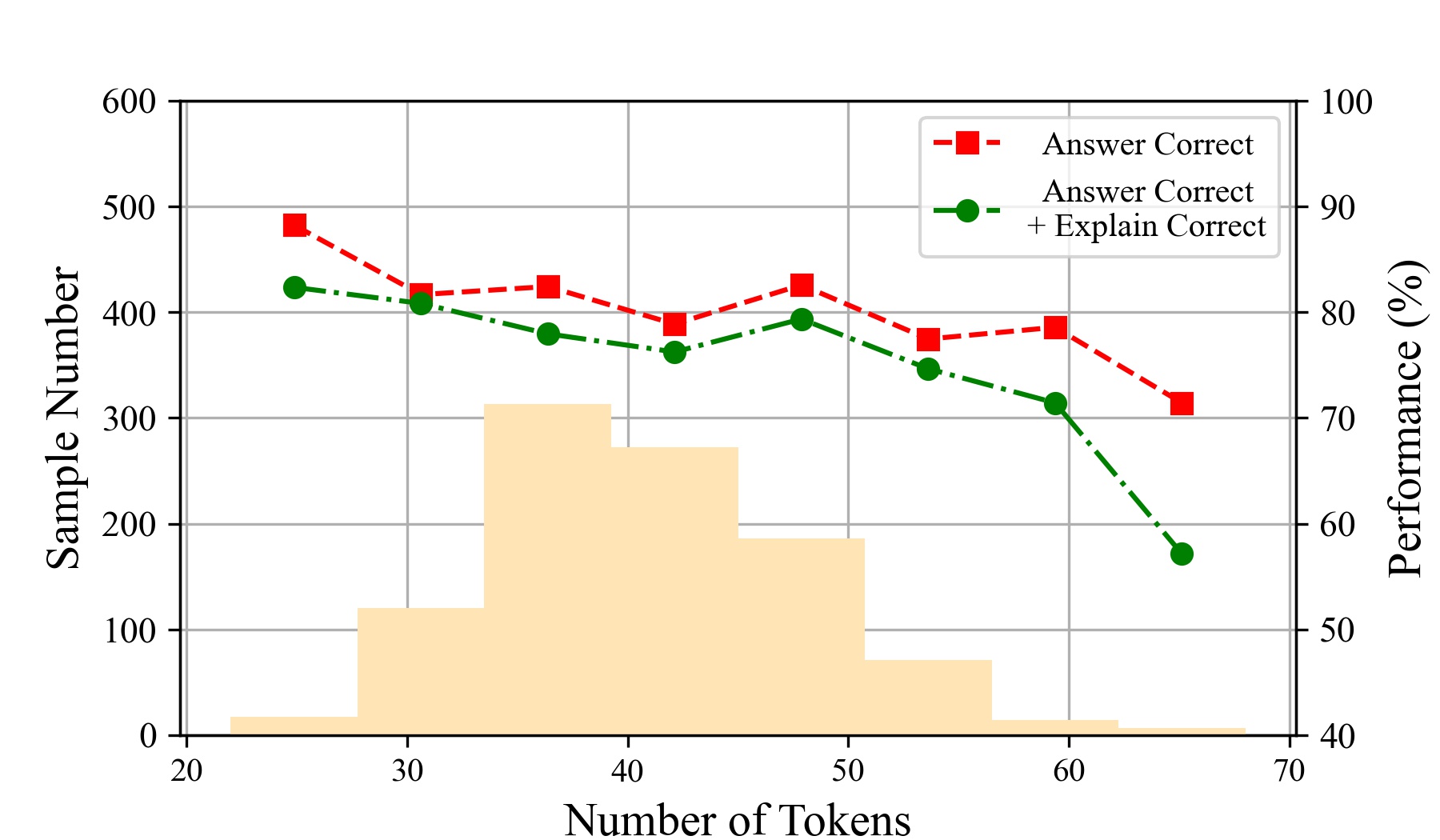}
        \vspace{-0.1cm}
        \subcaption{$\alpha$-NLI (Ab.).}
    \end{minipage}
    \begin{minipage}[t]{0.245\linewidth}
        \large
        \centering
        \includegraphics[scale=0.28]{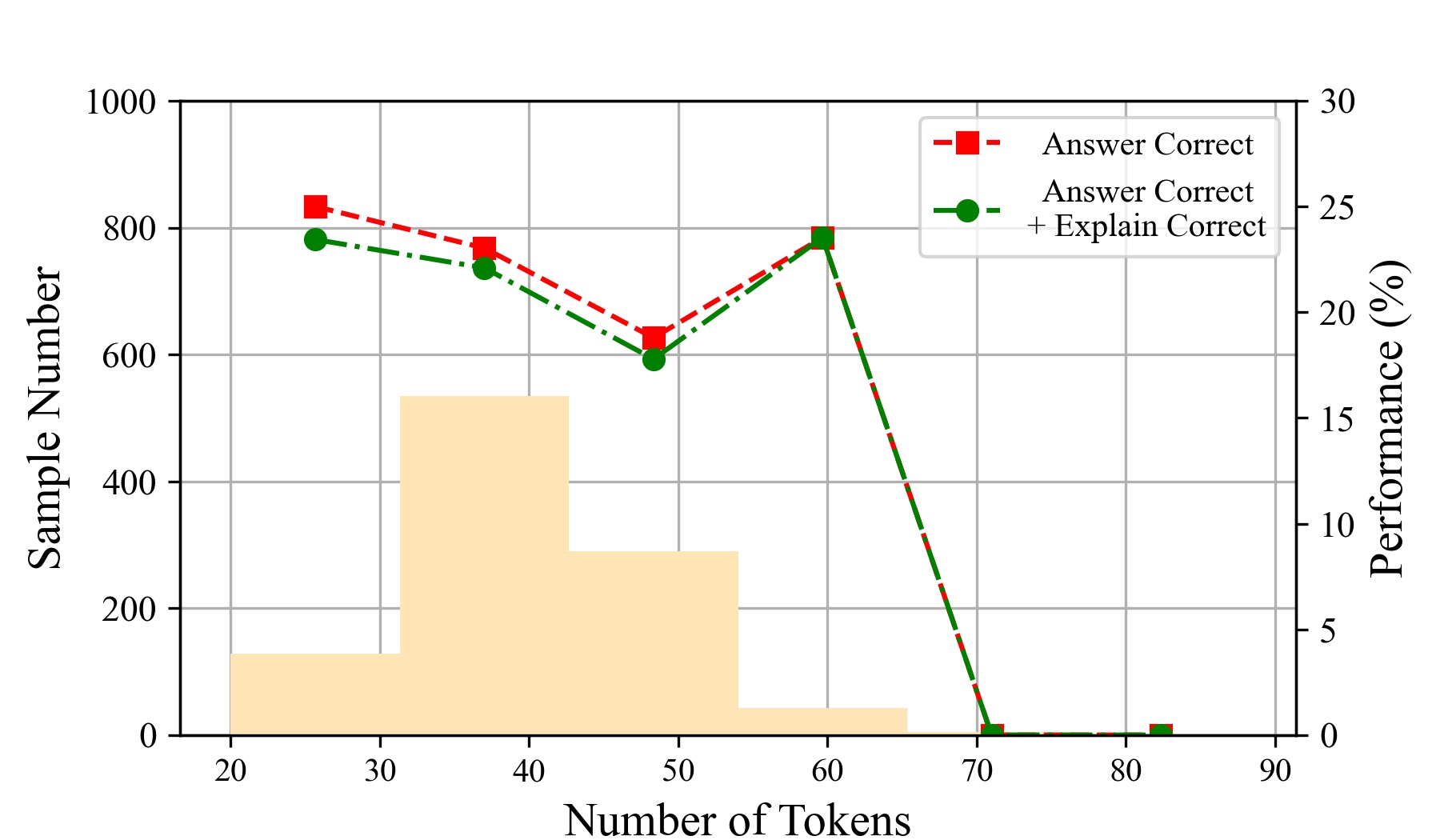}
        \vspace{-0.1cm}
        \subcaption{$\alpha$-NLG (Ab.).}
    \end{minipage}
    \begin{minipage}[t]{0.245\linewidth}
        \large
        \centering
        \includegraphics[scale=0.28]{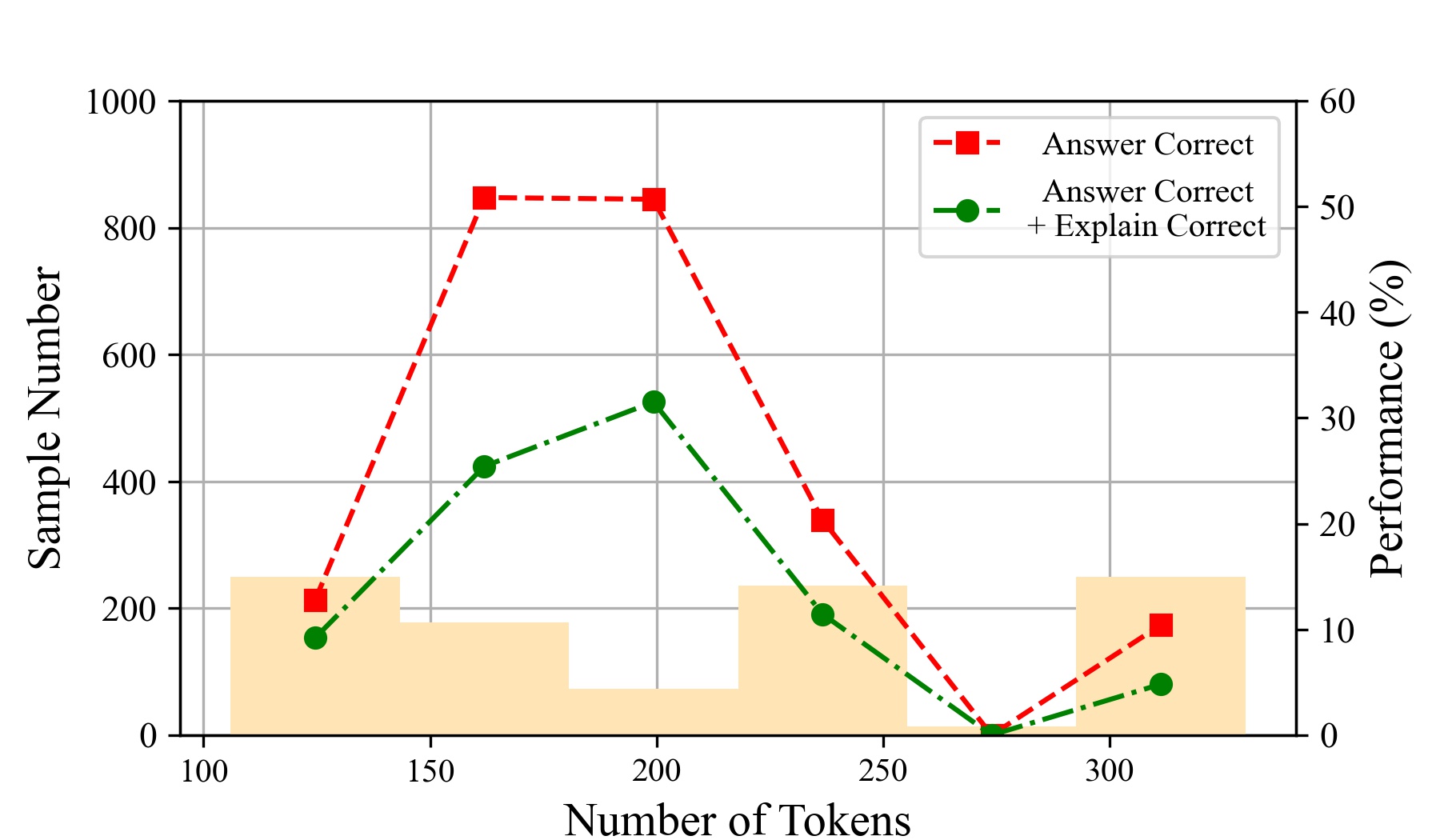}
        \vspace{-0.1cm}
        \subcaption{AbductiveRules (Ab.).}
    \end{minipage}
    \begin{minipage}[t]{0.245\linewidth}
        \large
        \centering
        \includegraphics[scale=0.28]{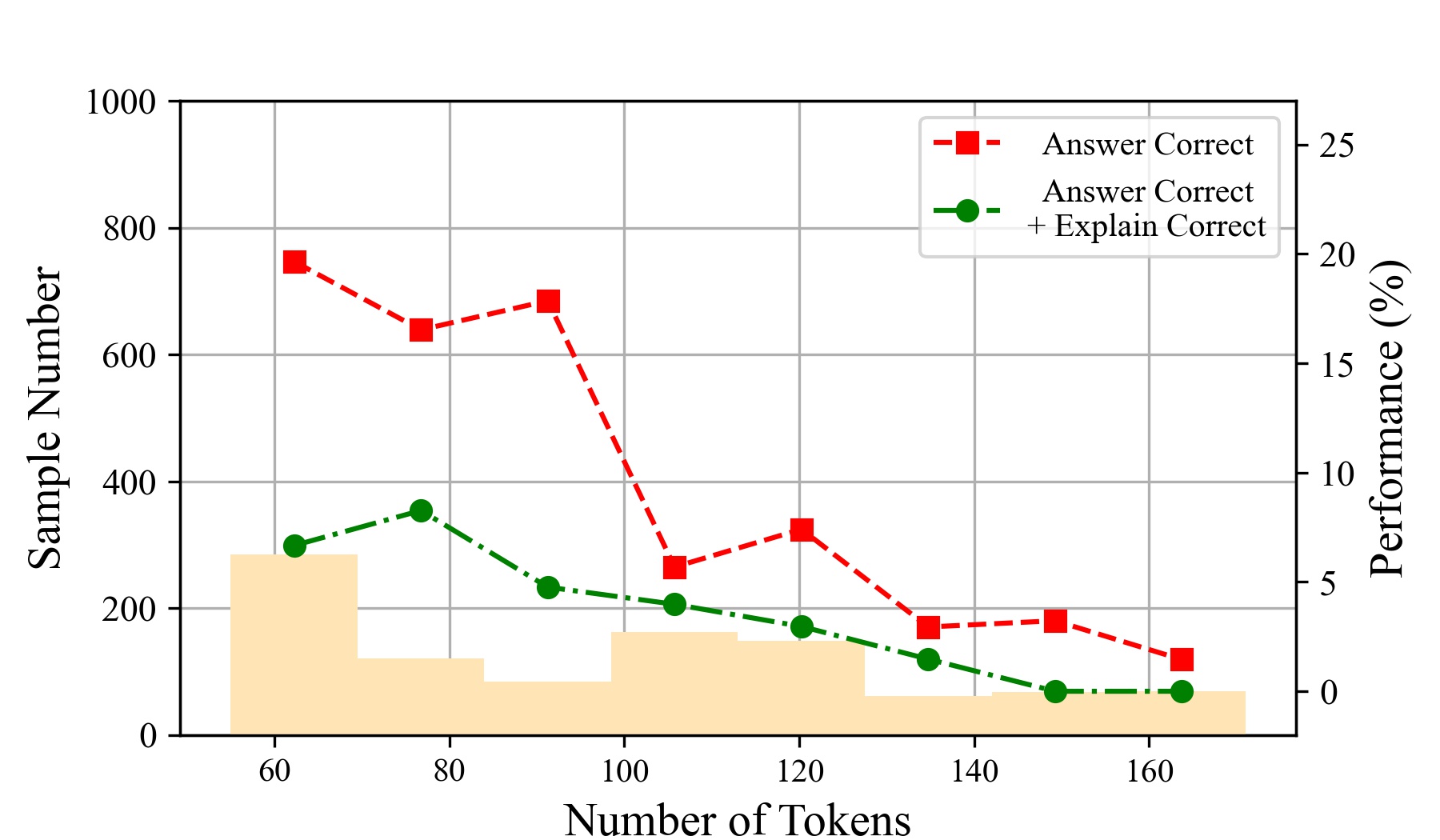}
        \vspace{-0.1cm}
        \subcaption{D*-Ab (Ab.).}
    \end{minipage}
    
    \begin{minipage}[t]{0.245\linewidth}
        \large
        \centering
        \includegraphics[scale=0.28]{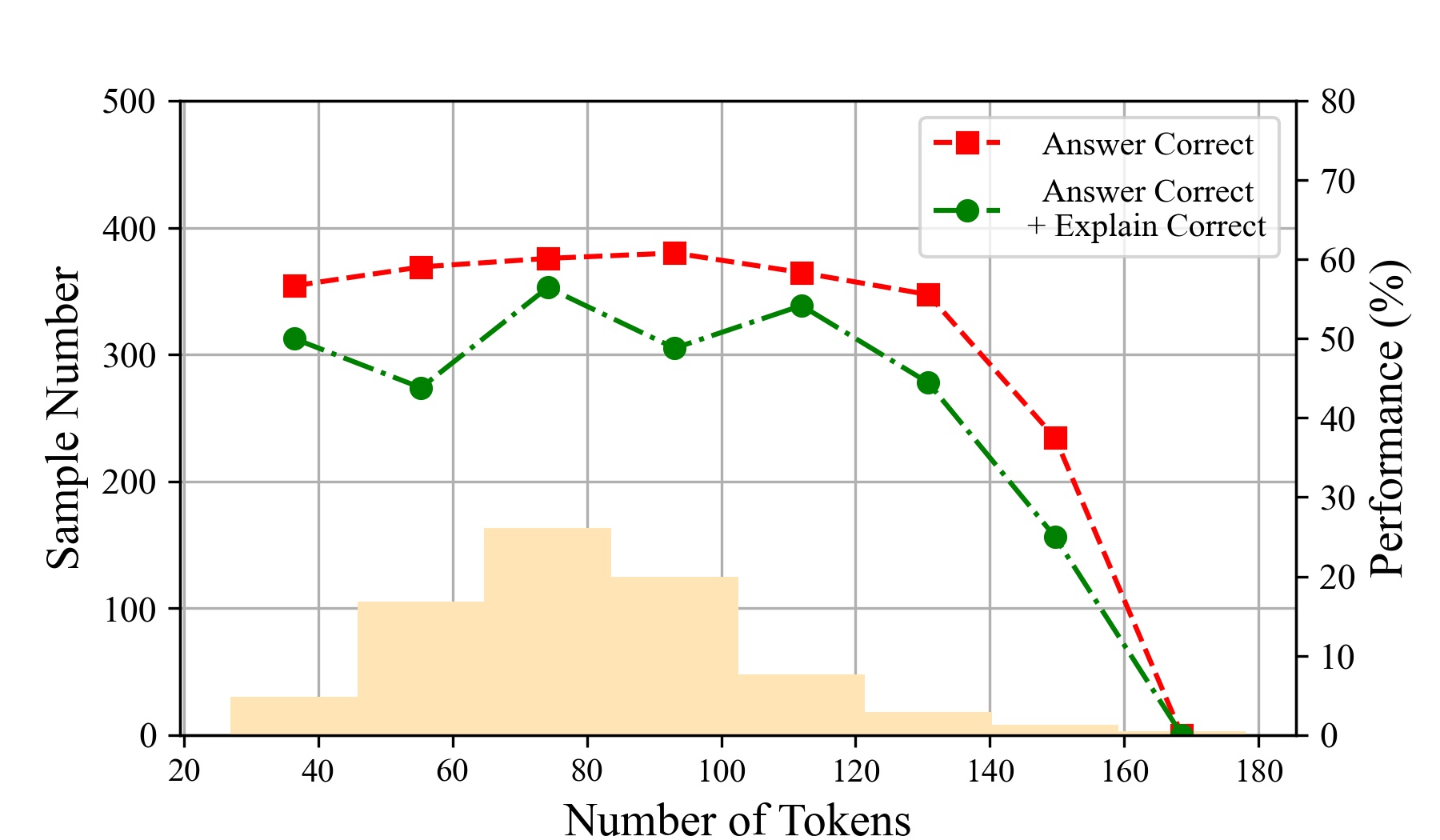}
        \vspace{-0.1cm}
        \subcaption{ReClor (Mix).}
    \end{minipage}
    \begin{minipage}[t]{0.245\linewidth}
        \large
        \centering
        \includegraphics[scale=0.28]{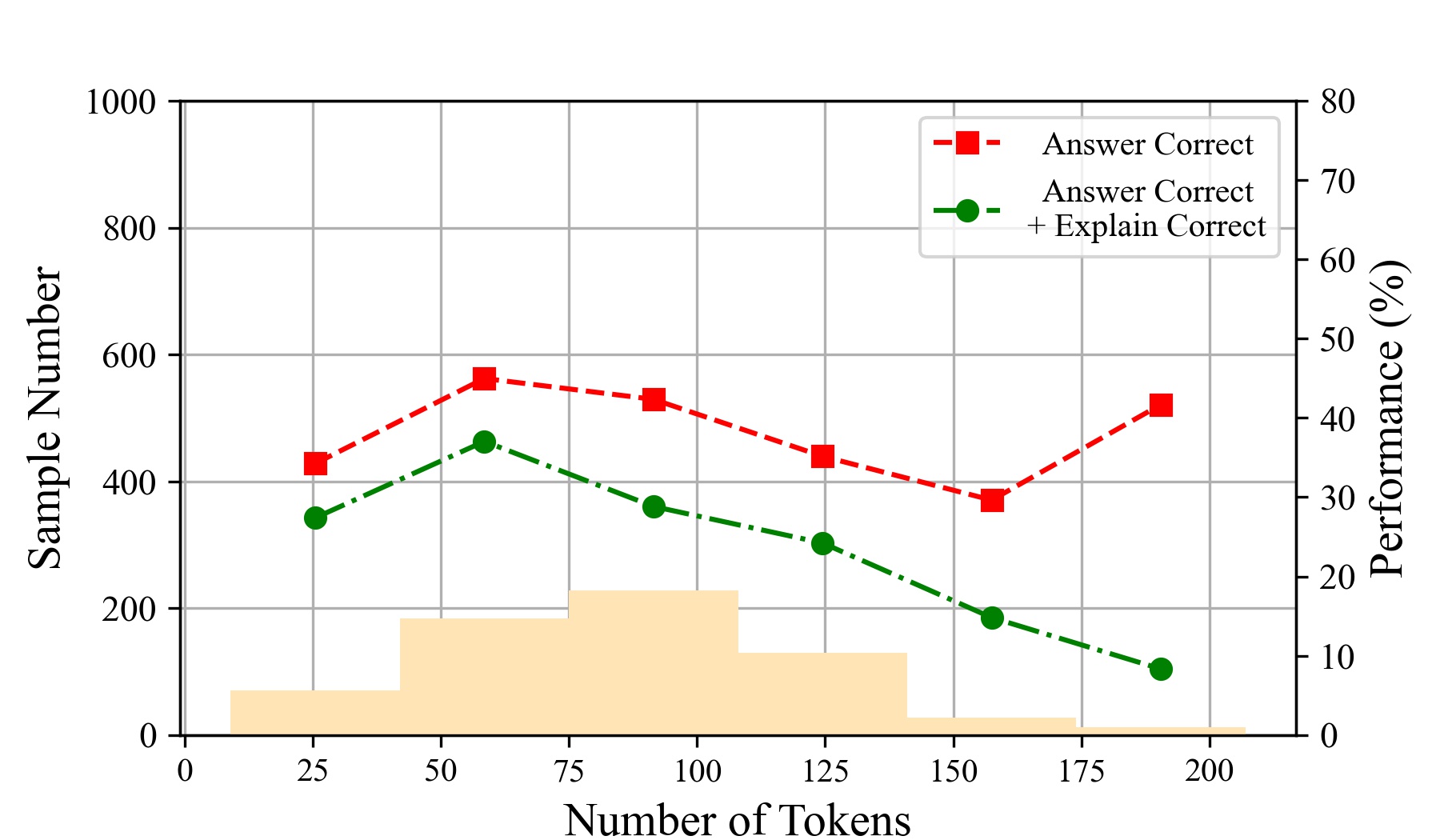}
        \vspace{-0.1cm}
        \subcaption{LogiQA (Mix).}
    \end{minipage}
    \begin{minipage}[t]{0.245\linewidth}
        \large
        \centering
        \includegraphics[scale=0.28]{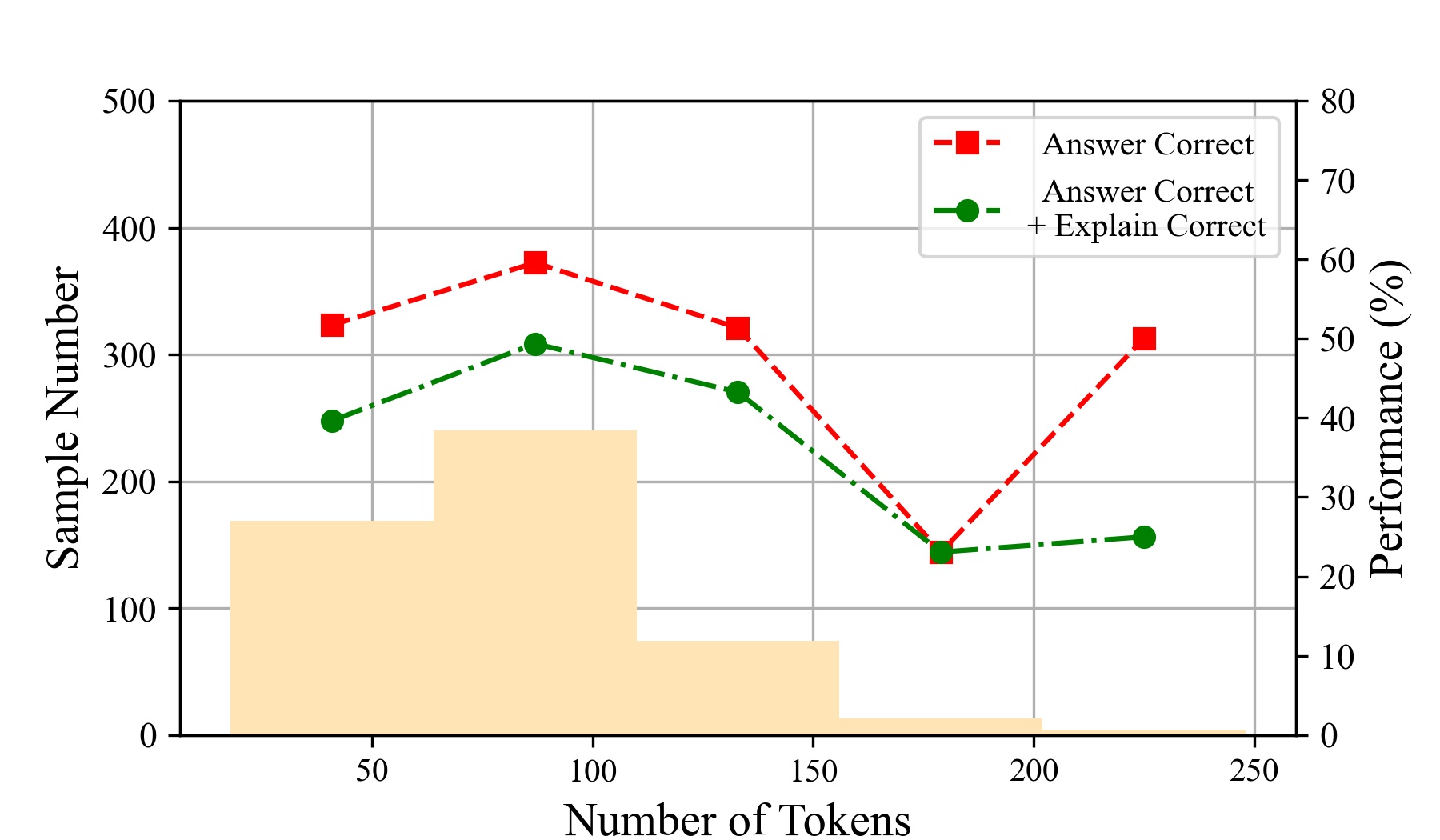}
        \vspace{-0.1cm}
        \subcaption{LogiQA2.0 (Mix).}
    \end{minipage}
    \begin{minipage}[t]{0.245\linewidth}
        \large
        \centering
        \includegraphics[scale=0.28]{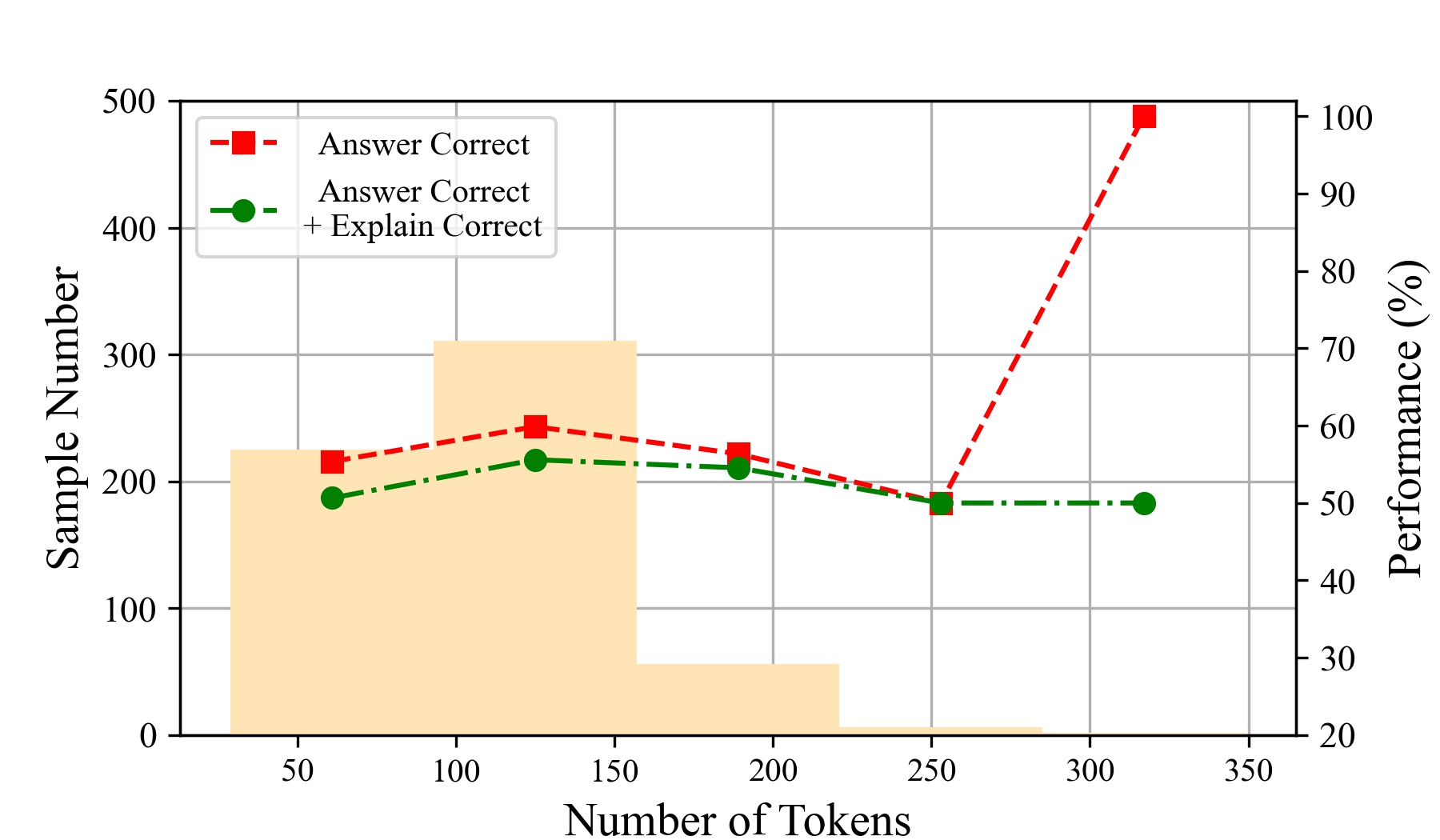}
        \vspace{-0.1cm}
        \subcaption{LogiQA2NLI (Mix).}
    \end{minipage}
    \caption{The performances of ChatGPT with different tokens on various datasets.{\color{black} The red line denotes \emph{Answer Correct} while the green line is \emph{Answer Correct+Explain Correct}.}}	
    \label{num_tokens}
\end{figure*}

\begin{figure*}[t]
	\begin{minipage}[t]{0.33\linewidth}
		\large
		\centering
		\includegraphics[scale=0.42]{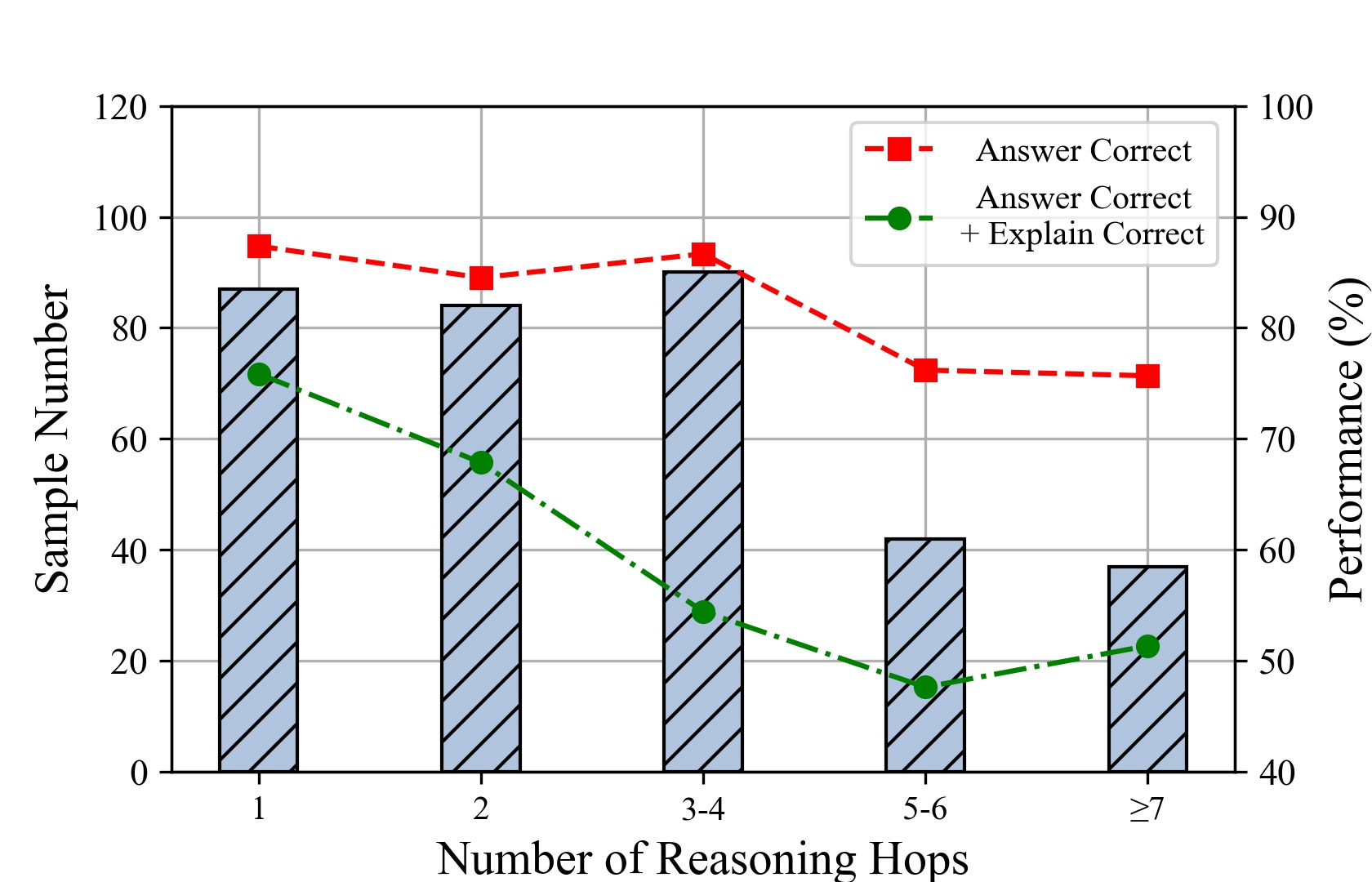}
		\vspace{-0.1cm}
		\subcaption{Deductive (EntailmentBank).}
		\label{de}
	\end{minipage}
	\begin{minipage}[t]{0.33\linewidth}
		\large
		\centering
		\includegraphics[scale=0.42]{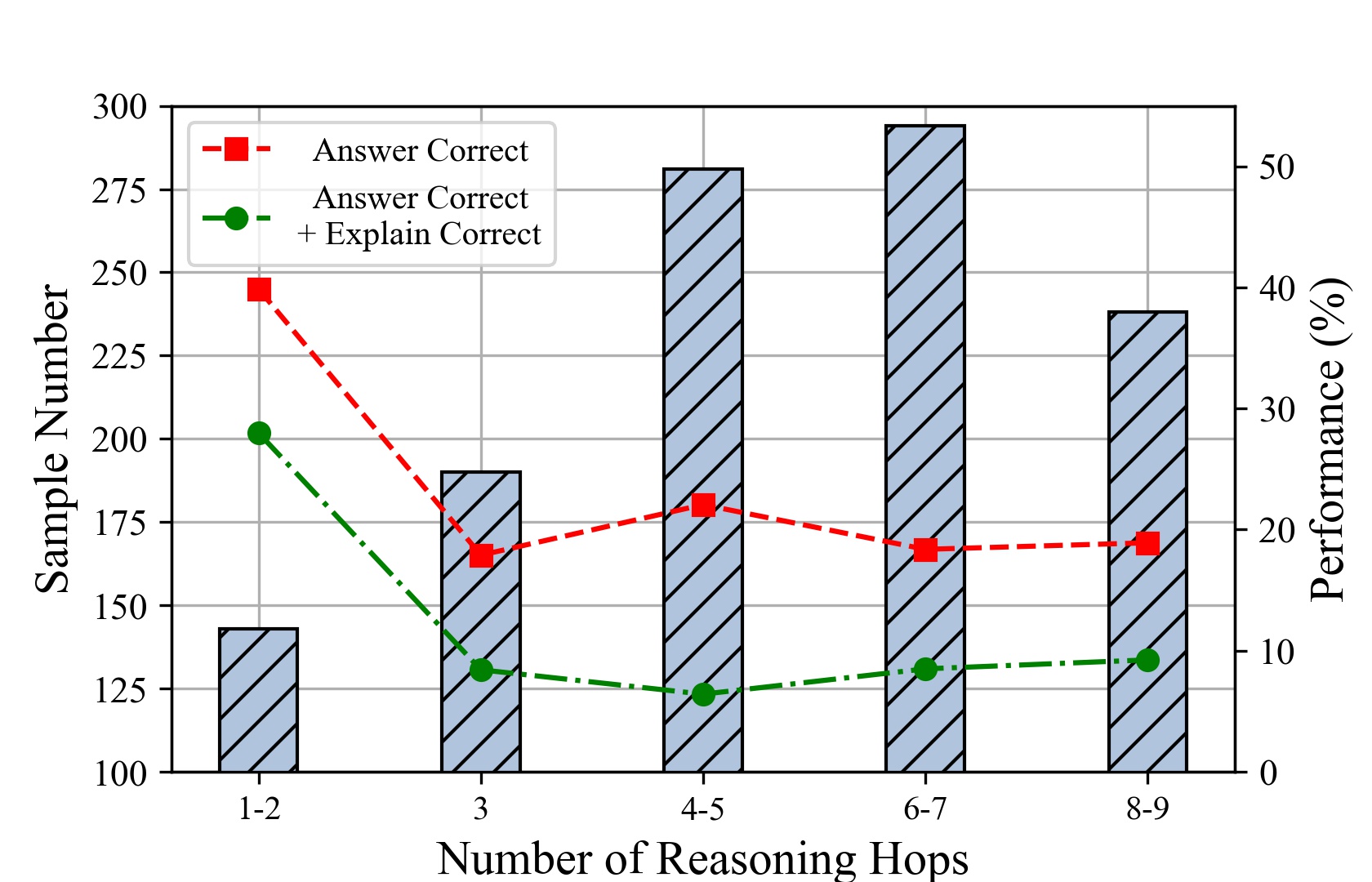}
		\vspace{-0.1cm}
		\subcaption{Inductive (CLUTRR).}
		\label{in}
	\end{minipage}
	\begin{minipage}[t]{0.33\linewidth}
		\large
		\centering
		\includegraphics[scale=0.42]{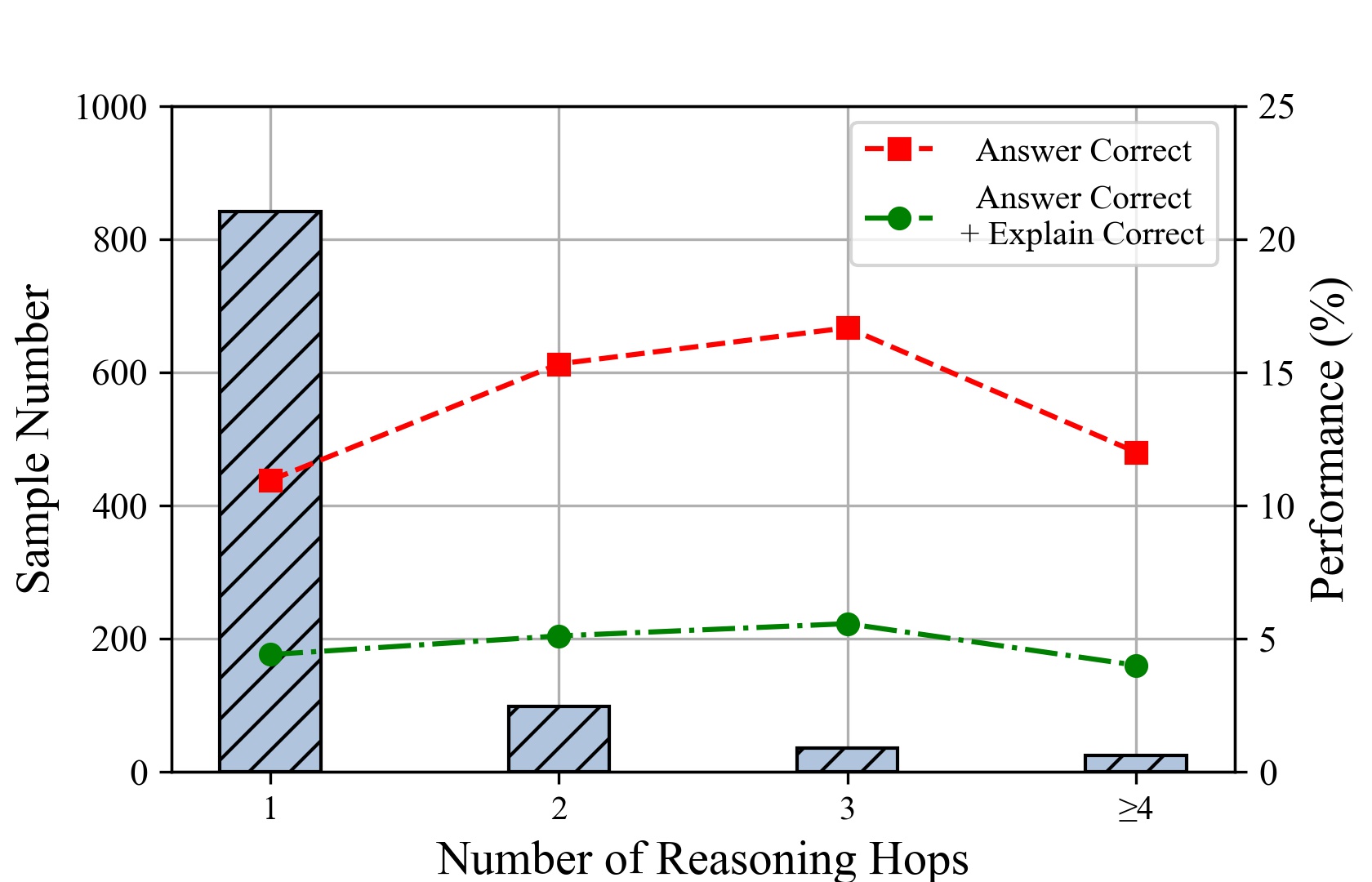}
		\vspace{-0.1cm}
		\subcaption{Abductive (D*-Ab).}
		\label{in}
	\end{minipage}
	\caption{The performances of ChatGPT under different number of hops. Comparison of Deductive, Inductive and Abductive reasoning settings.{\color{black} The red line denotes \emph{Answer Correct} while the green line is \emph{Answer Correct+Explain Correct}.}}
	\label{num_hop}
\end{figure*}

In the typical definition, hallucination refers to generated content that contradicts commonsense or current facts. To align with logical reasoning tasks, we expand the definition to include cases where facts are employed that contradict the context or are not verified by the context. Fig.~\ref{heatmap_hallu} displays the weighted performances of cases with no hallucinations. The darker color indicates better performance in avoiding hallucinations.
Based on the results, ChatGPT exhibits strong and consistent competitiveness, ranking first in deductive, inductive, and mixed-form settings. It also ranks second in abductive reasoning tasks, albeit with slight disadvantages. Conversely, BARD displays poor performance in avoiding hallucinations and maintaining clarity during reasoning. Across all four reasoning settings, BARD ranks last with significant gaps.

On average, LLMs induce hallucination in 27.46\% of the failure cases. 
{\color{black}
LLM hallucinations are more prevalent in deductive reasoning tasks among the settings of deductive, inductive, and abductive reasoning, whereas LLMs may exhibit clearer judgment in the inductive setting.
}

\subsection{How Does the Number of Statements Affect the LLMs' Performances?}

\begin{table*}[t]
	\centering
	\small
	\caption{Statistics and evaluation results on NeuLR. \emph{Num.} represents the number of samples in the dataset. \emph{\#Hop} represents the hop number of samples in the dataset. \emph{COT} represents the chain-of-thought strategy under the 1-shot setting.}
	\begin{tabular}{c|c|c|ccc|ccc|ccc}
		\toprule
            \multicolumn{1}{c|}{\multirow{2}{*}{\textbf{Dataset}}} &\multicolumn{1}{c|}{\multirow{2}{*}{\textbf{Num.}}} &\multicolumn{1}{c|}{\multirow{2}{*}{\textbf{\#Hop}}} & \multicolumn{3}{c|}{\textbf{text-davinci-003}} & \multicolumn{3}{c|}{\textbf{ChatGPT}} & \multicolumn{3}{c}{\textbf{BARD}}\\
            \multicolumn{1}{c|}{} & & & \textbf{0-shot} & \textbf{1-shot} & \textbf{COT} & \textbf{0-shot} & \textbf{1-shot} & \textbf{COT} & \textbf{0-shot} & \textbf{1-shot} & \textbf{COT}\\
            \midrule 
            \textbf{NeuLR} &3,000 &1$\sim$5 &50.93 &59.17 &67.90 &37.27 &48.13 &48.00 &63.67 &65.07 &66.00 \\
            \midrule 
            - Deductive &1,000 &2 &59.00 &69.40 &86.10 &85.20 &69.10 &68.30 &87.40 &93.10 &91.90 \\
            - Inductive &1,000 &3 &86.90 &89.60 &95.60 &15.10 &68.60 &69.60 &96.00 &92.60 &96.30\\
            - Abductive &1,000 &1$\sim$5 &6.90 &18.50 &22.00 &11.50 &6.70 &6.10 &7.60 &9.50 &9.80 \\
		\bottomrule
	\end{tabular}
	\vspace{-0.2cm}
	\label{neulr}
\end{table*}

In the following, we explore some of the key factors to affect the reasoning performances of LLMs. Since the length of the input context can be different for the datasets, we report the model performances with the number of statements in Fig.~\ref{num_sta}. We take ChatGPT for analysis and choose six datasets with specific counts of statements for illustration, covering the three reasoning manners. The first four subfigures are related to the deductive reasoning manner, i.e., EntailmentBank, RuleTaker, FOLIO, and Leap-of-Thought. Fig.~\ref{inductive_sta} is CLUTRR in the inductive setting and Fig.~\ref{abductive_sta} is D\*-Ab in the abductive setting. The horizontal axis denotes the number of statements. The left vertical axis denotes the number of samples for different numbers of statements. And the right vertical axis represents the performances with different numbers of statements.

From the overall results, LLMs can keep the correctness of both the answer and explanations with fewer input statements. With the statement number increasing, the performances drop a lot and LLMs struggle to give the correct explanations. Interestingly, five (out of six) datasets witness performance gains when the number of statements reaches certain values. For example, in the RuleTaker dataset, the best performances are achieved when the number of statements is larger than 17. And when the number is between 13 to 16, ChatGPT is capable of keeping stable performances. We argue that the larger number of statements can provide richer information and sometimes can help control the reasoning direction of LLMs.

Furthermore, we investigate the impact of the number of tokens in the context.
Fig.~\ref{num_tokens} illustrates the ChatGPT performances on twelve datasets. Results vary a lot across different datasets and different reasoning settings. In general, as the number of tokens increases, the performances of ChatGPT tend to decline. Detailedly, ChatGPT performs stably in deductive settings. Especially for EntailmentBank and Leap-of-Thought datasets, ChatGPT maintains relatively consistent accuracy with token numbers increasing. Considering that deductive setting is a more common form of reasoning in reality, LLMs are well-trained on it to tackle the various lengths of inputs.

\subsection{How Does the Number of Reasoning Hops Affect the LLMs' Performances?}

Also, it is interesting to explore the influences of reasoning hops for LLMs. Among the selected dataset, three of them offer the number of hops for each sample, which are EntailmentBank in deductive reasoning, CLUTRR in inductive reasoning and D*-Ab in abductive reasoning. Fig.~\ref{num_hop} presents the performance of ChatGPT with different hop numbers (Results of other LLMs are listed in the Appendix). Alongside the simple accuracy results, we also report the rigorous reasoning cases where both the answer and explanations are correct. 

In the deductive setting, with the number of hops increasing, the performances witness obvious drops, particularly influencing the rigor of the LLM reasoning. It illustrates that reasoning hops have great effects on deductive reasoning. In inductive reasoning, when the hop number is greater than two, the performance of ChatGPT decreases sharply. When the number ranges from three to nine, the performance of ChatGPT keeps stable at a relatively low level. Combined with the weak performance of ChatGPT in inductive reasoning tasks, it demonstrates that ChatGPT can only work on simple induction, and it obviously struggles in cases when more hops are needed. In the abductive reasoning setting, the majority of the test samples only need one-hop reasoning. When the hop number increases, the performances of ChatGPT witness slight improvements. It shows that ChatGPT may have the potential capability of multi-hop reasoning in the abductive setting. 

{\color{black}
\subsection{Does Chain-of-Thought or Program-of-Thought Help Logical Reasoning?}
As a supplement to the original results, we add the comparison between chain-of-thought and direct prompting in Table~\ref{main_new}.
It is observed that chain-of-thought does not always lead to evident improvements
Especially under abductive reasoning, COT tends to have adverse effects in the majority of instances.
Given the distinct disparities between logical reasoning and fact-based reasoning, the model is required to engage in abstract reasoning.
But COT actually works more for superficial reasoning over facts.

Recently, the program-of-thought (POT) prompting strategy~\cite{chen2022program,gao2023pal} stands out as another selection to improve the model performance. Since POT relies heavily on the accessibility of external solvers, we merely include several logical reasoning datasets for evaluation, i.e., ProofWriter and RuleTaker. In the implementation, LLMs are prompted to generate \emph{First-order Logic}, which can be executed by the \emph{Pyke} solver to obtain the final answer~\cite{xu-etal-2024-symbol,pan2023logic}. We include the experimental results in Appendix.

It can be concluded that \textbf{more powerful LLMs can derive greater advantages from the PoT.} 
It is observed that open-source LLMs (i.e., LLaMA3.1-Chat and Mistral-Instruct-v0.3) benefit more from the chain-of-thought strategy, but fail in the program-of-thought method. Conversely, proprietary powerful LLMs can derive more benefits from POT than from COT. It demonstrates that POT requires the basic symbolic generation capability, which is challenging for most of the current trending LLMs. The symbol-centric modeling for LLMs is also a promising direction.

}

\section{Neutral-content Logical Reasoning}

Considering the current benchmarks may not provide neutral content for fair evaluation, we propose the new dataset NeuLR to benchmark the neutral-content logical reasoning tasks. In column 1$\sim$3 of TABLE \ref{neulr}, we provide the statistics of NeuLR. It contains 3k samples in total, with 1k for deductive reasoning, 1k for inductive reasoning and 1k for abductive reasoning. Limited by space, we provide the details of the construction of NeuLR in the Appendix.

To evaluate the performances of LLMs on NeuLR, we conduct the experiments shown in TABLE \ref{neulr}. Especially, we provide three different test settings, i.e., zero-shot, one-shot and chain-of-thought~\cite{DBLP:conf/nips/Wei0SBIXCLZ22} settings. Details of the prompt forms are included in the Appendix.

Firstly, from the results, few-shot prompting and chain-of-thought prompting can both boost the performances of LLMs in most cases. Overall, chain-of-thought helps most to the model accuracy. Especially for text-davinci-003, it witnesses consistent gains with the aid of few-shot prompting and chain-of-thought prompting strategies.

Secondly, among the zero-shot results of three LLMs, BARD achieves the best performances on NeuLR while ChatGPT ranks last. The differences of zero-shot settings are significant. However, with the help of few-shot and chain-of-thought prompting strategies, text-davinci-003 large narrows the gaps with BARD, and it surpasses BARD with chain-of-thought strategy. Overall, the performances of LLMs on NeuLR still have great room for improvement.

Thirdly, from the perspective of different reasoning settings, there exist huge differences in results compared with the previous findings. 
Generally, the LLMs' performances in inductive reasoning are better than deductive reasoning and abductive reasoning. Especially, LLMs present obvious weakness in the abductive setting. While from fifteen classical datasets, the performances among the reasoning settings are sorted as \emph{deductive} $\textgreater$ \emph{abductive} $\textgreater$ \emph{inductive}. Such findings can also motivate future studies.
Content neutrality can enhance strategies for boosting inductive reasoning performance.

\begin{figure*}[t]
	\begin{minipage}[t]{0.24\linewidth}
		\large
		\centering
		\includegraphics[scale=0.42]{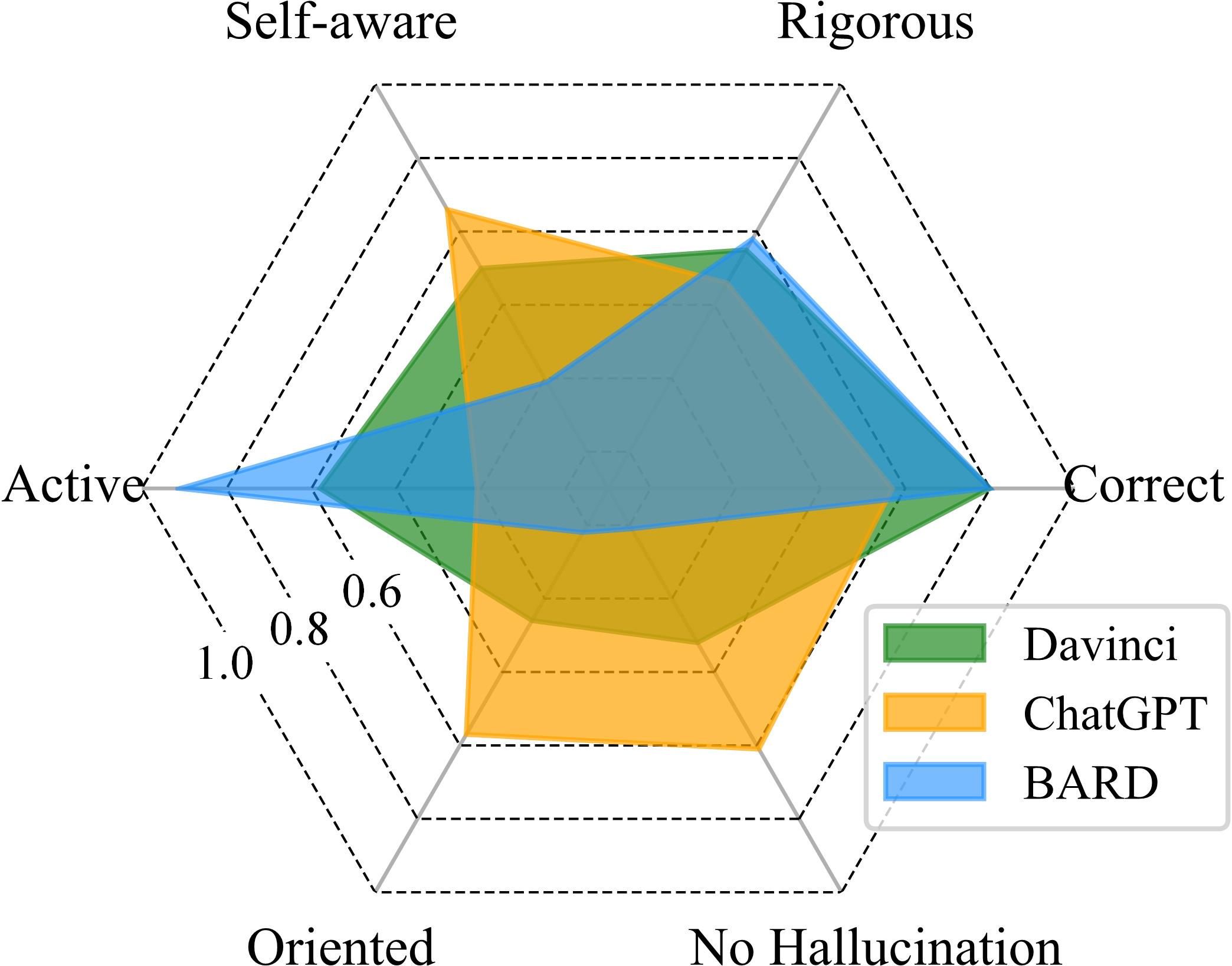}
		\vspace{-0.1cm}
		\subcaption{Deductive.}
		\label{res_de}
	\end{minipage}
	\begin{minipage}[t]{0.24\linewidth}
		\large
		\centering
		\includegraphics[scale=0.42]{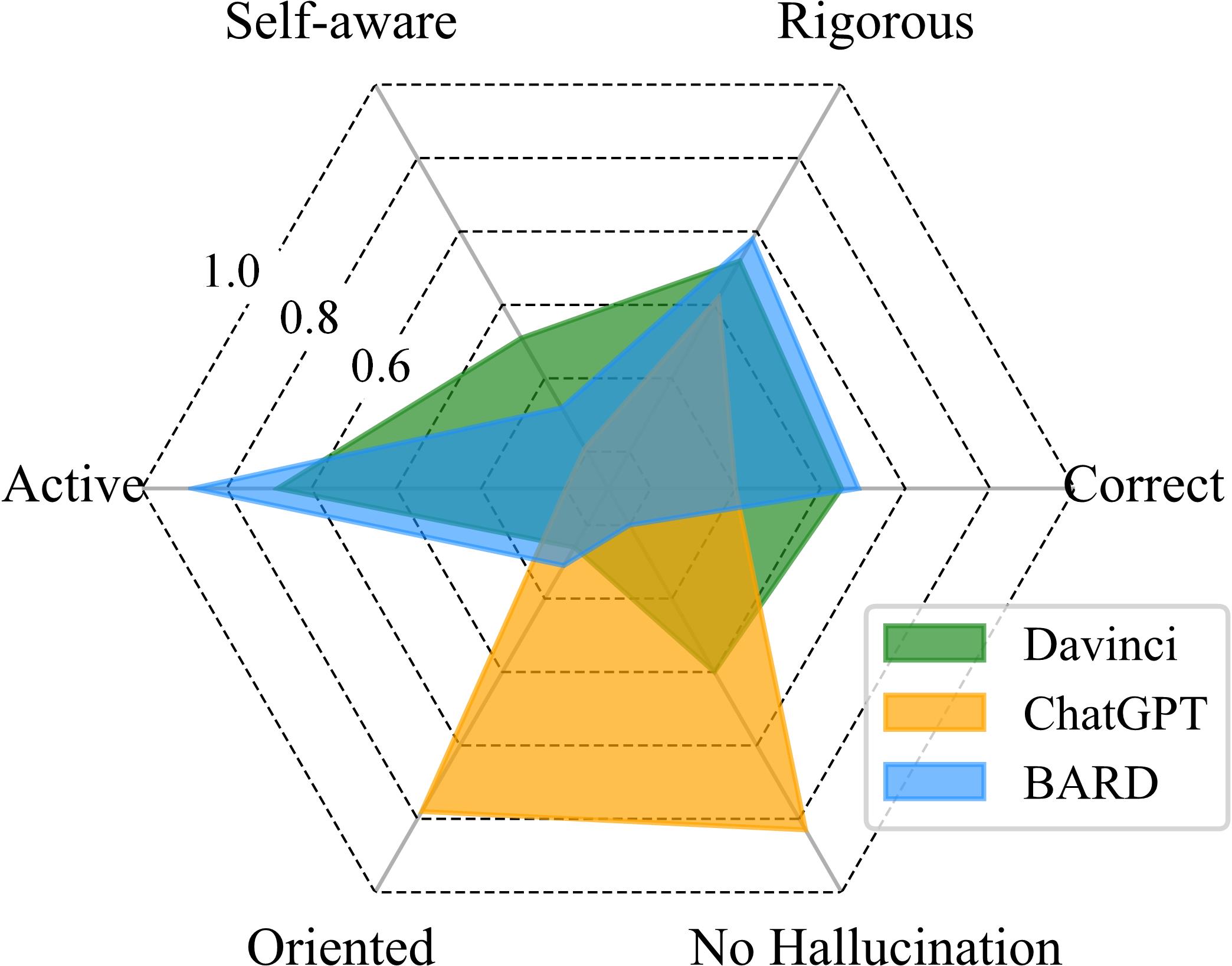}
		\vspace{-0.1cm}
		\subcaption{Inductive.}
		\label{res_in}
	\end{minipage}
	\begin{minipage}[t]{0.24\linewidth}
		\large
		\centering
		\includegraphics[scale=0.42]{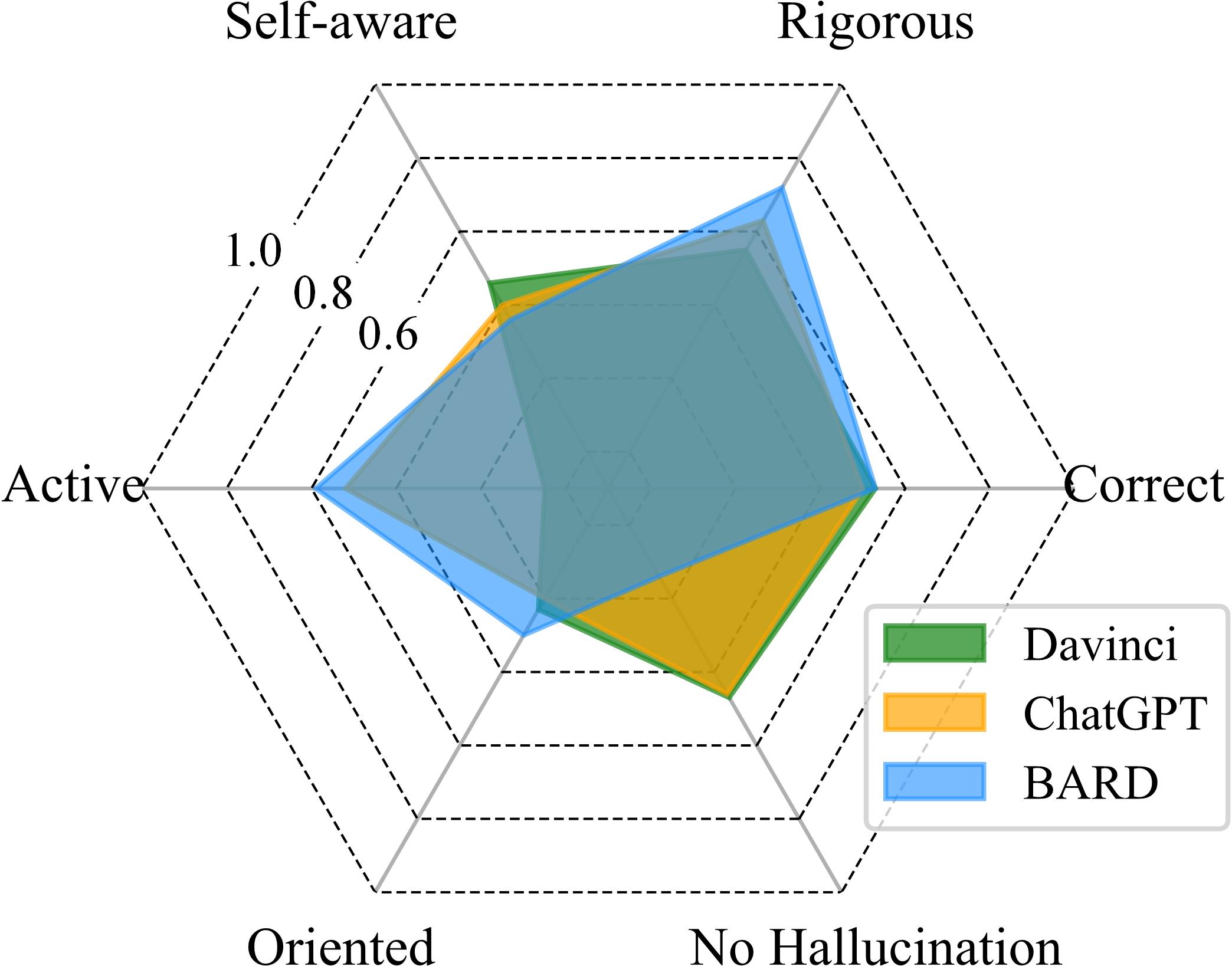}
		\vspace{-0.1cm}
		\subcaption{Abductive.}
		\label{res_ab}
	\end{minipage}
	\begin{minipage}[t]{0.24\linewidth}
		\large
		\centering
		\includegraphics[scale=0.42]{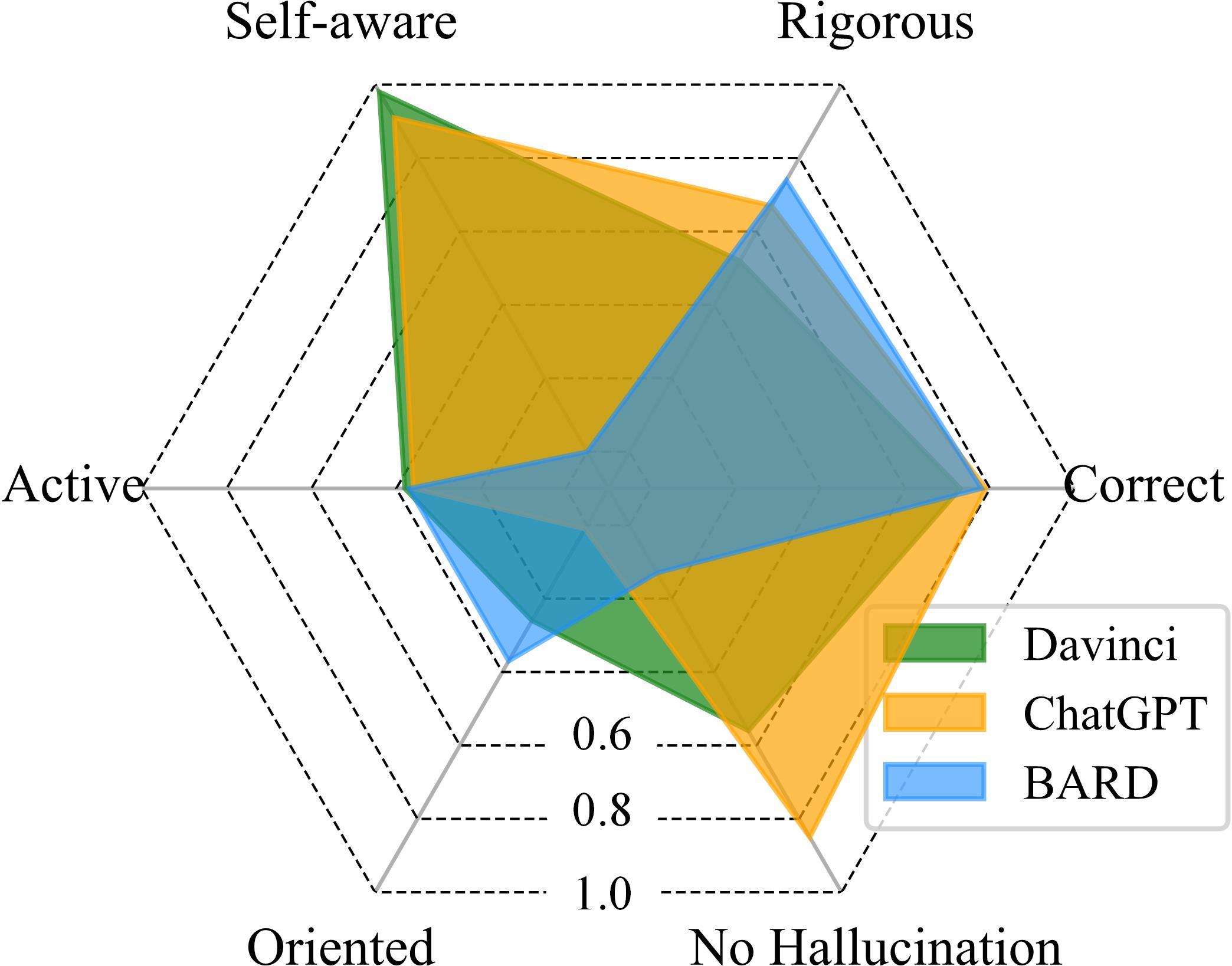}
		\vspace{-0.1cm}
		\subcaption{Mixed-form.}
		\label{res_mix}
	\end{minipage}
	\caption{Visualization of three early-stage LLM capabilities under four reasoning settings.}
	\label{res_llm}
\end{figure*}

\begin{figure*}[t]
    \begin{minipage}[t]{0.23\linewidth}
        \large
        \centering
        \includegraphics[scale=0.47]{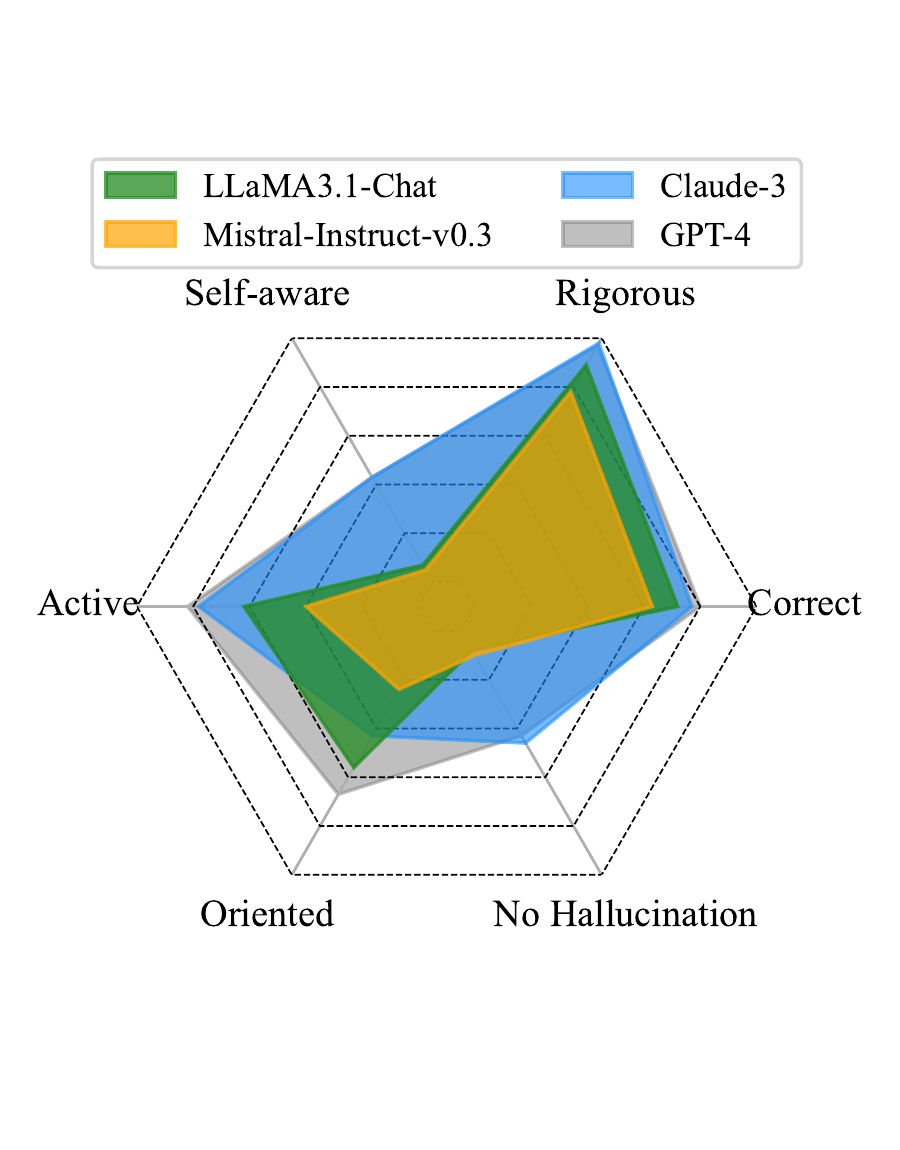}
        \vspace{-1.5cm}
        \subcaption{Deductive.}
        \label{heatmap_active}
    \end{minipage}
        \begin{minipage}[t]{0.23\linewidth}
        \large
        \centering
        \includegraphics[scale=0.47]{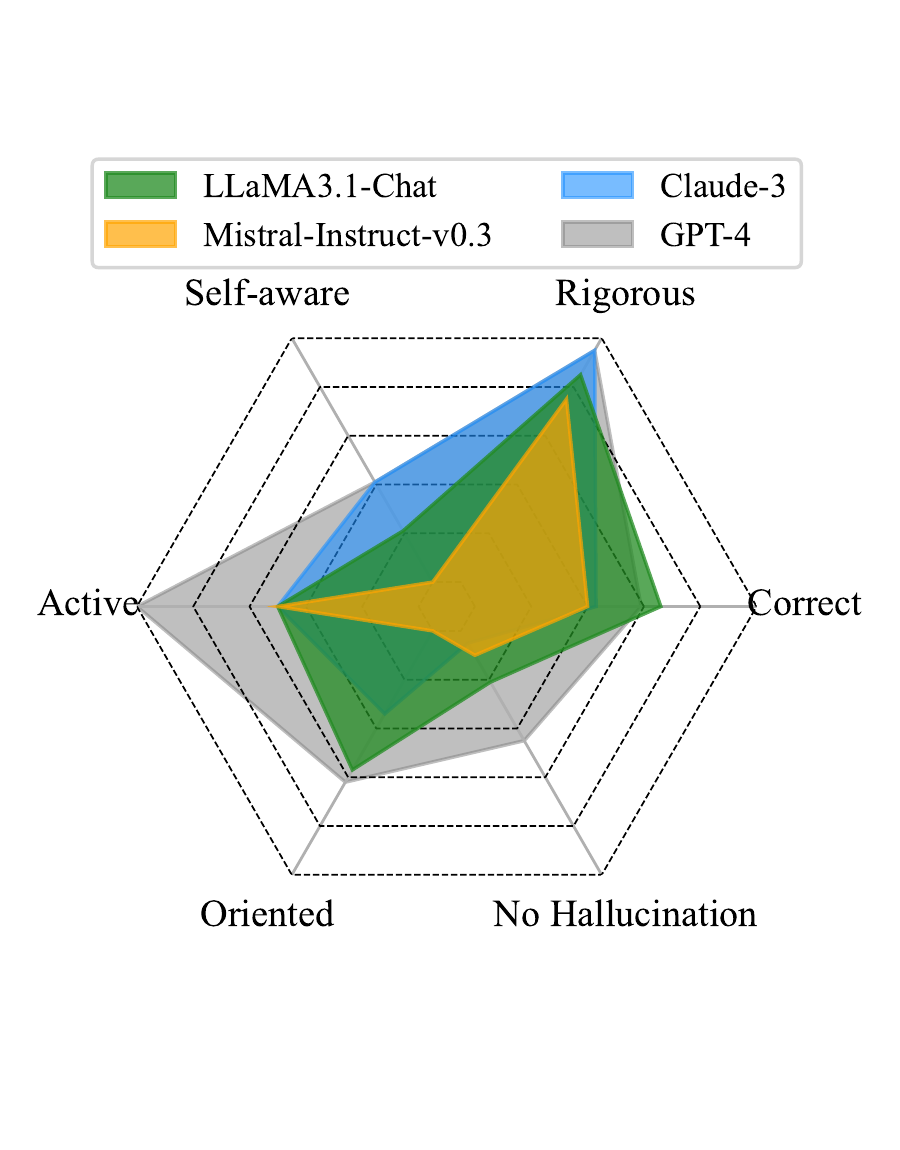}
        \vspace{-1.5cm}
        \subcaption{Inductive.}
        \label{heatmap_oriented}
    \end{minipage}
    \begin{minipage}[t]{0.23\linewidth}
        \large
        \centering
        \includegraphics[scale=0.47]{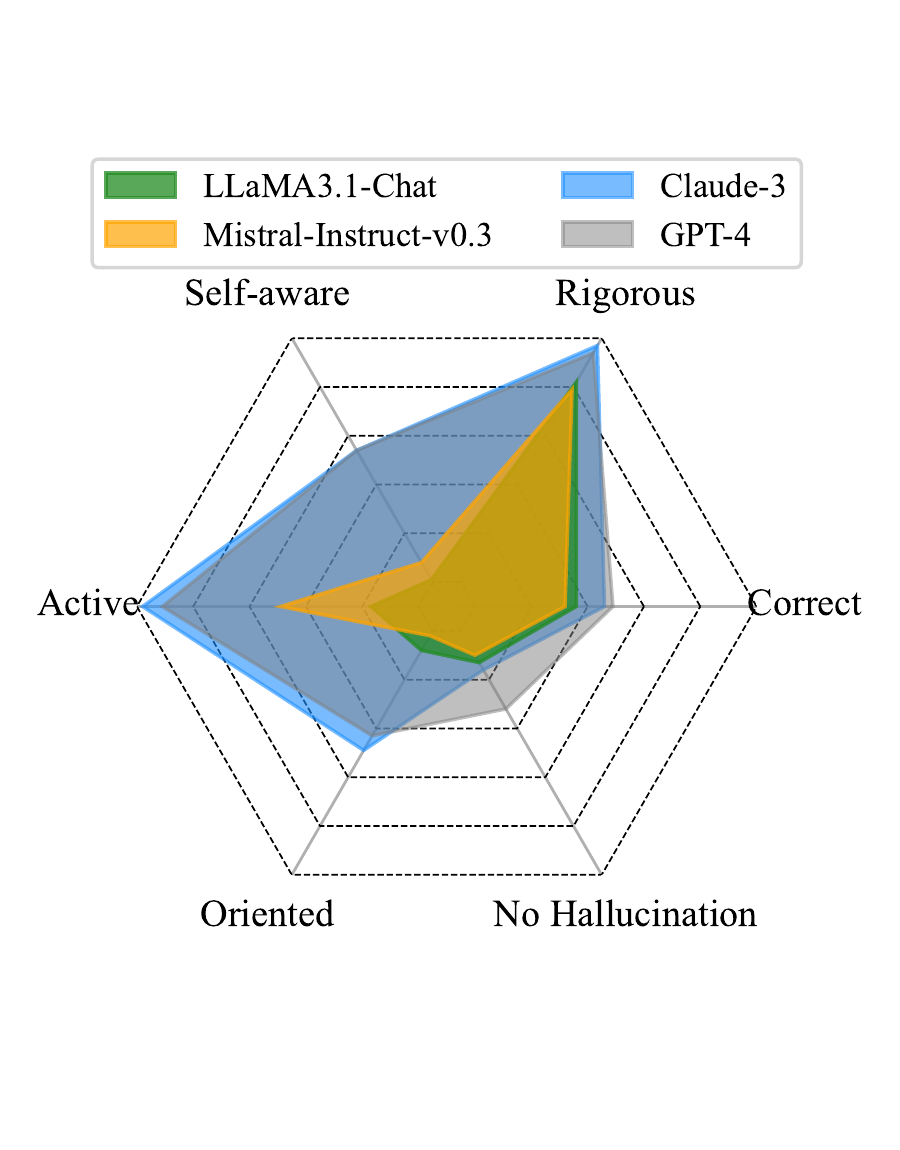}
        \vspace{-1.5cm}
        \subcaption{Abductive.}
        \label{heatmap_hallu}
    \end{minipage}
    \begin{minipage}[t]{0.23\linewidth}
        \large
        \centering
        \includegraphics[scale=0.47]{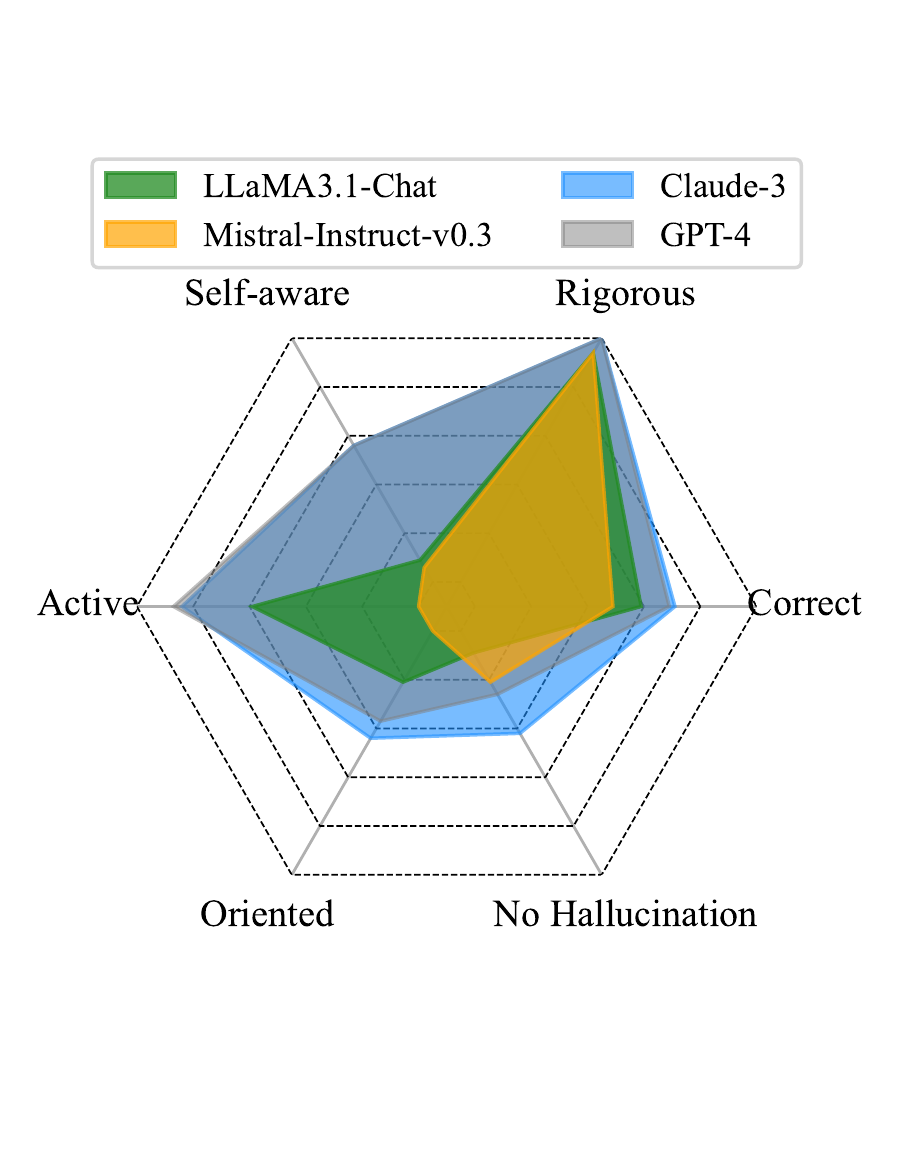}
        \vspace{-1.5cm}
        \subcaption{Mixed form.}
        \label{heatmap_hallu}
    \end{minipage}
    \caption{{\color{black}Visualization of LLM logical reasoning capability under four reasoning settings, including LLaMA3.1-Chat (8B), Mistral-Instruct-v0.3, Claude-3, and GPT-4 for comparisons.}}
    \label{dimensions}
\end{figure*}

\section{Conclusion}
In this paper, in-depth evaluations are conducted on logical reasoning tasks, discussing whether LLMs are really good logical reasoners. First, the logical reasoning evaluations are organized from deductive, inductive, abductive and mixed-form views. We select fifteen logical reasoning datasets to evaluate on three representative LLMs (i.e., text-davinci-003, ChatGPT and BARD) under both zero-shot and few-shot settings. Second, this paper provides fine-level evaluations on four metrics, covering both objective and subjective views. For problematic cases, extensive error attributions are conducted from two dimensions, forming five error types. It uncovers the logical flaws of LLMs and we provide deep analysis on the results. Third, to achieve a fair and pure benchmark for logical reasoning capability, we propose a dataset with neutral content, covering deductive, inductive and abductive settings.

Based on the evaluation results above, we abstract six dimensions to measure the logical reasoning capability of LLMs: (1) \emph{Correct}, (2) \emph{Rigorous}, (3) \emph{Self-aware}, (4) \emph{Active}, (5) \emph{Oriented} and (6) \emph{No hallucination}. All these properties can be calculated with the evaluation methods proposed in this paper. Therefore, we propose an evaluation scheme for the logical reasoning capability of LLMs. Considering the different performances of LLMs on deductive, inductive, abductive and mixed-form settings, we respectively visualize each ability map in Fig.~\ref{res_llm}. 

According to the results, text-davinci-003 can maintain balanced performances in deductive and mixed-form settings. But it usually fails to keep oriented for reasoning in the inductive setting, and it also shows laziness in the abductive reasoning tasks. Since it is the earliest released LLM of the three, it is understandable that text-davinci-003 has some limitations in logical reasoning tasks, especially in the more complex settings (e.g., inductive and abductive).

From the perspective of common evaluations, ChatGPT is the weakest LLM of the three, since it performs badly in showing correct and rigorous reasoning under deductive, inductive and abductive settings. Also, it seems to be the laziest reasoner in deductive and inductive settings. However, it surprises us that it shows unique advantages in maintaining oriented reasoning and avoiding hallucination, especially in deductive and inductive settings. In addition, it shows its comprehensive capability in the mixed-form setting. We argue that ChatGPT is specially designed for chatting, thus it does pretty well in keeping rational but is not good at solving complex reasoning problems. 

BARD is the most active reasoner and it keeps great competitiveness as a correct and rigorous reasoner. However, it also shows obvious flaws compared with other LLMs. BARD tends to generate redundant content, easily fails to find the correct reasoning directions and it usually fails to avoid hallucinations. In short, BARD shows great advantages in current benchmarks with objective metrics, due to the larger model size and massive training data. But it still has much room for improvement in some implicit aspects, i.e., self-awareness, orientation and non-hallucination. 

Overall, it can be observed that \textbf{all LLMs exhibit specific limitations in logical reasoning, with relative strength in deductive reasoning but evident struggles in inductive settings}. Moreover, current evaluation benchmarks, which primarily depend on objective metrics, are not sufficient to comprehensively evaluate LLMs.

{\color{black}
Additionally, we supplement the visualization of four trending LLMs in Fig.~\ref{dimensions}.
Different from Fig.~\ref{res_llm}, these metrics are calculated in the automatic mode.
It also verifies the scalability of our evaluation system.
Please refer to the Appendix for more analysis on these four LLMs and evaluation settings. 
}

\section{Future Directions}
Based on the evaluation results, this paper concludes six future directions for logical reasoning tasks.

\noindent \textbf{Strengthen the reasoning ability of inductive reasoning.}
Inductive reasoning draws broad conclusions from specific observations, requiring a more abstract and comprehensive understanding of real-world knowledge compared to deductive or abductive reasoning. However, LLMs have shown poor performance in this area, as demonstrated in Section 4.1. Therefore, it is crucial to develop pre-training or fine-tuning strategies to enhance their inductive reasoning abilities. One such strategy could be constructing more inductive instructions to guide LLMs.

\noindent \textbf{Enhance the LLM's perception of its capability boundaries.}
LLMs excel at providing answers and explanations for a wide range of reasoning questions, irrespective of complexity. However, as evidenced in Sections 5.1 and 5.4, they may introduce irrelevant information or fabrications, leading to nonsensical or illogical outcomes. A proficient logical reasoner should recognize its limitations and knows when to refrain from answering. Future investigations could explore human self-awareness in cognitive science and neuroscience to strengthen LLMs' understanding of their operational boundaries.

\noindent \textbf{Strengthen the rigorous reasoning to apply to real-world scenarios.}
Table~\ref{explain_true_complete} illustrates that current LLMs are not sufficiently rigorous for deductive, inductive, and abductive reasoning.
As a result, there exist significant gaps between their capabilities and potential applications in real-world scenarios,
particularly those that require detailed intermediate explanations.
For instance, using LLMs to solve mathematical problems and provide precise intelligent Q\&A services in the education field remains a significant challenge~\cite{drori2022neural}.

\noindent \textbf{Minimize the occurrence of hallucinations.}
Similar to the behaviors in other problem-solving contexts~\cite{DBLP:journals/csur/JiLFYSXIBMF23},
LLMs may generate false or irrelevant hallucinations during logical reasoning tasks.
It suggests that LLMs may not fully comprehend the question and can not solve it correctly. To address this issue, future research should develop more comprehensive evaluation metrics for hallucinations and explore specific strategies to minimize their occurrence.

\noindent \textbf{Improve the multi-hop reasoning capability, especially in inductive and abductive settings.}
Combined with the results from Fig.~\ref{num_hop} and Fig. 1 in the Appendix, the multi-hop reasoning capability of LLMs still have much room for improvement. 
In fact, humans are better at decomposing complex questions. It can be an interesting topic for LLMs to capture the ability to divide and conquer questions, thus benefiting multi-hop reasoning.

\noindent \textbf{Increase explainability.}
Finally, the explainability of LLMs will be essential for building trust, detecting and mitigating biases, improving performance, promoting user understanding, and complying with regulations. A commonsense-based neuro-symbolic AI framework, such as the one proposed by~\cite{camnt7} for sentiment analysis, can help increase the explainability of the reasoning processes required for decision-making, which is crucial for sensitive applications involving ethics, privacy and health.

\ifCLASSOPTIONcompsoc
  \section*{Acknowledgments}
\else
  \section*{Acknowledgment}
\fi

This work was supported by National Key Research and Development Program of China (2020AAA0108800),
National Natural Science Foundation of China (62137002, 61937001, 62176209, 62176207, 62106190, and 62050194).

\ifCLASSOPTIONcaptionsoff
  \newpage
\fi

\bibliographystyle{IEEEtran}
\bibliography{tkde}

\newpage

\appendix
\appendices
The following sections firstly provide comprehensive information about the seven LLMs analyzed in this study. Additionally, this file includes a detailed description of the evaluation methodology, prompt engineering techniques, NeuLR dataset, and supplementary analyses. Finally, the case studies for each dataset are presented. Together, these sections provide a comprehensive and detailed account of the methods and results of the study.

\section{Details of Large Language Models}
{\color{black}In this study, we select three previously representative LLMs (i.e., text-davinci-003, ChatGPT and BARD) and four up-to-date LLMs (i.e., LLaMA3.1-Chat, Mistral-Instruct-v0.3, Claude-3 and GPT-4) for evaluation and analysis.}
The specifics are itemized in Table~\ref{model}, which includes the affiliation, usage charge, pre-trained data, and model size of each LLM.

\begin{table}[H]
	\centering
	\small
	\caption{Details of the selected LLMs. \emph{Affi.} is short for \emph{Affiliation}. \emph{Charge} represents the charges for 1M tokens in the format of (input charge / output charge). \emph{Data} is the latest time of the utilized training data.
 \emph{B} in the last column represents the one billion.}
	\resizebox{\linewidth}{!}{
        \begin{tabular}{c|cccc}
            \toprule
            \textbf{Model} &\textbf{Affi.} &\textbf{Charge}  &\textbf{Data} &\textbf{Size}\\
            \midrule
            text-davinci-003 &Open-AI &20\$ &Sep. 2021 &-  \\
            ChatGPT &Open-AI &2\$ &Jun. 2021 &- \\
            BARD &Google &- & Not report &540B \\
            \midrule
            LLaMA3.1-Chat &Meta &- &2023 &8B \\
            Mistral-Inst-v0.3 &Mistral &- &2023 &7B \\
            Claude-3 &Anthropic &3\$/15\$ &2023 &- \\
            GPT-4 &Open-AI &30\$/60\$ &2023 &-  \\
		\bottomrule
	\end{tabular}
	\vspace{-0.2cm}
	\label{model}
        }
\end{table}

\section{Details of Evaluations}
This section supplements some details of the evaluation.

\subsection{Evaluated Datasets}

Considering the huge annotation load, we make a balance between the comprehensiveness and annotation cost. We try to keep the full data of each dataset to evaluate ChatGPT, except for some large-scale datasets, i.e., RuleTaker, $\alpha$-NLI, $\alpha$-NLG, AbductiveRules, D*-Ab and LogiQA2NLI. We limit the number of evaluation samples in these datasets to 1000, 1000, 1000, 1000, 1000 and 600 respectively. For text-davinci-003 and BARD, we randomly sample 100 examples for each dataset. It is still more comprehensive than previous works, that only utilize dozens of samples for testing (e.g., 30 samples for each dataset).

{\color{black} Notably, as for the supplemented experiments on the four latest LLMs (LLaMA3.1-Chat, Mistral-Instruct-v0.3, Claude-3 and GPT-4), we randomly sample 100 cases for each dataset.
It is expected to maintain a fair comparison.}

{\color{black} For the mixed-form reasoning datasets,
we calculate the fact length distributions of four datasets for reference.
It is observed that most samples have 1-2 or 3-4 facts for reasoning.}

\begin{table}[H]
	\centering
	\caption{Fact length distributions (\%) of mixed-form datasets.}
	\begin{tabular}{c|ccccc}
            \toprule
            \multirow{2}{*}{\textbf{Model}} &\multicolumn{5}{c}{\textbf{Length}}\\
            &\textbf{1-2} &\textbf{3-4}  &\textbf{5-6} &\textbf{7-8}&\textbf{$\geq$ 9}\\
            \midrule
            ReClor &28.20 &61.60 &9.00 &1.00&0.20  \\
            LogiQA &76.65 &13.36 &7.22 & 1.84&0.92 \\
            LogiQA 2.0 &39.40 &45.80 & 12.60 &1.60&0.60 \\
            LogiQA2NLI &34.94 &43.56 & 16.95 &3.07&1.46 \\
		\bottomrule
	\end{tabular}
	\label{fact_length}
\end{table}

\subsection{Evaluation Modes}
As mentioned in the main paper, we introduce four metrics to evaluate the LLM performances, i.e., \emph{Answer Correctness}, \emph{Explain Correctness}, \emph{Explain Completeness} and \emph{Explain Redundancy}. Also, we employ five metrics to attribute the error cases. {\color{black} Although these metrics are subjective, their standards are clear and well-defined.}

As for the three previously representative LLMs,
their evaluations are conducted through manual annotation. 
In detail, we output the LLM answers through API calling and hire ten graduate students (all majoring in Natural Language Processing) to annotate the answers based on our metric definitions. 

{\color{black}
As for the four trending LLMs,
their evaluations are under the \textbf{automatic mode}.
Specifically, we utilize GPT-4o model to act as the LLM judge.
The definitions of the metrics are fed to the LLM judge as instructions,
and it outputs the evaluation scores the same as the human annotators.
}

\section{Supplementary Results on Latest LLMs}

{\color{black}
First of all, we conduct the pilot experiments to demonstrate the consistency between the manual labeling and GPT-4o. The analysis is conducted under the four evaluation metrics (i.e., answer correctness, explanation correctness, explanation completeness, and explanation redundancy) and the two error categories (i.e., evidence selection and reasoning errors).
Fig.~\ref{llm_human} shows the results.

\begin{figure}[h]
    \centering
    \includegraphics[scale=0.6]{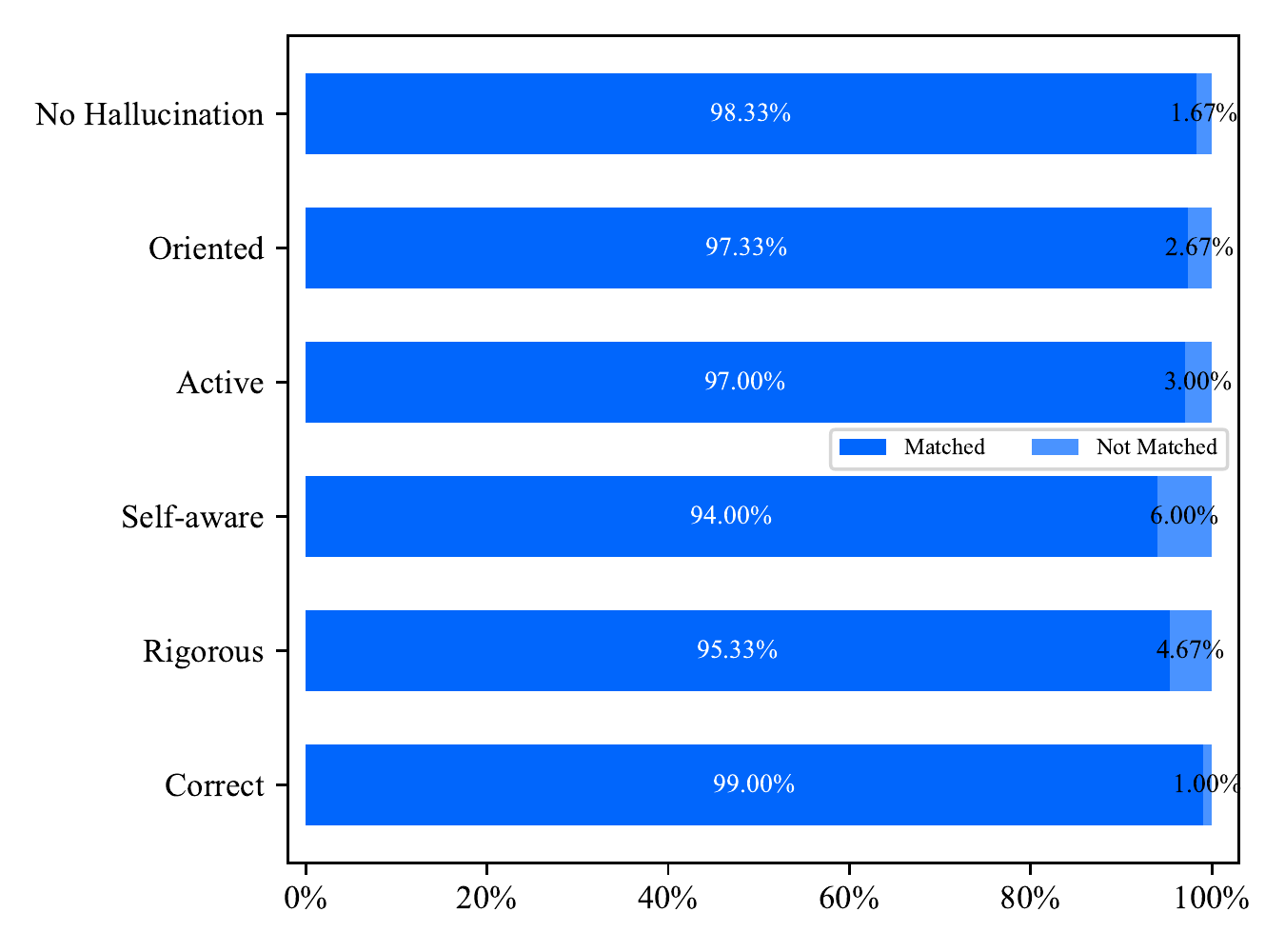}
    \caption{The evaluation matches between LLM judge and human annotations on the 300 samples.}
    \label{llm_human}
\end{figure}

\begin{figure*}[h]
    \begin{minipage}[t]{0.33\linewidth}
        \large
        \centering
        \includegraphics[scale=0.72]{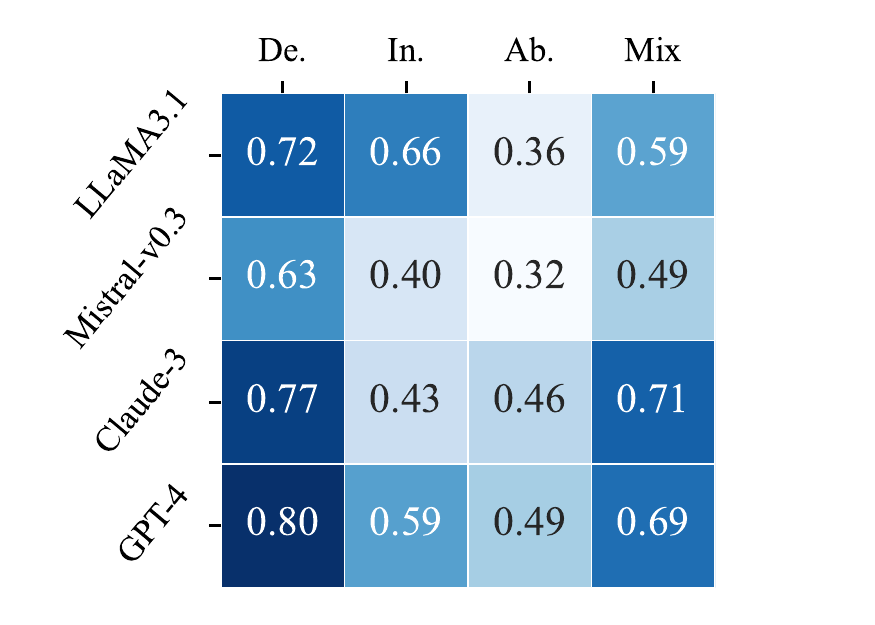}
        \subcaption{Correctness.}
        \label{heatmap_active}
    \end{minipage}
        \begin{minipage}[t]{0.33\linewidth}
        \large
        \centering
        \includegraphics[scale=0.72]{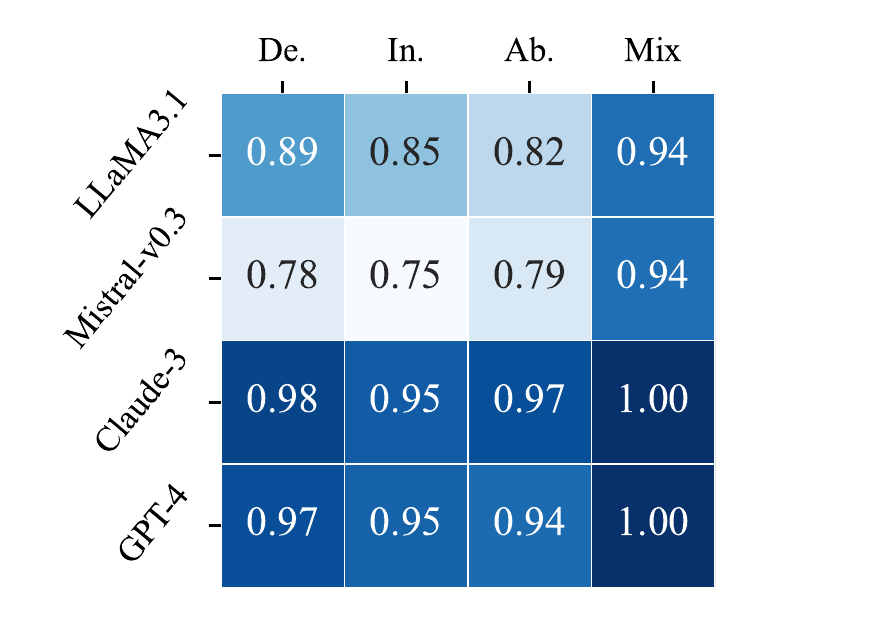}
        \subcaption{Rigorousness.}
        \label{heatmap_oriented}
    \end{minipage}
    \begin{minipage}[t]{0.33\linewidth}
        \large
        \centering
        \includegraphics[scale=0.72]{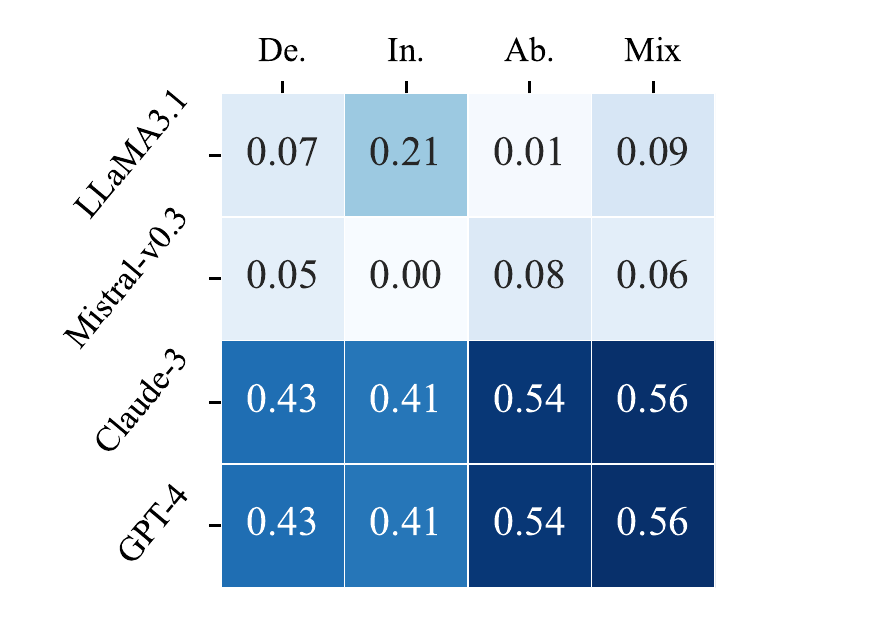}
        \subcaption{Self-awareness.}
        \label{heatmap_hallu}
    \end{minipage}
    \caption{Heatmap results for the correctness, rigorousness, and self-awareness metrics.}
    \vspace{-0.2cm}
    \label{heatmap_metrics}
\end{figure*}

\begin{figure*}[h]
    \begin{minipage}[t]{0.33\linewidth}
        \large
        \centering
        \includegraphics[scale=0.72]{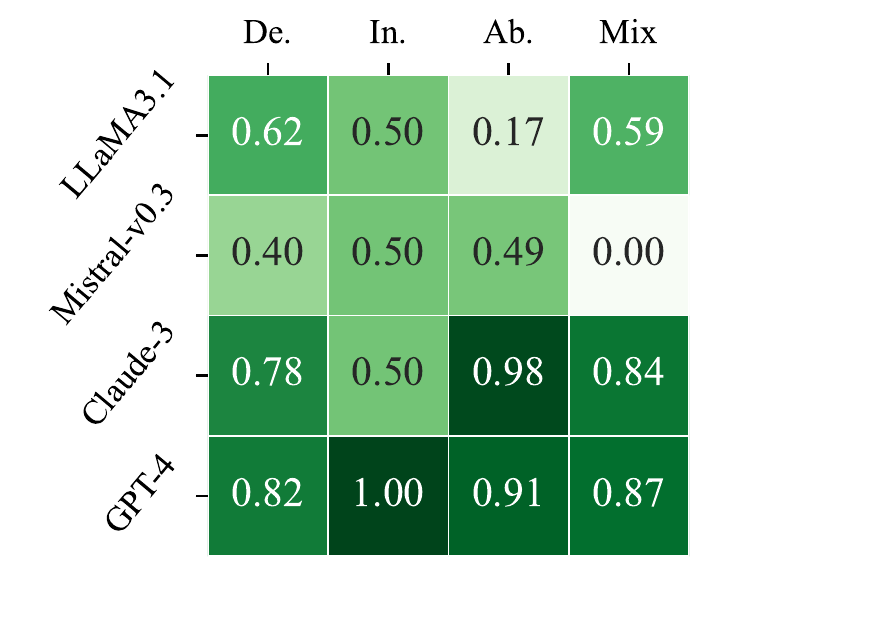}
        \subcaption{Active.}
        \label{heatmap_active}
    \end{minipage}
        \begin{minipage}[t]{0.33\linewidth}
        \large
        \centering
        \includegraphics[scale=0.72]{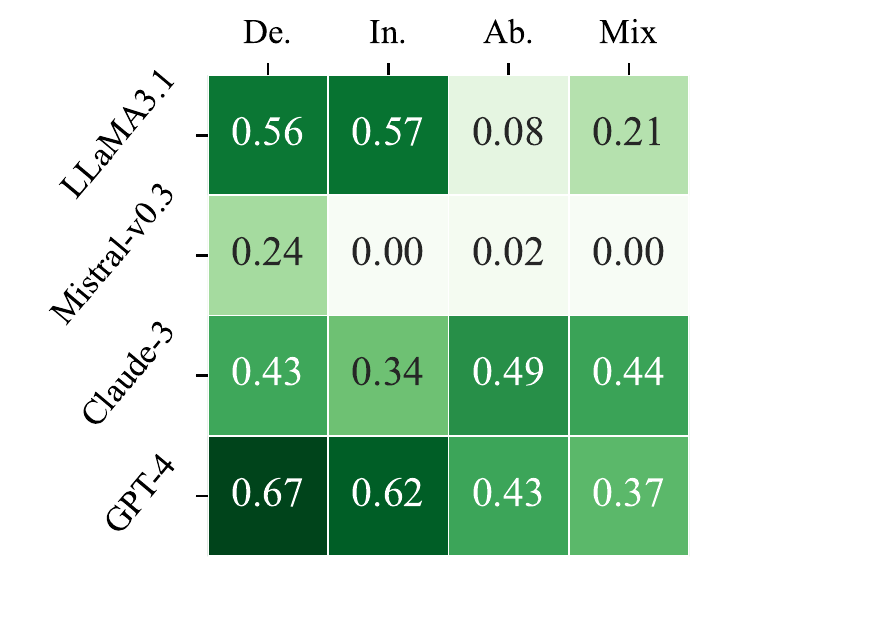}
        \subcaption{Oriented.}
        \label{heatmap_oriented}
    \end{minipage}
    \begin{minipage}[t]{0.33\linewidth}
        \large
        \centering
        \includegraphics[scale=0.72]{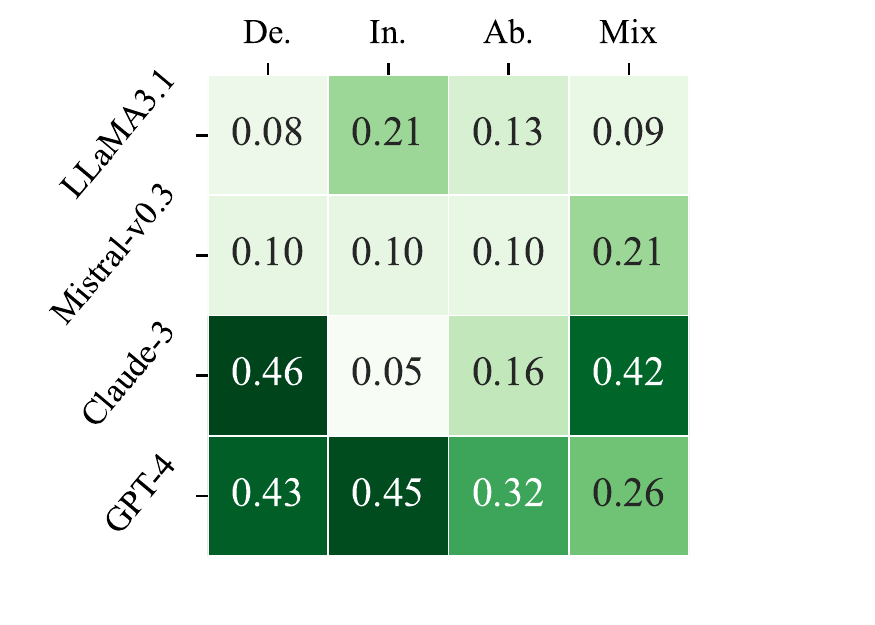}
        \subcaption{No hallucination.}
        \label{heatmap_hallu}
    \end{minipage}
    \caption{Heatmap results for the activity, orientation and no-hallucination of LLMs.}
    \vspace{-0.1cm}
    \label{heatmap_error}
\end{figure*}

It illustrates that LLM-as-judge can well align with human performance in our evaluation settings.

Next, we implement the \textbf{automatic evaluations} on the four trending LLMs (i.e., LLaMA3.1-Chat-8B, Mistral-Instruct-v0.2-7B, Claude-3, and GPT-4). Fig.~\ref{heatmap_metrics} and~\ref{heatmap_error} provide the respective results on the evaluation of six dimensions.

}

\section{Prompt Engineering}
In the implementation, we consider both the zero-shot and few-shot settings. To obtain appropriate model outputs, we construct the prompts, shown in Table \ref{prompt} of Appendix.

\begin{table*}[h]
	\centering
	\small
	\caption{Evaluated Datasets. \emph{Gen.} distinguishes whether the 
	predicted answer is generated text or classified labels. \emph{Explain} 
	denotes whether the explanation is required in the task. \# Davinci, \# ChatGPT and \# BARD columns represent the number of evaluation samples of each dataset for the three LLMs.
 } 
	\begin{tabular}{c|c|c|c|c|ccc}
		\toprule
            \textbf{Categories} & \textbf{Dataset} &\textbf{Source} & \textbf{Gen.} &\textbf{Explain} &\textbf{\# Davinci} &\textbf{\# ChatGPT} &\textbf{\# BARD} \\
            \midrule
            \multicolumn{1}{c|}{\multirow{5}{*}{Deductive}} & \textbf{bAbI-15} &~\cite{DBLP:journals/corr/WestonBCM15} &\checkmark &\checkmark &100 &1,000 &100\\
            & \textbf{EntailmentBank}&~\cite{DBLP:conf/emnlp/DalviJTXSPC21} &\checkmark & &100 &340 &100\\
            & \textbf{RuleTaker} &~\cite{DBLP:conf/ijcai/ClarkTR20} & &\checkmark &100 &1,000 &100\\
            & \textbf{FOLIO} &~\cite{DBLP:journals/corr/abs-2209-00840} & & & 100 &204 &100 \\
            & \textbf{Leap-Of-Thought} &~\cite{DBLP:conf/nips/TalmorTCGB20} & && 100 &1,289 &100 \\
            \midrule
            \multicolumn{1}{c|}{\multirow{2}{*}{Inductive}} & \textbf{bAbI-16} &~\cite{DBLP:journals/corr/WestonBCM15} &\checkmark & &100 &1,000 &100 \\
            & \textbf{CLUTRR} &~\cite{DBLP:conf/emnlp/SinhaSDPH19} &\checkmark & &100 &1,146 &100 \\
            \midrule
             \multicolumn{1}{c|}{\multirow{4}{*}{Abductive}} & \textbf{$\alpha$-NLI} &~\cite{DBLP:conf/iclr/BhagavatulaBMSH20} & & &100 &1,000 &100\\
            & \textbf{$\alpha$-NLG} &~\cite{DBLP:conf/iclr/BhagavatulaBMSH20} &\checkmark & &100 &1,000 &100 \\
            & \textbf{AbductiveRules} &~\cite{DBLP:conf/acl/Young0BW22} &\checkmark &  &100 &1,000 &100 \\
            & \textbf{D*-Ab} &~\cite{DBLP:conf/acl/TafjordDC21} &\checkmark &\checkmark &100 &1,000 &100 \\
		\midrule
            \multicolumn{1}{c|}{\multirow{4}{*}{mixed-form}} & \textbf{ReClor} &~\cite{DBLP:conf/iclr/YuJDF20} & & &100 &500 &100 \\
            & \textbf{LogiQA} &~\cite{DBLP:conf/ijcai/LiuCLHWZ20} & & &100 &651 &100 \\
            & \textbf{LogiQA 2.0} &~\cite{DBLP:journals/corr/abs-2304-03439} & & &100 &500 &100\\
            & \textbf{LogiQA2NLI} &~\cite{DBLP:journals/corr/abs-2304-03439} & & &100 &600 &100 \\
		\bottomrule
	\end{tabular}
	\vspace{-0.2cm}
	\label{datasets}
\end{table*}

\begin{figure*}[h]
	\begin{minipage}[t]{0.33\linewidth}
		\large
		\centering
		\includegraphics[scale=0.4]{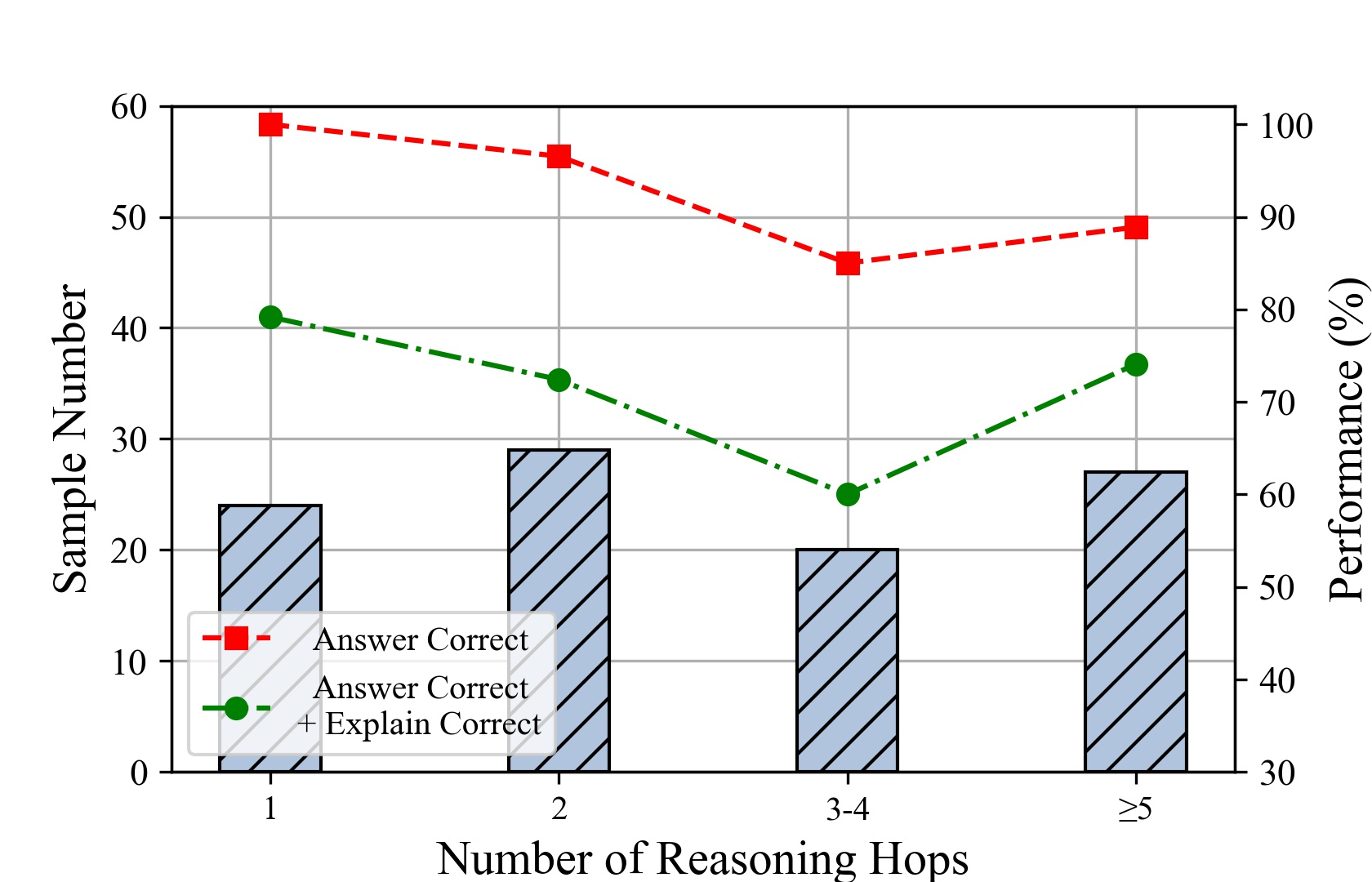}
		\vspace{-0.1cm}
		\subcaption{Deductive setting (EntailmentBank) on text-davinci-003.}
		\label{de}
	\end{minipage}
	\begin{minipage}[t]{0.33\linewidth}
		\large
		\centering
		\includegraphics[scale=0.4]{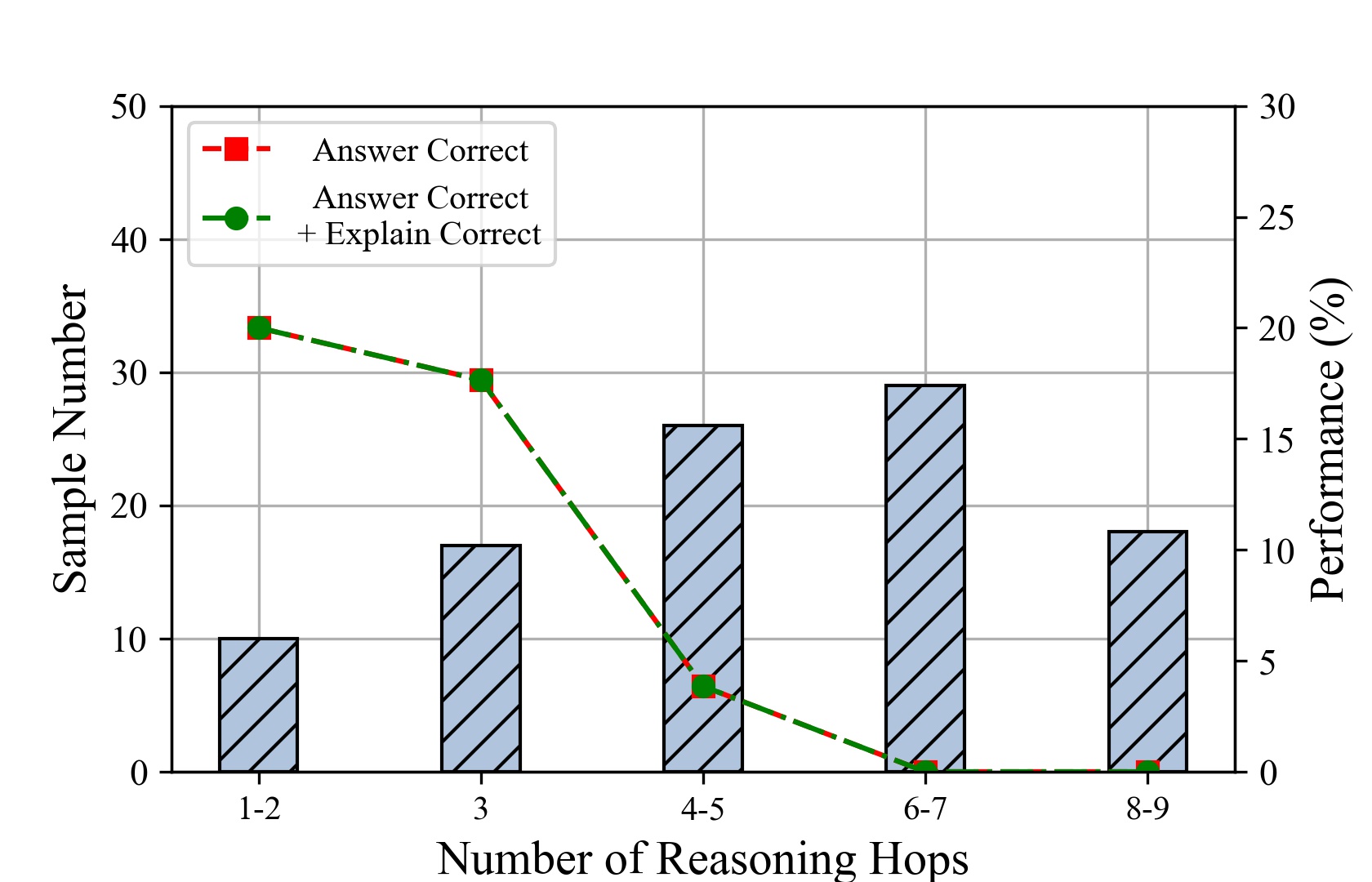}
		\vspace{-0.1cm}
		\subcaption{Inductive setting (CLUTRR) on text-davinci-003.}
		\label{in}
	\end{minipage}
	\begin{minipage}[t]{0.33\linewidth}
		\large
		\centering
		\includegraphics[scale=0.4]{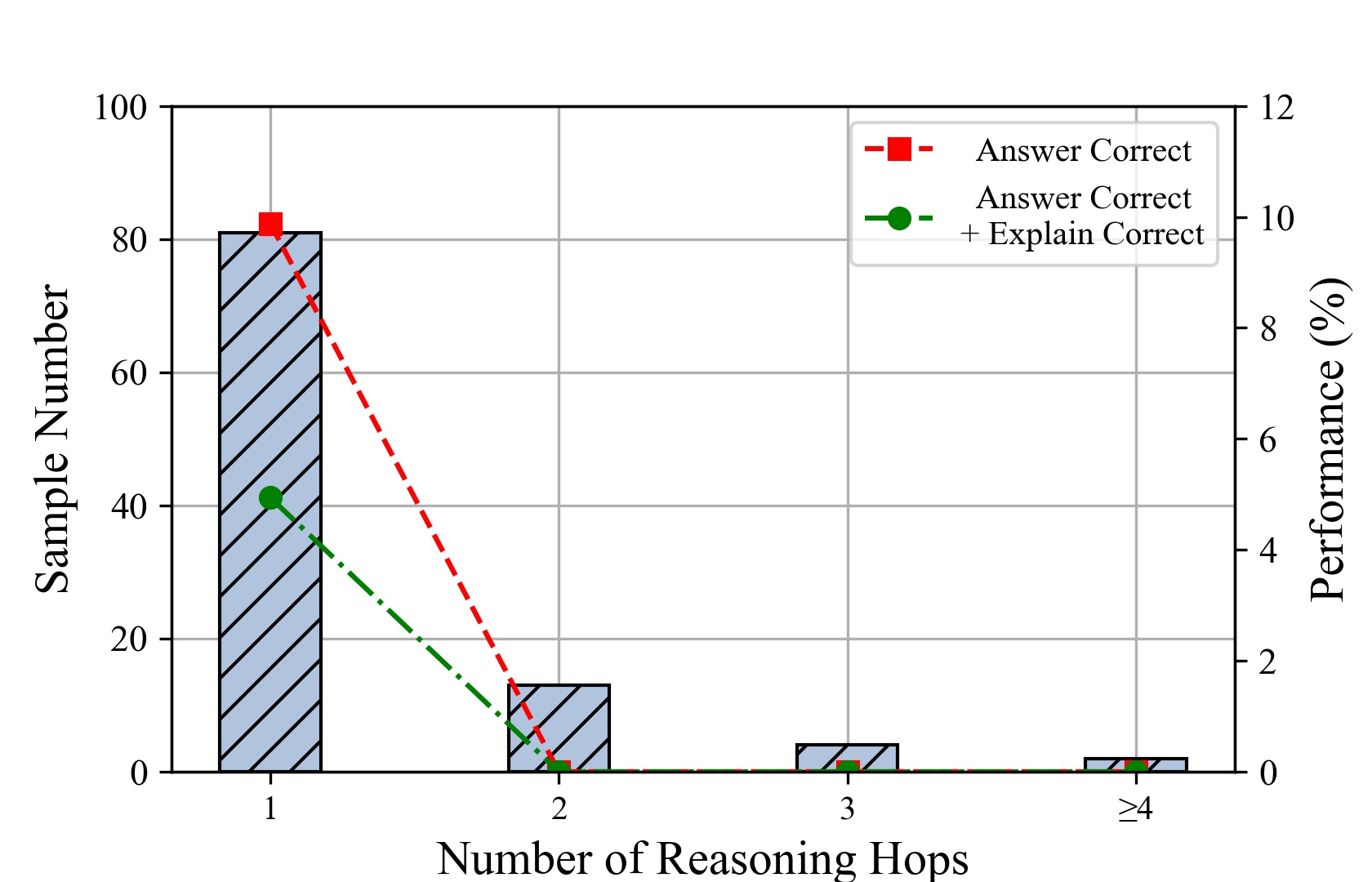}
		\vspace{-0.1cm}
		\subcaption{Abductive setting (D*-Ab) on text-davinci-003.}
		\label{in}
	\end{minipage}
        \\
	\begin{minipage}[t]{0.33\linewidth}
		\large
		\centering
		\includegraphics[scale=0.4]{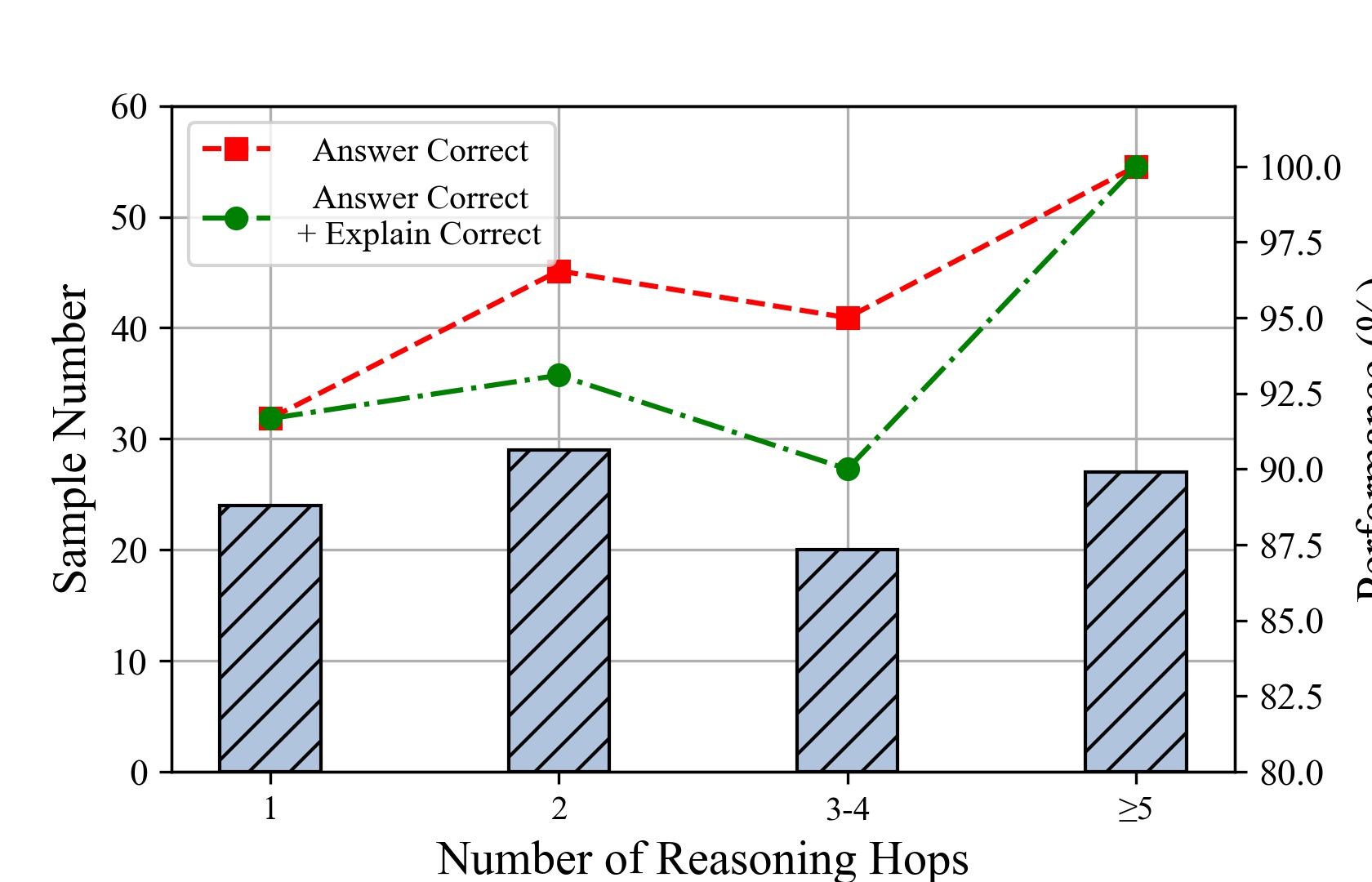}
		\vspace{-0.1cm}
		\subcaption{Deductive setting (EntailmentBank) on BARD.}
		\label{de}
	\end{minipage}
	\begin{minipage}[t]{0.33\linewidth}
		\large
		\centering
		\includegraphics[scale=0.4]{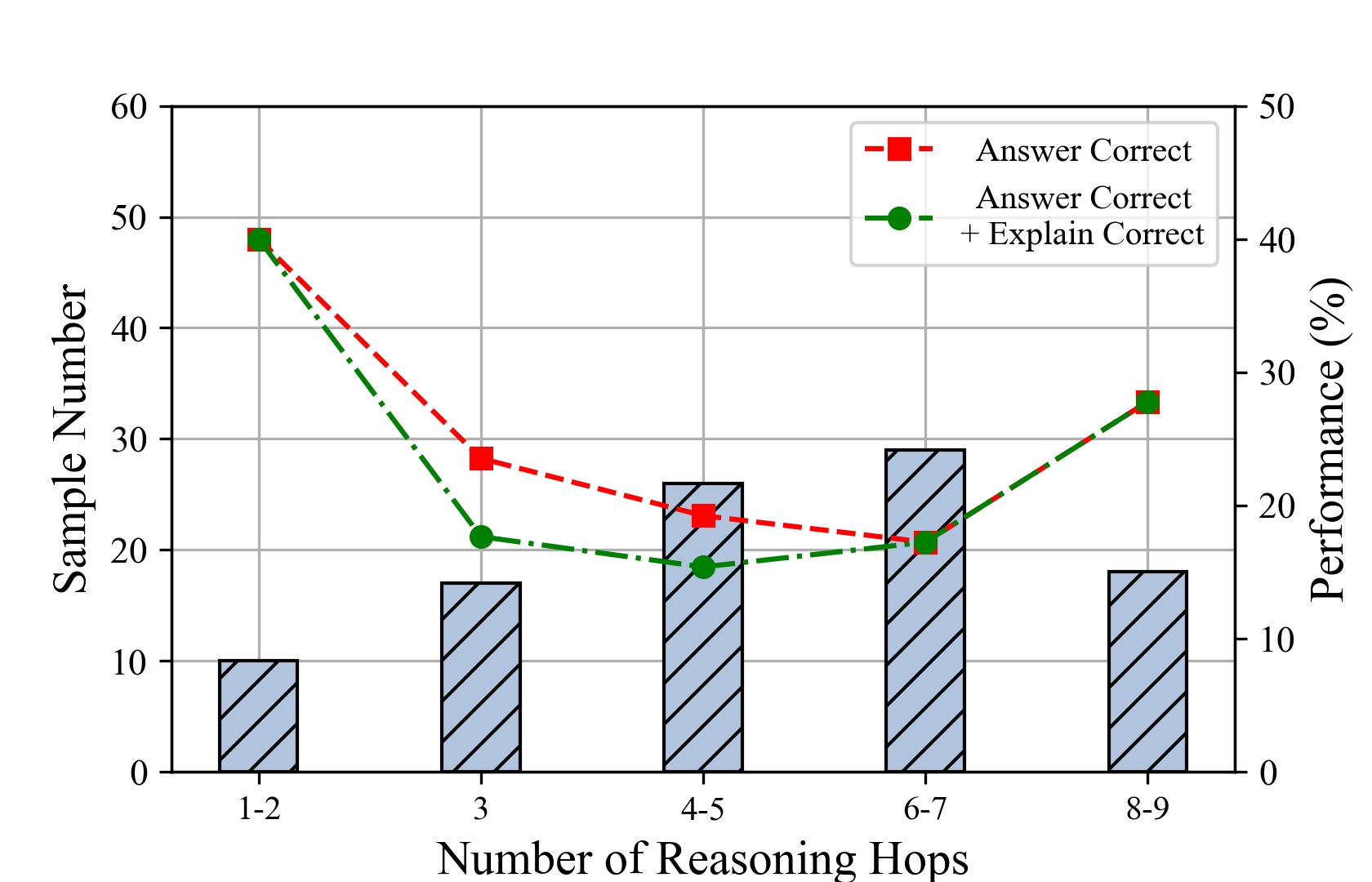}
		\vspace{-0.1cm}
		\subcaption{Inductive setting (CLUTRR) on BARD.}
		\label{in}
	\end{minipage}
	\begin{minipage}[t]{0.33\linewidth}
		\large
		\centering
		\includegraphics[scale=0.4]{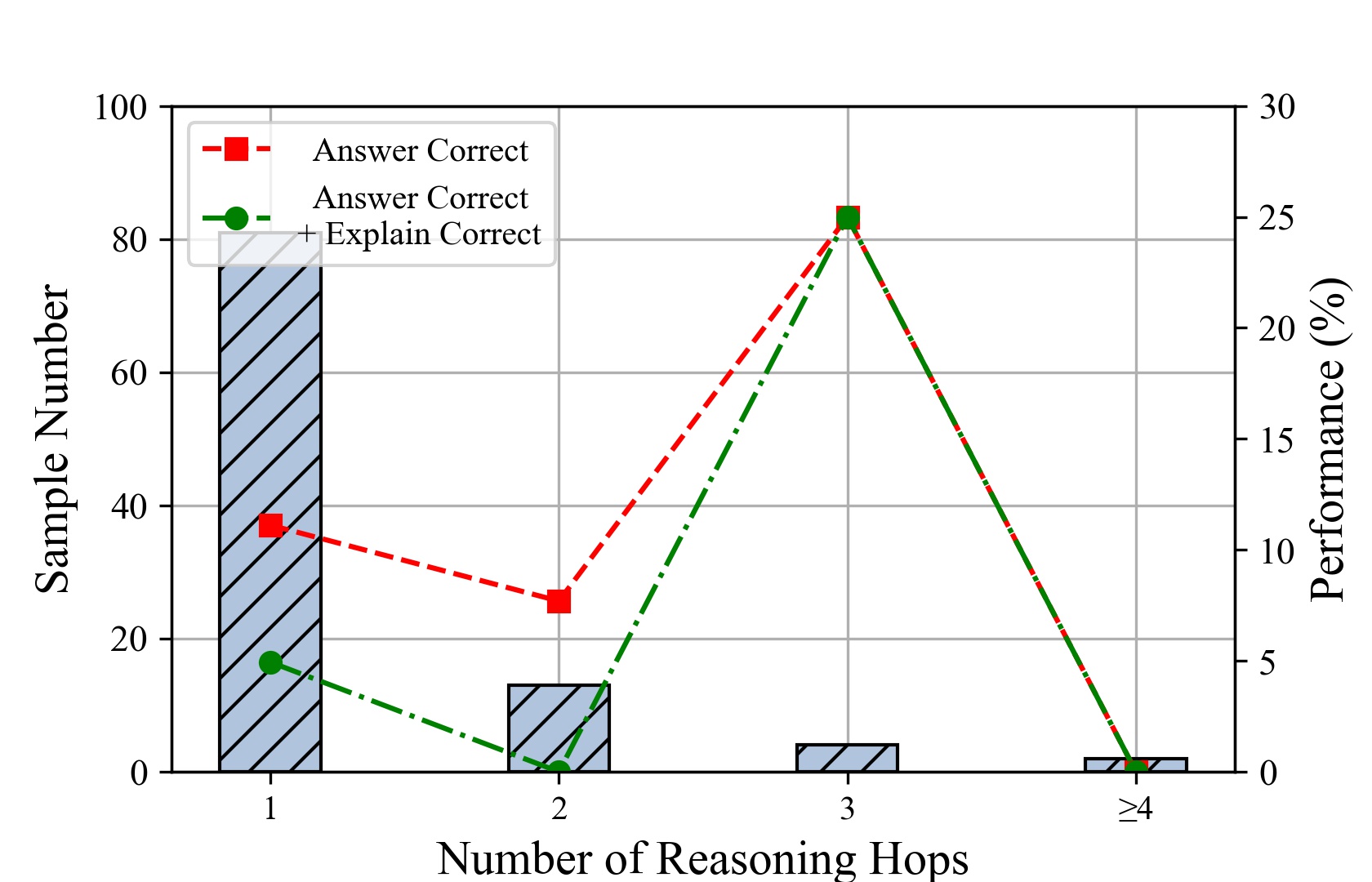}
		\vspace{-0.1cm}
		\subcaption{Abductive setting (D*-Ab) on BARD.}
		\label{in}
	\end{minipage}
	\caption{The performances of text-davinci-003 and BARD under different number of hops. Comparison of deductive, inductive and abductive reasoning settings.}
	\label{num_hop_appendix}
\end{figure*}

\section{NeuLR dataset}
In this section, we will provide more details on how NeuLR is constructed and what is the form of the prompt.

\subsection{Construction of NeuLR}
According to the explanations in the main paper, current logical reasoning may fail to keep the neutral content for evaluation. Therefore, we propose the new dataset NeuLR to benchmark the pure logical reasoning capability of LLMs. The data is sourced from \cite{DBLP:journals/corr/WestonBCM15,DBLP:conf/acl/TafjordDC21}. Detailedly, we transform some of the important content (e.g., entities and properties) into random strings. In NeuLR, we neutralize three types of words, i.e., name, species and property. We replace each type of words with the concatenation of a prefix (i.e., NP, SP, ADP) and six random characters (combination of uppercase and lowercase letters and numbers). Table \ref{construction} illustrates examples of the replacement process.

\begin{table}[H]
	\centering
	\small
	\caption{The construction of neutral content. \emph{Example} column presents some example words. \emph{Prefix} represents the prefix for each type of words. \emph{6-character} represents examples of randomly generated combination of characters.}
	\begin{tabular}{c|cccc}
            \toprule
            \textbf{Type} &\textbf{Example} &\textbf{Prefix}  &\textbf{6-character}\\
            \midrule
            Name &Bob/Lily &NP &e.g., uF52pT \\
            Species &Dog/Sheep &SP &e.g., 7gfO2k \\
            Property & Big/Red/Smart &ADP & e.g., PT01mx \\
		\bottomrule
	\end{tabular}
	\vspace{-0.2cm}
	\label{construction}
\end{table}

\subsection{Prompt Forms}
Different from previous evaluations, we consider adding descriptions of content neutrality in the prompt forms. Specifically, we replace the previous task description prefix with the following form (take abductive reasoning for illustration):

\begin{itemize}
    \item This is a neutral-content abductive reasoning task. The strings starting with SP represent species. The strings starting with NP represent names. The strings starting with ADP represent property. Given a context and a fact, it is required to generate a short missing fact.
\end{itemize}
Other parts of the prompts are the same as the previous form.

Further, we consider both zero-shot setting, few-shot setting and chain-of-thought strategy in NeuLR. The zero-shot and few-shot prompt forms are same as the forms in TABLE~\ref{prompt}. For chain-of-thought prompt, we utilize the explanation provided with the dataset to help generate the reasoning chains. Next, we will offer examples for each reasoning setting.

\noindent \textbf{- Deductive Reasoning}

\noindent [chain-of-thought]: There is one example of deductive reasoning: The facts are: SP1Ggwz1 are afraid of SPU4g85d. NPdmbdKB is a SP1Ggwz1. SPArMPn0 are afraid of SPU4g85d. SPU4g85d are afraid of SPArMPn0. NPd0WTGi is a SP1Ggwz1. NP1cTSnm is a SPU4g85d. SPFURodYs are afraid of SP1Ggwz1. NPpEQFEK is a SPU4g85d. The question is: What is NPpEQFEK afraid of? Because NPpEQFEK is a SPU4g85d and SPU4g85d are afraid of SPArMPn0. Therefore, the answer is: SPArMPn0.

\noindent \textbf{- Inductive Reasoning}

\noindent [chain-of-thought]: There is one example of inductive reasoning: The facts are: NP16WBQQ is a SPQFyx7i. NPerfGLN is a SP4Vge77. NP0V3buK is a SPGbXv1C. NPU9kOFg is a SPQFyx7i. NPDHuODI is a SP4Vge77. NPerfGLN is ADPHQuIP9. NP0V3buK is ADPmBU2ts. NP16WBQQ is ADP0rI59W. NPDHuODI is ADPHQuIP9.The question is: What property is NPU9kOFg? Because NPU9kOFg is a SPQFyx7i, NP16WBQQ is a SPQFyx7i and NP16WBQQ is ADP0rI59W. Therefore, the answer is: ADP0rI59W.

\noindent \textbf{- Abductive Reasoning}

\noindent [chain-of-thought]: ... Based on rule2 and tripleM, we can derive a new fact. Combine it with triple4. Based on rule1, we can derive a new fact. Combine it with rule1. In this way, we can finally derive the given fact...

\section{Supplementary Analysis}

We also provide the performances of text-davinci-003 and BARD with different number of hops in Figure~\ref{num_hop_appendix}. In the deductive setting of text-davinci-003, model performance drops with the number of hops increasing. But when the hop number is over five, it witnesses slight gains in performance, which illustrates that text-davinci-003 has the potential to conduct multi-hop reasoning in the deductive reasoning setting. However, in the inductive and abductive settings, the performances of text-davinci-003 decrease sharply when the number of hops increases. Especially, it fails all the cases when the hop number is over six in inductive reasoning, and also when the hop number is greater than one in abductive reasoning. It is inferior to ChatGPT.

For BARD, the situation is quite different. In deductive reasoning, the performance of BARD increases with the hop number adds. Especially, when the hop number is over five, the accuracy reaches 100\% without reducing the rigor of reasoning. In inductive reasoning, the performance of BARD also drops at first, but it keeps stable and witnesses obvious gains when the hop number is over six. It demonstrates that BARD is better at conducting inductive reasoning and processing multi-hop scenarios compared with text-davinci-003 and ChatGPT. In abductive reasoning, BARD struggles a lot, inferior to ChatGPT but is better than text-davinci-003.

In all, in the face of complex multi-hop scenarios, LLMs still have much room for improvement. From the results, they do relatively well in deductive reasoning settings. But they are far from good in the inductive and abductive settings, which can also inspire future research on it.

\section{Case Studies}
\vspace{-0.1cm}
We show one reasoning case for each dataset in Table 11-25, where the context and question as well as output of 0-shot ChatGPT, 1-shot ChatGPT, 3-shot ChatGPT,
0-shot Davinci-003 and 0-shot BARD are displayed.
We also provided the annotated information about the answer correctness: \textcolor{red}{\faCheckCircle}, explain correctness: \textcolor{cyan}{\faCheckCircle}, explain completeness: \textcolor{teal}{\faCheckCircle}, explain redundancy: \textcolor{purple}{\faCheckCircle}, evidence wrong selection: \textcolor{green}{\usym{1F5F9}}, hallucination: \textcolor{blue}{\usym{1F5F9}}, no reasoning: \textcolor{brown}{\usym{1F5F9}}, perspective mistake: \textcolor{magenta}{\usym{1F5F9}}, and process mistake: \textcolor{violet}{\usym{1F5F9}}.  (the last five indicators are annotated when the explain explanation is false).

\vspace{-0.3cm}


\onecolumn
{\small
\begin{longtable}{c|m{8cm}<{\centering}|m{6cm}<{\centering}}
\caption{Prompt Engineering.}\\
		\toprule
            \textbf{Dataset} & \textbf{Prompt of zero-shot} & \textbf{Prompt of k-shot} \\
		\hline
            \hline
            \multicolumn{3}{c}{\textbf{Deductive Reasoning}} \\
            \hline
            bAbI (task 15)
            &\makecell[{{p{8cm}}}]{\textcolor{red}{[zero-shot prompt]}: Given facts: [Context]. Based on the given facts above, answer the following question using deductive reasoning and give simple explanations.
            The question is: [Question]}
            &\makecell[{{p{6cm}}}]{There are [k] examples of deductive reasoning:\\
            Given facts: [Context]\\
            The question is: [Question]\\
            The answer is: [Answer] (display k samples)\\
            \textcolor{red}{[zero-shot prompt]}
            }\\
            \hline
            EntailmentBank
            &\makecell[{{p{8cm}}}]{\textcolor{red}{[zero-shot prompt]}: Given facts: [Context]. [Question]. Please answer the question in one sentence using deductive reasoning. And give simple explanations.}
            &\makecell[{{p{6cm}}}]{There are [k] examples of deductive reasoning:\\
            Given facts: [Context]\\
            The question is: [Question]\\
            The answer is: [Answer] (display k samples)\\
            \textcolor{red}{[zero-shot prompt]}
            }\\
            \hline
            RuleTaker
            &\makecell[{{p{8cm}}}]{\textcolor{red}{[zero-shot prompt]}: Given facts: [Context]. Based on the given facts above, determine whether the following statement is true using deductive reasoning and give simple explanations. The statement is: [Statement].}
            &\makecell[{{p{6cm}}}]{There are [k] examples of deductive reasoning:\\
            Given facts: [Context]\\
            The statement is: [Statement]\\
            The answer is: [Answer] (display k samples)\\
            \textcolor{red}{[zero-shot prompt]}
            }\\
            \hline
            FOLIO
            &\makecell[{{p{8cm}}}]{\textcolor{red}{[zero-shot prompt]}: Given facts: [Context]. Based on the given facts above, determine whether the following statement is true, false, or uncertain using deductive reasoning and give simple explanations. The statement is: [Statement].}
            &\makecell[{{p{6cm}}}]{There are [k] examples of deductive reasoning:\\
            Given facts: [Context]\\
            The statement is: [Statement]\\
            The answer is: [Answer] (display k samples)\\
            \textcolor{red}{[zero-shot prompt]}
            }\\
            \hline
            Leap-Of-Thought 
            &\makecell[{{p{8cm}}}]{\textcolor{red}{[zero-shot prompt]}: Given facts: [Context]. Based on the given facts above, determine whether the following statement is true using deductive reasoning and give simple explanations. The statement is: [Statement].}
            &\makecell[{{p{6cm}}}]{There are [k] examples of deductive reasoning:\\
            Given facts: [Context]\\
            The statement is: [Statement]\\
            The answer is: [Answer] (display k samples)\\
            \textcolor{red}{[zero-shot prompt]}
            }\\
            \hline
            \hline
            \multicolumn{3}{c}{\textbf{Inductive Reasoning}} \\
            \hline
            bAbI-16
            &\makecell[{{p{8cm}}}]{\textcolor{red}{[zero-shot prompt]}: Given facts: [context]. Based on the given facts above, answer the following question using inductive reasoning and give simple explanations. The question is: [Question].}
            &\makecell[{{p{6cm}}}]{There are [k] examples of inductive reasoning:\\
            Given facts: [Context]\\
            The question is: [Question]\\
            The answer is: [Answer] (display k samples)\\
            \textcolor{red}{[zero-shot prompt]}
            }\\
            \hline
            CLUTRR
            &\makecell[{{p{8cm}}}]{\textcolor{red}{[zero-shot prompt]}: Given facts: [context]. [Question]. Please answer the question in one sentence using inductive reasoning. And give simple explanations.}
            &\makecell[{{p{6cm}}}]{There are [k] examples of inductive reasoning:\\
            Given facts: [Context]\\
            The question is: [Question]\\
            The answer is: [Answer] (display k samples)\\
            \textcolor{red}{[zero-shot prompt]}
            }\\
            \hline
            \hline
            \multicolumn{3}{c}{\textbf{Abductive Reasoning}} \\
            \hline
            $\alpha$-NLI
            &\makecell[{{p{8cm}}}]{\textcolor{red}{[task description]}: Given a context, the abductive reasoning task is to choose the more likely explanation from a given pair of hypotheses choices. And give simple explanations.\\
            \textcolor{red}{[zero-shot prompt]}: The context is: [Context]. The hypothesis choices are: A. [Option A]. B. [Option B].}
            &\makecell[{{p{6cm}}}]{\textcolor{red}{[task description]} \\Next, I will give you [k] example(s) for test. The context is [Context]. The hypothesis choice are: A. [Option A]. B. [Option B]. The correct choice is: [Label]. \\
            Next, I will give you an example for test.\\
            \textcolor{red}{[zero-shot prompt]}
            }\\
            \hline
            $\alpha$-NLG
            &\makecell[{{p{8cm}}}]{\textcolor{red}{[task description]}: Given a context, the abductive reasoning task is to generate a valid and short hypothesis.\\ 
            \textcolor{red}{[zero-shot prompt]}: The context is: [Context]. Please generate a short hypothesis for the context and give simple explanations.}
            &\makecell[{{p{6cm}}}]{\textcolor{red}{[task description]} \\Next, I will give you [k] example(s) for test. The context is [Context]. The correct answer is [Label].\\
            Next, I will give you an example for test.\\
            \textcolor{red}{[zero-shot prompt]}
            }\\
            \hline
            AbductiveRules
            &\makecell[{{p{8cm}}}]{
            \textcolor{red}{[task description]}: Given a context and an observation, the abductive reasoning task is to generate a valid and short explanation.\\
            \textcolor{red}{[zero-shot prompt]}: The context is: [Context]. The observation is: [Observation]. Please generate a short explanation for the given context and observation.}
            &\makecell[{{p{6cm}}}]{\textcolor{red}{[task description]} \\Next, I will give you [k] example(s) for test. The context is: [Context]. The observation is: [Observation]. The explanation is: [Explanation].\\
            Next, I will give you an example for test.\\
            \textcolor{red}{[zero-shot prompt]}
            }\\
            \hline
            D*-Ab
            &\makecell[{{p{8cm}}}]{\textcolor{red}{[task description]}: Given a context and a fact, the abductive reasoning task is to generate a short missing fact.\\
            \textcolor{red}{[zero-shot prompt]}: The context is: [Context+Rule]. The fact is: [Fact]. Please generate a short missing fact for the given context and fact. And give simple explanations.}
            &\makecell[{{p{6cm}}}]{\textcolor{red}{[task description]} \\Next, I will give you [k] example(s) for test. The context is: [Context+Rule]. The observation is: [Observation]. The explanation is: [Explanation].\\
            Next, I will give you an example for test.\\
            \textcolor{red}{[zero-shot prompt]}
            }\\
            \hline
            \hline
            \multicolumn{3}{c}{\textbf{mixed-form Reasoning}} \\
            \hline
            ReClor
            &\makecell[{{p{8cm}}}]{
            \textcolor{red}{[task description]}: This is a Mahcine Reading Comprehension task, given the context and question, you are required to choose the correct answer from the answer set and give explanations.\\
            \textcolor{red}{[zero-shot prompt]}: The context is: [Context]. The question is: [Question]. [Option A]. [Option B]. [Option C]. [Option D]. Please give the correct answer and simple explanations.}
            &\makecell[{{p{6cm}}}]{\textcolor{red}{[task description]} \\Next, I will give you [k] example(s) for test. The context is: [Context]. The question is: [Question]. The correct choice is: [Label].\\
            Next, I will give you an example for test.\\
            \textcolor{red}{[zero-shot prompt]}
            }\\
            \hline
            LogiQA
            &\makecell[{{p{8cm}}}]{
            \textcolor{red}{[task description]}: This is a Mahcine Reading Comprehension task, given the context and question, you are required to choose the correct answer from the answer set and give explanations.\\
            \textcolor{red}{[zero-shot prompt]}: The context is: [Context]. The question is: [Question]. [Option A]. [Option B]. [Option C]. [Option D]. Please give the correct answer and simple explanations.}
            &\makecell[{{p{6cm}}}]{\textcolor{red}{[task description]} \\Next, I will give you [k] example(s) for test. The context is: [Context]. The question is: [Question]. The correct choice is: [Label].\\
            Next, I will give you an example for test.\\
            \textcolor{red}{[zero-shot prompt]}
            }\\
            \hline
            LogiQA 2.0
            &\makecell[{{p{8cm}}}]{
            \textcolor{red}{[task description]}: This is a Mahcine Reading Comprehension task, given the context and question, you are required to choose the correct answer from the answer set and give explanations.\\
            \textcolor{red}{[zero-shot prompt]}: The context is: [Context]. The question is: [Question]. [Option A]. [Option B]. [Option C]. [Option D]. Please give the correct answer and simple explanations.}
            &\makecell[{{p{6cm}}}]{\textcolor{red}{[task description]} \\Next, I will give you [k] example(s) for test. The context is: [Context]. The question is: [Question]. The correct choice is: [Label].\\
            Next, I will give you an example for test.\\
            \textcolor{red}{[zero-shot prompt]}
            }\\
            \hline
            LogiQA2NLI
            &\makecell[{{p{8cm}}}]{
            \textcolor{red}{[task description]}: This is a Natural Language Inference task. Please tell whether the premise and conclusion entailed or NOT entailed. And give simple explanations.\\
            \textcolor{red}{[zero-shot prompt]}: The premise is: [Premise]. The conclusion is: [Conclusion]. Please give the correct answer and simple explanations.}
            &\makecell[{{p{6cm}}}]{\textcolor{red}{[task description]} \\Next, I will give you [k] example(s) for test. The premise is: [Premise]. The context is: [Context]. The correct answer is: [Label].\\
            Next, I will give you an example for test.\\
            \textcolor{red}{[zero-shot prompt]}
            }\\
		\bottomrule
            \label{prompt}
	\end{longtable}
}
\twocolumn

\onecolumn
{\small
\begin{longtable}{m{0.5cm}|l}
\caption{A reasoning case in bAbI-15. }\label{tab:long}\\
		\toprule
            \rotatebox{90}{\makecell[c]{Question, answer\\and explanation}} & \makecell[{{p{16.3cm}}}]{Context: 1 Sheep are afraid of wolves. 2  Mice are afraid of wolves. 3  Winona is a sheep. 4  Gertrude is a sheep. 5  Wolves are afraid of sheep. 6  Emily is a sheep. 7  Cats are afraid of sheep. 8  Jessica is a sheep.\\
            Question: What is jessica afraid of?\\
            Answer: \textcolor{red}{Wolf}\\
            Explanation: IsSheep(jessica) (8)$\xrightarrow{1}$ Afraid(jessica, wolf)
            }\\
            \hline
            \rotatebox{90}{\makecell[c]{ChatGPT\\of 0-shot}}& \makecell[{{p{16cm}}}]{
            Jessica is afraid of \textcolor{red}{nothing}, as it is not specified in the given facts that she has any known natural predator or fear. \\
            Answer and Explain: \textcolor{red}{\faTimesCircle} \textcolor{cyan}{\faTimesCircle} \textcolor{teal}{\faTimesCircle} \textcolor{purple}{\faTimesCircle}$\;\;\;\;\;\;\;\;$Explain Error Type: \textcolor{green}{\usym{1F5F9}} \textcolor{blue}{\usym{1F5F5}} \textcolor{brown}{\usym{1F5F9}} \textcolor{magenta}{\usym{1F5F5}}  \textcolor{violet}{\usym{1F5F5}}
            }\\
            \hline
            \rotatebox{90}{\makecell[c]{ChatGPT\\of 1-shot}}& \makecell[{{p{16cm}}}]{
            Jessica is a sheep and sheep are afraid of \textcolor{red}{wolves}. Therefore, it can be deduced that Jessica is afraid of wolves.\\
            Answer and Explain: \textcolor{red}{\faCheckCircle} \textcolor{cyan}{\faCheckCircle} \textcolor{teal}{\faCheckCircle} \textcolor{purple}{\faTimesCircle}
            }\\
            \hline
            \rotatebox{90}{\makecell[c]{ChatGPT\\of 3-shot}}& \makecell[{{p{16cm}}}]{
            Sheep are afraid of wolves and Jessica is a sheep.\\
            Answer and Explain: \textcolor{red}{\faTimesCircle} \textcolor{cyan}{\faTimesCircle} \textcolor{teal}{\faCheckCircle} \textcolor{purple}{\faTimesCircle}$\;\;\;\;\;\;\;\;$Explain Error Type: \textcolor{green}{\usym{1F5F5}} \textcolor{blue}{\usym{1F5F5}} \textcolor{brown}{\usym{1F5F9}} \textcolor{magenta}{\usym{1F5F5}}  \textcolor{violet}{\usym{1F5F5}}
            }\\
            \hline
            \rotatebox{90}{\makecell[c]{Davinci-003\\of 0-shot}}& \makecell[{{p{16cm}}}]{
            Jessica is likely afraid of \textcolor{red}{wolves}, since wolves are a common source of fear that other animals, including sheep, \textcolor{blue}{mice, and cats, have in common}. \textcolor{purple}{Wolves are known to be predators of sheep}, so Jessica being a sheep, would likely be scared of wolves.\\
            Answer and Explain: \textcolor{red}{\faCheckCircle} \textcolor{cyan}{\faTimesCircle} \textcolor{teal}{\faCheckCircle} \textcolor{purple}{\faCheckCircle}$\;\;\;\;\;\;\;\;$Explain Error Type: \textcolor{green}{\usym{1F5F9}} \textcolor{blue}{\usym{1F5F9}} \textcolor{brown}{\usym{1F5F5}} \textcolor{magenta}{\usym{1F5F9}}  \textcolor{violet}{\usym{1F5F5}}
            }\\
            \hline
            \rotatebox{90}{\makecell[c]{Bard\\of 0-shot}}& \makecell[{{p{16cm}}}]{
            Jessica is a sheep. Sheep are afraid of wolves. Therefore, Jessica is afraid of wolves.\\
            Here is the deductive reasoning:\\
            1. Sheep are afraid of wolves.\\
            2. Jessica is a sheep.\\
            3. Therefore, Jessica is afraid of wolves.\\
            The first premise is supported by the fact that \textcolor{purple}{sheep are prey animals and wolves are predators. Wolves are known to prey on sheep, and sheep have evolved to be afraid of wolves as a way of self-preservation}.\\
            The second premise is supported by the fact that Jessica is a sheep. This is a given fact in the prompt.\\
            The third premise follows logically from the first two premises. If sheep are afraid of wolves, and Jessica is a sheep, then Jessica must be afraid of wolves.\\
            Therefore, the conclusion is that Jessica is afraid of wolves.\\
            Answer and Explain: \textcolor{red}{\faCheckCircle} \textcolor{cyan}{\faCheckCircle} \textcolor{teal}{\faCheckCircle} \textcolor{purple}{\faCheckCircle}
            }\\
		\bottomrule
	\end{longtable}
}
\twocolumn

\onecolumn
{\small
\begin{longtable}{m{0.5cm}|l}
\caption{A reasoning case in EntailmentBank. }\label{tab:long}\\
		\toprule
            \rotatebox{90}{\makecell[c]{Question, answer\\and explanation}} & \makecell[{{p{16.3cm}}}]{Context: 1 pollination is when pollinating animals / wind / water carry pollen from one flower to another flower. 2  if something is a part of a process then that something is required for that process. 3  helping something has a positive impact on that something. 4  if something is required for a process then that something positively impacts that process. 5  an insect is a pollinating animal.\\
            Question: life cycle of some plants includes the process of pollination. Pollination is helped by?\\
            Answer: \textcolor{red}{wind, water, and insects}\\
            Explanation: sent1 \& sent5$\rightarrow$int1: pollination is when insects / wind / water carry pollen from one flower to another; int1 \& sent2$\rightarrow$int2: insects / wind / water carrying pollen is required for pollination; int2 \& sent4$\rightarrow$int3: insects / wind / water have a positive impact on pollination by carrying pollen; int3 \& sent3$\rightarrow$hypothesis
            }\\
            \hline
            \rotatebox{90}{\makecell[c]{ChatGPT\\of 0-shot}}& \makecell[{{p{16cm}}}]{
            Pollination in the life cycle of some plants is helped by \textcolor{red}{pollinating animals, such as insects}, which are required for the process of pollination and have a positive impact on it. \\
            Answer and Explain: \textcolor{red}{\faTimesCircle} \textcolor{cyan}{\faTimesCircle} \textcolor{teal}{\faTimesCircle} \textcolor{purple}{\faTimesCircle}$\;\;\;\;\;\;\;\;$Explain Error Type: \textcolor{green}{\usym{1F5F9}} \textcolor{blue}{\usym{1F5F5}} \textcolor{brown}{\usym{1F5F9}} \textcolor{magenta}{\usym{1F5F5}}  \textcolor{violet}{\usym{1F5F5}}
            }\\
            \hline
            \rotatebox{90}{\makecell[c]{ChatGPT\\of 1-shot}}& \makecell[{{p{16cm}}}]{
            Pollination in plants is helped by \textcolor{red}{pollinating animals such as insects}, which carry pollen from one flower to another as a necessary part of the process, positively impacting the plants' life cycle.\\
            Answer and Explain: \textcolor{red}{\faTimesCircle} \textcolor{cyan}{\faTimesCircle} \textcolor{teal}{\faTimesCircle} \textcolor{purple}{\faTimesCircle}$\;\;\;\;\;\;\;\;$Explain Error Type: \textcolor{green}{\usym{1F5F9}} \textcolor{blue}{\usym{1F5F5}} \textcolor{brown}{\usym{1F5F9}} \textcolor{magenta}{\usym{1F5F5}}  \textcolor{violet}{\usym{1F5F5}}
            }\\
            \hline
            \rotatebox{90}{\makecell[c]{ChatGPT\\of 3-shot}}& \makecell[{{p{16cm}}}]{
            Pollinating  animals such as insects, which are required for the process of pollination to occur.\\
            Answer and Explain: \textcolor{red}{\faTimesCircle} \textcolor{cyan}{\faTimesCircle} \textcolor{teal}{\faTimesCircle} \textcolor{purple}{\faTimesCircle}$\;\;\;\;\;\;\;\;$Explain Error Type: \textcolor{green}{\usym{1F5F9}} \textcolor{blue}{\usym{1F5F5}} \textcolor{brown}{\usym{1F5F9}} \textcolor{magenta}{\usym{1F5F5}}  \textcolor{violet}{\usym{1F5F5}}
            }\\
            \hline
            \rotatebox{90}{\makecell[c]{Davinci-003\\of 0-shot}}& \makecell[{{p{16cm}}}]{
            Pollination is \textcolor{red}{helped by pollinating animals, such as insects}, carrying pollen from one flower to another.\\
            Answer and Explain: \textcolor{red}{\faTimesCircle} \textcolor{cyan}{\faTimesCircle} \textcolor{teal}{\faTimesCircle} \textcolor{purple}{\faTimesCircle}$\;\;\;\;\;\;\;\;$Explain Error Type: \textcolor{green}{\usym{1F5F9}} \textcolor{blue}{\usym{1F5F5}} \textcolor{brown}{\usym{1F5F9}} \textcolor{magenta}{\usym{1F5F5}}  \textcolor{violet}{\usym{1F5F5}}
            }\\
            \hline
            \rotatebox{90}{\makecell[c]{Bard\\of 0-shot}}& \makecell[{{p{16cm}}}]{
            Sure, here is the answer to the question in one sentence using deductive reasoning:\\
            Pollination is helped by \textcolor{red}{pollinating animals, such as insects}.\\
            Here is the simple explanation:\\
            \textcolor{purple}{Pollination is the process of transferring pollen from the male part of a flower to the female part of a flower.}\\
            Pollinating animals, such as insects, help pollination by carrying pollen from one flower to another.\\
            \textcolor{purple}{Without pollination, plants cannot reproduce and produce seeds.}\\
            Therefore, pollinating animals help the life cycle of some plants by helping them to reproduce.\\
            Answer and Explain: \textcolor{red}{\faTimesCircle} \textcolor{cyan}{\faTimesCircle} \textcolor{teal}{\faTimesCircle} \textcolor{purple}{\faCheckCircle}$\;\;\;\;\;\;\;\;$Explain Error Type: \textcolor{green}{\usym{1F5F9}} \textcolor{blue}{\usym{1F5F5}} \textcolor{brown}{\usym{1F5F9}} \textcolor{magenta}{\usym{1F5F5}}  \textcolor{violet}{\usym{1F5F5}}
            }\\
		\bottomrule
	\end{longtable}
}
\twocolumn

\onecolumn
{\small
\begin{longtable}{m{0.5cm}|l}
\caption{A reasoning case in RuleTaker. }\label{tab:long}\\
		\toprule
            \rotatebox{90}{\makecell[c]{Question, answer\\and explanation}} & \makecell[{{p{16.3cm}}}]{Context: 1 If the current runs through the bell then the bell is ringing. 2 The circuit includes the battery. 3 If the circuit does not include the switch then the circuit is complete. 4 The circuit includes the bell. 5 If the circuit includes the switch and the switch is on then the circuit is complete. 6 If the current runs through the light bulb then the light bulb is glowing. 7 If the wire is plastic then the wire is not conducting. 8 The circuit includes the switch. 9 If the circuit includes the battery and the battery is not flat then the circuit is powered. 10 If the current runs through the circuit and the circuit includes the bell then the current runs through the bell. 11 The wire is metal. 12 If the wire is metal then the wire is conducting. 13 If the current runs through the circuit and the circuit includes the light bulb then the current runs through the light bulb. 14 If the current runs through the radio then the radio is playing. 15 If the current runs through the circuit and the circuit includes the radio then the current runs through the radio. 16 If the circuit is powered and the circuit is complete and the wire is conducting then the current runs through the circuit.\\
            Statement: The current does not run through the circuit.\\
            Question: Determine whether the statement is true using deductive reasoning and give explanations.\\
            Answer: \textcolor{red}{True}  
            }\\
            \hline
            \rotatebox{90}{\makecell[c]{ChatGPT\\of 0-shot}}& \makecell[{{p{16cm}}}]{
            The circuit is powered, complete, and the wire is conducting. Therefore, we can use deductive reasoning to determine that the statement ``The current does not run through the circuit" is false based on the given facts. \\
            Answer and Explain: \textcolor{red}{\faTimesCircle} \textcolor{cyan}{\faTimesCircle} \textcolor{teal}{\faTimesCircle} \textcolor{purple}{\faTimesCircle}$\;\;\;\;\;\;\;\;$Explain Error Type: \textcolor{green}{\usym{1F5F9}} \textcolor{blue}{\usym{1F5F5}} \textcolor{brown}{\usym{1F5F9}} \textcolor{magenta}{\usym{1F5F5}}  \textcolor{violet}{\usym{1F5F5}}
            }\\
            \hline
            \rotatebox{90}{\makecell[c]{ChatGPT\\of 1-shot}}& \makecell[{{p{16cm}}}]{
            If the battery is not flat and the circuit includes the battery, then the circuit is powered: ``If the circuit includes the battery and the battery is not flat then the circuit is powered." If the circuit is powered and the wire is metal (which is conducting), then the current runs through the circuit: ``If the circuit is powered and the circuit is complete and the wire is conducting then the current runs through the circuit." However, it is stated that the wire is plastic and if the wire is plastic, then it is not conducting: ``If the wire is plastic then the wire is not conducting." Therefore, the statement ``The current does not run through the circuit" is true.\\
            Answer and Explain: \textcolor{red}{\faTimesCircle} \textcolor{cyan}{\faTimesCircle} \textcolor{teal}{\faTimesCircle} \textcolor{purple}{\faTimesCircle}$\;\;\;\;\;\;\;\;$Explain Error Type: \textcolor{green}{\usym{1F5F9}} \textcolor{blue}{\usym{1F5F5}} \textcolor{brown}{\usym{1F5F9}} \textcolor{magenta}{\usym{1F5F5}}  \textcolor{violet}{\usym{1F5F5}}
            }\\
            \hline
            \rotatebox{90}{\makecell[c]{ChatGPT\\of 3-shot}}& \makecell[{{p{16cm}}}]{
            This statement is false based on the given facts. If the circuit is powered, complete, and the wire is conducting, then the current runs through the circuit, as stated in the first premise. Hence, we can't conclude that the current doesn't run through the circuit.\\
            Answer and Explain: \textcolor{red}{\faTimesCircle} \textcolor{cyan}{\faTimesCircle} \textcolor{teal}{\faTimesCircle} \textcolor{purple}{\faTimesCircle}$\;\;\;\;\;\;\;\;$Explain Error Type: \textcolor{green}{\usym{1F5F9}} \textcolor{blue}{\usym{1F5F5}} \textcolor{brown}{\usym{1F5F9}} \textcolor{magenta}{\usym{1F5F5}}  \textcolor{violet}{\usym{1F5F5}}
            }\\
            \hline
            \rotatebox{90}{\makecell[c]{Davinci-003\\of 0-shot}}& \makecell[{{p{16cm}}}]{
            Pollination is \textcolor{red}{helped by pollinating animals, such as insects}, carrying pollen from one flower to another.\\
            Answer and Explain: \textcolor{red}{\faTimesCircle} \textcolor{cyan}{\faTimesCircle} \textcolor{teal}{\faTimesCircle} \textcolor{purple}{\faTimesCircle}$\;\;\;\;\;\;\;\;$Explain Error Type: \textcolor{green}{\usym{1F5F9}} \textcolor{blue}{\usym{1F5F5}} \textcolor{brown}{\usym{1F5F9}} \textcolor{magenta}{\usym{1F5F5}}  \textcolor{violet}{\usym{1F5F5}}
            }\\
            \hline
            \rotatebox{90}{\makecell[c]{Bard\\of 0-shot}}& \makecell[{{p{16cm}}}]{
            Sure, here is the answer to the question in one sentence using deductive reasoning:\\
            Pollination is helped by \textcolor{red}{pollinating animals, such as insects}.\\
            Here is the simple explanation:\\
            \textcolor{purple}{Pollination is the process of transferring pollen from the male part of a flower to the female part of a flower.}\\
            Pollinating animals, such as insects, help pollination by carrying pollen from one flower to another.\\
            \textcolor{purple}{Without pollination, plants cannot reproduce and produce seeds.}\\
            Therefore, pollinating animals help the life cycle of some plants by helping them to reproduce.\\
            Answer and Explain: \textcolor{red}{\faTimesCircle} \textcolor{cyan}{\faTimesCircle} \textcolor{teal}{\faTimesCircle} \textcolor{purple}{\faCheckCircle}$\;\;\;\;\;\;\;\;$Explain Error Type: \textcolor{green}{\usym{1F5F9}} \textcolor{blue}{\usym{1F5F5}} \textcolor{brown}{\usym{1F5F9}} \textcolor{magenta}{\usym{1F5F5}}  \textcolor{violet}{\usym{1F5F5}}
            }\\
		\bottomrule
	\end{longtable}
}
\twocolumn

\onecolumn
{\small
\begin{longtable}{m{0.5cm}|l}
\caption{A reasoning case in FOLIO.}\label{tab:long}\\
		\toprule
            \rotatebox{90}{\makecell[c]{Question, answer\\and explanation}} & \makecell[{{p{16.3cm}}}]{Context: 1 All growth companies' stocks are volatile. 2 If the stock price is volatile, then it is not suitable for a retirement fund. 3 Some companies' stocks are growth companies' stocks. 4 All mature companies' stocks are suitable for a retirement fund. 5 KO is a mature company's stock.\\
            Statement: KO is a company stock and a growth company's stock.\\
            Question: Determine whether the statement is true, false, or uncertain using deductive reasoning and give explanations.\\
            Answer: \textcolor{red}{False}\\
            Explanation: MatureCompanies'Stocks(KO) (5)$\xrightarrow{4}$ SuitableForRetirementRund(KO) $\xrightarrow{2r}$$\neg$PriceVolatile(KO)$\xrightarrow{1r}$ $\neg$GrowthCompanies'Stocks(KO) (contradictory)  
            }\\
            \hline
            \rotatebox{90}{\makecell[c]{ChatGPT\\of 0-shot}}& \makecell[{{p{16cm}}}]{
            A mature company's stock cannot also be a growth company's stock since these are two distinct categories. The given facts say that \textcolor{purple}{some companies' stocks are growth companies' stocks} and all mature companies' stocks are suitable for a retirement fund. Additionally, KO is specifically identified as a mature company's stock. Therefore, the statement ``KO is a company stock and a growth company's stock" is \textcolor{red}{false}.\\
            Answer and Explain: \textcolor{red}{\faCheckCircle} \textcolor{cyan}{\faTimesCircle} \textcolor{teal}{\faTimesCircle} \textcolor{purple}{\faCheckCircle}$\;\;\;\;\;\;\;\;$Explain Error Type: \textcolor{green}{\usym{1F5F9}} \textcolor{blue}{\usym{1F5F5}} \textcolor{brown}{\usym{1F5F9}} \textcolor{magenta}{\usym{1F5F5}}  \textcolor{violet}{\usym{1F5F5}}
            }\\
            \hline
            \rotatebox{90}{\makecell[c]{ChatGPT\\of 1-shot}}& \makecell[{{p{16cm}}}]{
            We know from the given facts that KO is ``a mature company's stock," but there is no information to suggest that it is also a growth company's stock. Therefore, the statement ``KO is a company stock and a growth company's stock" is \textcolor{red}{uncertain} based on the given facts.\\
            Answer and Explain: \textcolor{red}{\faTimesCircle} \textcolor{cyan}{\faTimesCircle} \textcolor{teal}{\faTimesCircle} \textcolor{purple}{\faTimesCircle}$\;\;\;\;\;\;\;\;$Explain Error Type: \textcolor{green}{\usym{1F5F9}} \textcolor{blue}{\usym{1F5F5}} \textcolor{brown}{\usym{1F5F9}} \textcolor{magenta}{\usym{1F5F5}}  \textcolor{violet}{\usym{1F5F5}}
            }\\
            \hline
            \rotatebox{90}{\makecell[c]{ChatGPT\\of 3-shot}}& \makecell[{{p{16cm}}}]{
            The given facts state that KO is a mature company's stock. However, it is not stated whether KO is also a growth company's stock. Therefore, the statement ``KO is a company stock and a growth company's stock" is \textcolor{red}{uncertain}.\\
            Answer and Explain: \textcolor{red}{\faTimesCircle} \textcolor{cyan}{\faTimesCircle} \textcolor{teal}{\faTimesCircle} \textcolor{purple}{\faTimesCircle}$\;\;\;\;\;\;\;\;$Explain Error Type: \textcolor{green}{\usym{1F5F9}} \textcolor{blue}{\usym{1F5F5}} \textcolor{brown}{\usym{1F5F9}} \textcolor{magenta}{\usym{1F5F5}}  \textcolor{violet}{\usym{1F5F5}}
            }\\
            \hline
            \rotatebox{90}{\makecell[c]{Davinci-003\\of 0-shot}}& \makecell[{{p{16cm}}}]{
            \textcolor{red}{False}. While it is true that KO is a company stock, it is not true that it is a growth company's stock. Growth companies' stocks are volatile, whereas the fact given states that mature companies' stocks are suitable for retirement funds. Since KO is a mature company's stock, it follows that it is not a growth company's stock.\\
            Answer and Explain: \textcolor{red}{\faCheckCircle} \textcolor{cyan}{\faTimesCircle} \textcolor{teal}{\faTimesCircle} \textcolor{purple}{\faTimesCircle}$\;\;\;\;\;\;\;\;$Explain Error Type: \textcolor{green}{\usym{1F5F9}} \textcolor{blue}{\usym{1F5F5}} \textcolor{brown}{\usym{1F5F9}} \textcolor{magenta}{\usym{1F5F5}}  \textcolor{violet}{\usym{1F5F5}}
            }\\
            \hline
            \rotatebox{90}{\makecell[c]{Bard\\of 0-shot}}& \makecell[{{p{16cm}}}]{
            The statement ``KO is a company stock and a growth company's stock" is \textcolor{red}{false}.\\
            Explanation:\\
            1. All growth companies' stocks are volatile. This means that the stock price of a growth company is likely to fluctuate more than the stock price of a mature company.\\
            2. If the stock price is volatile, then it is not suitable for a retirement fund. This is because retirement funds are designed to provide long-term growth, and volatile stocks are more likely to lose value over time.\\
            3. Some companies' stocks are growth companies' stocks. This means that some companies are growing at a faster rate than others.\\
            4. All mature companies' stocks are suitable for a retirement fund. This is because mature companies are less likely to experience sudden changes in their stock price.\\
            5. KO is a mature company's stock. This means that KO is a company that is not growing at a rapid pace.\\
            Based on the above facts, we can conclude that KO is not a growth company's stock. This is because KO is a mature company, and mature companies' stocks are not volatile. Therefore, the statement ``KO is a company stock and a growth company's stock" is false.\\
            Answer and Explain: \textcolor{red}{\faCheckCircle} \textcolor{cyan}{\faCheckCircle} \textcolor{teal}{\faTimesCircle} \textcolor{purple}{\faTimesCircle}
            }\\
		\bottomrule
	\end{longtable}
}
\twocolumn

\onecolumn
{\small
\begin{longtable}{m{0.5cm}|l}
\caption{A reasoning case in Leap-Of-Thought. }\label{tab:long}\\
		\toprule
            \rotatebox{90}{\makecell[c]{Question, answer\\and explanation}} & \makecell[{{p{16.3cm}}}]{Context: 1 A bird is not capable of express feelings. 2 A quack-quack is not capable of suffering dehydration. 3 A privet is capable of suffering dehydration. 4 A music is capable of express feelings.\\
            Statement: A jazz is capable of express feelings.\\
            Question: Determine whether the statement is true using deductive reasoning and give explanations.\\
            Answer: \textcolor{red}{True}\\
            Explanation: IsMusic(jazz) (common sense)$\xrightarrow{4}$ ExpressFeeling(jazz)
            }\\
            \hline
            \rotatebox{90}{\makecell[c]{ChatGPT\\of 0-shot}}& \makecell[{{p{16cm}}}]{
            The statement ``A jazz is capable of expressing feelings" can be inferred with a certain level of probability based on the given facts. The fact that "A music is capable of expressing feelings" suggests that jazz, which is a genre of music, is likely to be capable of expressing feelings as well. The fact that "A bird is not capable of expressing feelings" and "A quack-quack is not capable of suffering dehydration" are logically irrelevant to the statement about whether jazz is capable of expressing feelings. The fact that "A privet is capable of suffering dehydration" is also irrelevant, as it does not provide any information or indication one way or the other about the statement. Therefore, based on the given facts, the statement about jazz being capable of expressing feelings can be inferred with a certain level of probability and is likely \textcolor{red}{true}.\\
            Answer and Explain: \textcolor{red}{\faCheckCircle} \textcolor{cyan}{\faCheckCircle} \textcolor{teal}{\faCheckCircle} \textcolor{purple}{\faTimesCircle}
            }\\
            \hline
            \rotatebox{90}{\makecell[c]{ChatGPT\\of 1-shot}}& \makecell[{{p{16cm}}}]{
            The statement ``A jazz is capable of express feelings" is \textcolor{red}{true} using deductive reasoning based on the given facts.\\
            Answer and Explain: \textcolor{red}{\faCheckCircle} \textcolor{cyan}{\faTimesCircle} \textcolor{teal}{\faTimesCircle} \textcolor{purple}{\faTimesCircle}$\;\;\;\;\;\;\;\;$Explain Error Type: \textcolor{green}{\usym{1F5F9}} \textcolor{blue}{\usym{1F5F5}} \textcolor{brown}{\usym{1F5F9}} \textcolor{magenta}{\usym{1F5F5}}  \textcolor{violet}{\usym{1F5F5}}
            }\\
            \hline
            \rotatebox{90}{\makecell[c]{ChatGPT\\of 3-shot}}& \makecell[{{p{16cm}}}]{
            This is \textcolor{red}{true}, as the given fact ``a music is capable of express feelings" includes the subcategory of music that is jazz. Therefore, we can logically deduce that jazz music is also capable of expressing feelings.\\
            Answer and Explain: \textcolor{red}{\faCheckCircle} \textcolor{cyan}{\faCheckCircle} \textcolor{teal}{\faCheckCircle} \textcolor{purple}{\faTimesCircle}
            }\\
            \hline
            \rotatebox{90}{\makecell[c]{Davinci-003\\of 0-shot}}& \makecell[{{p{16cm}}}]{
            \textcolor{red}{True}. \textcolor{purple}{Although humans usually associate the sound of a quack-quack with birds, the quack-quack mentioned here is not a bird and is not capable of expressing feelings. Additionally, a privet is capable of suffering dehydration but not expressing feelings.} Therefore, the only species that has been mentioned in the facts that is capable of expressing feelings is the music. Although a jazz is not explicitly mentioned in the facts, it would also be categorically considered to be a type of music, so it can logically be assumed that a jazz is capable of expressing feelings.\\
            Answer and Explain: \textcolor{red}{\faCheckCircle} \textcolor{cyan}{\faCheckCircle} \textcolor{teal}{\faCheckCircle} \textcolor{purple}{\faCheckCircle}
            }\\
            \hline
            \rotatebox{90}{\makecell[c]{Bard\\of 0-shot}}& \makecell[{{p{16cm}}}]{
            The statement "A jazz is capable of express feelings" is \textcolor{red}{not true}. The fact that a music is capable of expressing feelings does not mean that a jazz is capable of expressing feelings. A jazz is a type of music, but it is not the only type of music. There are many other types of music, and \textcolor{blue}{not all of them are capable of expressing feelings}. For example, \textcolor{purple}{classical music is often seen as being more formal and less emotional than jazz.} Therefore, it is not possible to say that all music is capable of expressing feelings, simply because one type of music is capable of doing so.\\
            In order to determine whether a jazz is capable of expressing feelings, we would need more information about the specific jazz in question. We would need to know what the jazz is about, who wrote it, and how it was performed. Once we have this information, we can then use deductive reasoning to determine whether the jazz is capable of expressing feelings.\\
            For example, if the jazz is about a personal experience that the author had, then it is more likely that the jazz will be able to express feelings. This is because the author will have put their own emotions into the jazz, and this will be reflected in the music. However, if the jazz is about a more general topic, such as the beauty of nature, then it is less likely that the jazz will be able to express feelings. This is because the author will not have put their own emotions into the jazz, and the music will therefore be more objective.\\
            Ultimately, whether or not a jazz is capable of expressing feelings is a matter of opinion. There is no scientific way to determine whether a piece of music is capable of expressing feelings. However, by using deductive reasoning, we can gain a better understanding of the potential for a jazz to express feelings.\\
            Answer and Explain: \textcolor{red}{\faTimesCircle} \textcolor{cyan}{\faTimesCircle} \textcolor{teal}{\faCheckCircle} \textcolor{purple}{\faCheckCircle}$\;\;\;\;\;\;\;\;$Explain Error Type: \textcolor{green}{\usym{1F5F5}} \textcolor{blue}{\usym{1F5F9}} \textcolor{brown}{\usym{1F5F5}} \textcolor{magenta}{\usym{1F5F5}}  \textcolor{violet}{\usym{1F5F9}}
            }\\
		\bottomrule
	\end{longtable}
}
\twocolumn

\onecolumn
{\small
\begin{longtable}{m{0.5cm}|l}
\caption{A reasoning case in bAbI-16. }\\
		\toprule
            \rotatebox{90}{\makecell[c]{Question, answer\\and explanation}} & \makecell[{{p{16.3cm}}}]{Context: 1 Greg is a rhino. 2  Brian is a swan. 3  Bernhard is a frog. 4  Brian is green. 5  Greg is gray. 6  Bernhard is yellow. 7  Julius is a lion. 8  Julius is yellow. 9  Lily is a rhino.\\
            Question:  What color is Lily?\\
            Answer: \textcolor{red}{Gray}\\
            Explanation: IsRhino(Greg) (1), IsGray(Greg) (5)$\rightarrow$ IsGray(rhino), IsRhino(Lily) (9)$\rightarrow$IsGray(Lily)
            }\\
            \hline
            \rotatebox{90}{\makecell[c]{ChatGPT\\of 0-shot}}& \makecell[{{p{16cm}}}]{Based on the given facts and inductive reasoning, it is likely that Lily's color is \textcolor{red}{not definitively determined}. Although Lily is described as a rhino, there is no information given that directly pertains to Lily's color. Additionally, there is no other animal descriptions that could be used to make an inference regarding Lily's color. Therefore, based on the given facts, it is not possible to make an accurate determination of Lily's color using inductive reasoning.\\
            Answer and Explain: \textcolor{red}{\faTimesCircle} \textcolor{cyan}{\faTimesCircle} \textcolor{teal}{\faTimesCircle} \textcolor{purple}{\faTimesCircle}$\;\;\;\;\;\;\;\;$Explain Error Type: \textcolor{green}{\usym{1F5F9}} \textcolor{blue}{\usym{1F5F5}} \textcolor{brown}{\usym{1F5F9}} \textcolor{magenta}{\usym{1F5F5}}  \textcolor{violet}{\usym{1F5F5}}
            }\\
            \hline
            \rotatebox{90}{\makecell[c]{ChatGPT\\of 1-shot}}& \makecell[{{p{16cm}}}]{Lily's color is likely to be \textcolor{red}{unknown} as there is no information provided about her color among the other animals.\\
            Answer and Explain: \textcolor{red}{\faTimesCircle} \textcolor{cyan}{\faTimesCircle} \textcolor{teal}{\faTimesCircle} \textcolor{purple}{\faTimesCircle}$\;\;\;\;\;\;\;\;$Explain Error Type: \textcolor{green}{\usym{1F5F9}} \textcolor{blue}{\usym{1F5F5}} \textcolor{brown}{\usym{1F5F9}} \textcolor{magenta}{\usym{1F5F5}}  \textcolor{violet}{\usym{1F5F5}}
            }\\
            \hline
            \rotatebox{90}{\makecell[c]{ChatGPT\\of 3-shot}}& \makecell[{{p{16cm}}}]{Brian, Greg, and Bernhard, which do not directly provide clues about what color Lily may be. Therefore, we \textcolor{red}{cannot be certain} about her color based on the information given alone.\\
            Answer and Explain: \textcolor{red}{\faTimesCircle} \textcolor{cyan}{\faTimesCircle} \textcolor{teal}{\faTimesCircle} \textcolor{purple}{\faTimesCircle}$\;\;\;\;\;\;\;\;$Explain Error Type: \textcolor{green}{\usym{1F5F9}} \textcolor{blue}{\usym{1F5F5}} \textcolor{brown}{\usym{1F5F9}} \textcolor{magenta}{\usym{1F5F5}}  \textcolor{violet}{\usym{1F5F5}}
            }\\
            \hline
            \rotatebox{90}{\makecell[c]{Davinci-003\\of 0-shot}}& \makecell[{{p{16cm}}}]{
            Based on the given facts, we can use inductive reasoning to conclude that Lily is \textcolor{red}{gray}. This is because all of the other rhinos in the given facts are gray, and there is no evidence to suggest that Lily is any different. In fact, the only thing that distinguishes Lily from the other rhinos is her name, \textcolor{purple}{which is a flower that is typically white or yellow. However, we can safely assume that Lily's name is not a reference to her color, as there are many other animals and plants that share the same name.} Therefore, the most likely conclusion is that Lily is gray, just like the other rhinos. Here is a more detailed explanation of the inductive reasoning process used to reach this conclusion:\\
            1. We are given the following facts: * Greg is a rhino. * Brian is a swan. * Bernhard is a frog. * Brian is green. * Greg is gray. * Bernhard is yellow. * Julius is a lion. * Julius is yellow. * Lily is a rhino.\\
            2. We notice that all of the rhinos in the given facts are gray.\\
            3. We conclude that there is a strong likelihood that Lily is also gray, as there is no evidence to suggest that she is any different from the other rhinos.\\
            4. \textcolor{purple}{We can safely assume that Lily's name is not a reference to her color, as there are many other animals and plants that share the same name.}\\
            5. Therefore, the most likely conclusion is that Lily is gray.\\
            Answer and Explain: \textcolor{red}{\faCheckCircle} \textcolor{cyan}{\faCheckCircle} \textcolor{teal}{\faCheckCircle} \textcolor{purple}{\faCheckCircle}
            }\\
            \hline
            \rotatebox{90}{\makecell[c]{Bard\\of 0-shot}}& \makecell[{{p{16cm}}}]{
            Based on inductive reasoning, it is likely that Lily is \textcolor{red}{gray}, as both Greg and \textcolor{blue}{Julius are gray} and they are both rhinos, like Lily. Since the only \textcolor{purple}{other colors mentioned among the animals are green and yellow}, and none of the rhinos have these colors, it is reasonable to assume that Lily is gray.\\
            Answer and Explain: \textcolor{red}{\faCheckCircle} \textcolor{cyan}{\faTimesCircle} \textcolor{teal}{\faCheckCircle} \textcolor{purple}{\faCheckCircle}$\;\;\;\;\;\;\;\;$Explain Error Type: \textcolor{green}{\usym{1F5F5}} \textcolor{blue}{\usym{1F5F9}} \textcolor{brown}{\usym{1F5F5}} \textcolor{magenta}{\usym{1F5F9}}  \textcolor{violet}{\usym{1F5F5}}
            }\\
		\bottomrule
	\end{longtable}
}
\twocolumn

\onecolumn
{\small
\begin{longtable}{m{0.5cm}|l}
\caption{A reasoning case in CLUTRR. }\label{tab:long}\\
		\toprule
            \rotatebox{90}{\makecell[c]{Question, answer\\and explanation}} & \makecell[{{p{16.3cm}}}]{Context: 1 [Chuck] took his daughter [Sara] to the park that afternoon to play. 2 [Judy] took her sister, [Maxine], out to dinner for her birthday. 3 [Chuck] took his sister [Judy] out to lunch after learning that she got accepted into her first choice for university. 4 [Maxine] asked her daughter, [Joyce], if she would like to go to a movie with her on Saturday night. 5 [Maxine] and her daughter [Joyce] went to see a movie yesterday and then got ice cream afterwards. 6 [Reynaldo]'s wife, [Maxine], was thrilled to learn she was pregnant. 7 [Naomi] went shoe shopping with her sister [Sara].\\
            Question: What is the kinship between Reynaldo and Naomi?\\
            Answer: \textcolor{red}{niece}\\
            Explanation: [(Reynaldo, wife, Maxine) 6, (Maxine, daughter, Joyce) 5] $\rightarrow$ (Reynaldo, daughter, Joyce) i1\\
            $\;$[(Maxine, sister, Judy) 2, (Judy, brother, Chuck) 3] $\rightarrow$ (Maxine, brother, Chuck) i2\\
            $\;$[(Joyce, mother, Maxine) 4, (Maxine, brother, Chuck) i2] $\rightarrow$ (Joyce, uncle, Chuck) i3\\
            $\;$[(Reynaldo, daughter, Joyce) i1, (Joyce, uncle, Chuck) i4] $\rightarrow$ (Reynaldo, brother, Chuck) i4\\
            $\;$[(Chuck, daughter, Sara) 1, (Sara, sister, Naomi) 7] $\rightarrow$ (Chuck, daughter, Naomi) i5\\
            $\;$[(Reynaldo, brother, Chuck) i4, (Chuck, daughter, Naomi) i5] $\rightarrow$ (Reynaldo, niece, Naomi)
            }\\
            \hline
            \rotatebox{90}{\makecell[c]{ChatGPT\\of 0-shot}}& \makecell[{{p{16cm}}}]{
            There is \textcolor{red}{no direct or indirect} relation mentioned between Reynaldo and Naomi in the given facts. \\
            Answer and Explain: \textcolor{red}{\faTimesCircle} \textcolor{cyan}{\faTimesCircle} \textcolor{teal}{\faTimesCircle} \textcolor{purple}{\faTimesCircle}$\;\;\;\;\;\;\;\;$Explain Error Type: \textcolor{green}{\usym{1F5F9}} \textcolor{blue}{\usym{1F5F5}} \textcolor{brown}{\usym{1F5F9}} \textcolor{magenta}{\usym{1F5F5}}  \textcolor{violet}{\usym{1F5F5}}
            }\\
            \hline
            \rotatebox{90}{\makecell[c]{ChatGPT\\of 1-shot}}& \makecell[{{p{16cm}}}]{
            Unclear, as there is \textcolor{red}{not enough} information about Naomi's relation to Reynaldo in the given facts.\\
            Answer and Explain: \textcolor{red}{\faTimesCircle} \textcolor{cyan}{\faTimesCircle} \textcolor{teal}{\faTimesCircle} \textcolor{purple}{\faTimesCircle}$\;\;\;\;\;\;\;\;$Explain Error Type: \textcolor{green}{\usym{1F5F9}} \textcolor{blue}{\usym{1F5F5}} \textcolor{brown}{\usym{1F5F9}} \textcolor{magenta}{\usym{1F5F5}}  \textcolor{violet}{\usym{1F5F5}}
            }\\
            \hline
            \rotatebox{90}{\makecell[c]{ChatGPT\\of 3-shot}}& \makecell[{{p{16cm}}}]{
            The given facts do not provide any information about the kinship between Reynaldo and Naomi, so the question \textcolor{red}{cannot be answered} using inductive reasoning.\\
            Answer and Explain: \textcolor{red}{\faTimesCircle} \textcolor{cyan}{\faTimesCircle} \textcolor{teal}{\faTimesCircle} \textcolor{purple}{\faTimesCircle}$\;\;\;\;\;\;\;\;$Explain Error Type: \textcolor{green}{\usym{1F5F9}} \textcolor{blue}{\usym{1F5F5}} \textcolor{brown}{\usym{1F5F9}} \textcolor{magenta}{\usym{1F5F5}}  \textcolor{violet}{\usym{1F5F5}}
            }\\
            \hline
            \rotatebox{90}{\makecell[c]{Davinci-003\\of 0-shot}}& \makecell[{{p{16cm}}}]{
            Reynaldo and Naomi have \textcolor{red}{no direct kinship}; \textcolor{blue}{they are both related to Maxine}, as Reynaldo is her husband and Naomi is Sara's sister.\\
            Answer and Explain: \textcolor{red}{\faTimesCircle} \textcolor{cyan}{\faTimesCircle} \textcolor{teal}{\faTimesCircle} \textcolor{purple}{\faCheckCircle}$\;\;\;\;\;\;\;\;$Explain Error Type: \textcolor{green}{\usym{1F5F9}} \textcolor{blue}{\usym{1F5F9}} \textcolor{brown}{\usym{1F5F9}} \textcolor{magenta}{\usym{1F5F5}}  \textcolor{violet}{\usym{1F5F5}}
            }\\
            \hline
            \rotatebox{90}{\makecell[c]{Bard\\of 0-shot}}& \makecell[{{p{16cm}}}]{
            Reynaldo and Naomi are \textcolor{red}{siblings-in-law}.\\
            Inductive reasoning is a type of reasoning that uses specific examples to make a general conclusion. In this case, we can use the following specific examples to make the general conclusion that Reynaldo and Naomi are siblings-in-law:\\
            Maxine is the wife of Reynaldo.\\
            \textcolor{blue}{Sara is the daughter of Maxine}.\\
            Naomi is the sister of Sara.\\
            Therefore, Naomi is the sister-in-law of Reynaldo.\\
            Here are some simple explanations for each of the specific examples:\\
            Maxine is the wife of Reynaldo because we are told that Reynaldo's wife is Maxine.\\
            \textcolor{blue}{Sara is the daughter of Maxine because we are told that Maxine has a daughter named Sara}.\\
            Naomi is the sister of Sara because we are told that Naomi went shoe shopping with her sister Sara.\\
            Answer and Explain: \textcolor{red}{\faTimesCircle} \textcolor{cyan}{\faTimesCircle} \textcolor{teal}{\faTimesCircle} \textcolor{purple}{\faCheckCircle}$\;\;\;\;\;\;\;\;$Explain Error Type: \textcolor{green}{\usym{1F5F9}} \textcolor{blue}{\usym{1F5F9}} \textcolor{brown}{\usym{1F5F9}} \textcolor{magenta}{\usym{1F5F5}}  \textcolor{violet}{\usym{1F5F5}}
            }\\
		\bottomrule
	\end{longtable}
}
\twocolumn

\onecolumn
{\small
\begin{longtable}{m{0.5cm}|l}
\caption{A reasoning case in $\alpha$-NLI. }\label{tab:long}\\
		\toprule
            \rotatebox{90}{\makecell[c]{Question, answer\\and explanation}} & \makecell[{{p{16.3cm}}}]{Context: When I was 19, I enlisted with the USMC. When I got back from the hospital, the USMC sent me home.\\
            Question: Choose the more likely explanation from following two options: A. I got injured the last day during a training event. B. I was shot in action but didn't regret it.\\
            Answer: \textcolor{red}{B}
            }\\
            \hline
            \rotatebox{90}{\makecell[c]{ChatGPT\\of 0-shot}}& \makecell[{{p{16cm}}}]{
            The more likely explanation is hypothesis choice \textcolor{red}{A}: I got injured the last day during a training event. This is because the context mentions the narrator enlisting with the USMC and being sent home after getting back from the hospital, suggesting a military-related scenario. Hypothesis choice A of the narrator getting injured during a training event aligns with these details and provides a probable explanation for why they may have been sent home, as injuries sustained during training may prevent soldiers from fulfilling duties and may require medical attention. Hypothesis choice B of the narrator being shot in action doesn't fit with the context, as there is no indication of any mention of being in combat or any explanation for how this would have been related to the narrator's hospitalization or being sent home, and it doesn't provide an explanation for what may have motivated the narrator's military-related interests or preferences.\\
            Answer and Explain: \textcolor{red}{\faTimesCircle} \textcolor{cyan}{\faTimesCircle} \textcolor{teal}{\faCheckCircle} \textcolor{purple}{\faTimesCircle}$\;\;\;\;\;\;\;\;$Explain Error Type: \textcolor{green}{\usym{1F5F5}} \textcolor{blue}{\usym{1F5F5}} \textcolor{brown}{\usym{1F5F5}} \textcolor{magenta}{\usym{1F5F5}}  \textcolor{violet}{\usym{1F5F9}}
            }\\
            \hline
            \rotatebox{90}{\makecell[c]{ChatGPT\\of 1-shot}}& \makecell[{{p{16cm}}}]{
            The correct choice is \textcolor{red}{A}. The context suggests that the person was enlisted in the USMC and got sent home from the hospital, indicating that they likely got injured during a training event. Choice B suggests an unlikely scenario where the person got shot in action, which cannot be inferred from the context and contradicts the fact that the person did not regret it.\\
            Answer and Explain: \textcolor{red}{\faTimesCircle} \textcolor{cyan}{\faTimesCircle} \textcolor{teal}{\faCheckCircle} \textcolor{purple}{\faTimesCircle}$\;\;\;\;\;\;\;\;$Explain Error Type: \textcolor{green}{\usym{1F5F5}} \textcolor{blue}{\usym{1F5F5}} \textcolor{brown}{\usym{1F5F5}} \textcolor{magenta}{\usym{1F5F5}}  \textcolor{violet}{\usym{1F5F9}}
            }\\
            \hline
            \rotatebox{90}{\makecell[c]{ChatGPT\\of 3-shot}}& \makecell[{{p{16cm}}}]{
            The correct choice is: \textcolor{red}{A}. The context suggests that the narrator enlisted with the USMC and was sent home from the hospital. Hypothesis A, where the narrator got injured the last day during a training event, is the more likely explanation because it fits with the idea of being sent home from the hospital after being injured in the line of duty. Hypothesis B, where the narrator was shot in action but didn't regret it, does not make sense as a plausible explanation for being sent home from the hospital expressed in the context.\\
            Answer and Explain: \textcolor{red}{\faTimesCircle} \textcolor{cyan}{\faTimesCircle} \textcolor{teal}{\faCheckCircle} \textcolor{purple}{\faTimesCircle}$\;\;\;\;\;\;\;\;$Explain Error Type: \textcolor{green}{\usym{1F5F5}} \textcolor{blue}{\usym{1F5F5}} \textcolor{brown}{\usym{1F5F5}} \textcolor{magenta}{\usym{1F5F5}}  \textcolor{violet}{\usym{1F5F9}}
            }\\
            \hline
            \rotatebox{90}{\makecell[c]{Davinci-003\\of 0-shot}}& \makecell[{{p{16cm}}}]{
            The more likely explanation is \textcolor{red}{A}. I got injured the last day during a training event. This could have been caused by some sort of accident with a weapon or a physical training exercise gone wrong. It is more likely that I would have been injured on the last day of service, as opposed to being shot in action, which would be less likely in the US Marine Corps.\\
            Answer and Explain: \textcolor{red}{\faTimesCircle} \textcolor{cyan}{\faTimesCircle} \textcolor{teal}{\faTimesCircle} \textcolor{purple}{\faTimesCircle}$\;\;\;\;\;\;\;\;$Explain Error Type: \textcolor{green}{\usym{1F5F5}} \textcolor{blue}{\usym{1F5F5}} \textcolor{brown}{\usym{1F5F5}} \textcolor{magenta}{\usym{1F5F5}}  \textcolor{violet}{\usym{1F5F9}}
            }\\
            \hline
            \rotatebox{90}{\makecell[c]{Bard\\of 0-shot}}& \makecell[{{p{16cm}}}]{
            The more likely explanation is \textcolor{red}{A}. I got injured the last day during a training event.\\
            This is because the context states that you enlisted with the USMC at 19, and that you were sent home from the hospital after being injured. This suggests that the injury occurred during training, and that it was serious enough to warrant your discharge from the military.\\
            Hypothesis B is less likely because it is not supported by the context. The context does not mention anything about you being shot in action, or about you not regretting it. Additionally, it is not clear why the USMC would send you home if you were not injured.\\
            Here is a simple explanation for why you might have been injured during training:\\
            You were involved in a training accident.\\
            You overexerted yourself and suffered an injury.\\
            You were injured by another Marine during training.\\
            It is important to remember that these are just possible explanations, and we cannot know for sure why you were injured without more information.\\
            Answer and Explain: \textcolor{red}{\faTimesCircle} \textcolor{cyan}{\faTimesCircle} \textcolor{teal}{\faCheckCircle} \textcolor{purple}{\faTimesCircle}$\;\;\;\;\;\;\;\;$Explain Error Type: \textcolor{green}{\usym{1F5F5}} \textcolor{blue}{\usym{1F5F5}} \textcolor{brown}{\usym{1F5F5}} \textcolor{magenta}{\usym{1F5F5}}  \textcolor{violet}{\usym{1F5F9}}
            }\\
		\bottomrule
	\end{longtable}
}
\twocolumn

\onecolumn
{\small
\begin{longtable}{m{0.5cm}|l}
\caption{A reasoning case in $\alpha$-NLG. }\label{tab:long}\\
		\toprule
            \rotatebox{90}{\makecell[c]{Question, answer\\and explanation}} & \makecell[{{p{16.3cm}}}]{Context: Joy's mom sent her in the donut store for a dozen donuts. She was happy as she left with 13 donuts instead of 12. \\
            Question: Generate a short hypothesis for the context and give simple explanations.\\
            Answer: \textcolor{red}{Joy got an extra donut.}
            }\\
            \hline
            \rotatebox{90}{\makecell[c]{ChatGPT\\of 0-shot}}& \makecell[{{p{16cm}}}]{
            Hypothesis: The fact that Joy's mom sent her to the donut store for a dozen donuts, and Joy left with 13 instead of 12 might have been a result of \textcolor{red}{the potential personal or environmental factors related to perception, generosity, or habit}.\\
            Explanation: The fact that Joy's mom sent her to the donut store for a dozen donuts suggests that it might have some potential personal or environmental factors related to perception, generosity, or habit. Having personal or environmental factors can involve various types of cognitive or practical factors, such as accuracy, generosity, or routine, and can reflect one's sense of responsibility or automaticity. Thus, the fact that Joy's mom sent her to the donut store for a dozen donuts might have been a result of the potential personal or environmental factors related to perception, generosity, or habit. Additionally, the fact that Joy left with 13 donuts instead of 12 implies that there might have been some additional potential cognitive or contextual factors related to accuracy, availability, or preference. Having cognitive or contextual factors can involve various types of perception or action, such as estimation, surplus, or choice, and can reflect one's sense of intention or opportunity. Furthermore, the fact that Joy was happy as she left with 13 donuts instead of 12 suggests that there might have been some potential affective or symbolic factors related to value, satisfaction, or normativity. Having affective or symbolic factors can involve various types of appraisal or intent, such as pleasure, gratitude, or conformity, and can reflect one's sense of social or moral obligation. Therefore, the hypothesis suggests that the fact that Joy's mom sent her to the donut store for a dozen donuts, and Joy left with 13 instead of 12 might have been a result of the potential personal or environmental factors related to perception, generosity, or habit, as well as the potential cognitive or contextual factors related to accuracy, availability, or preference.\\
            Answer and Explain: \textcolor{red}{\faTimesCircle} \textcolor{cyan}{\faTimesCircle} \textcolor{teal}{\faCheckCircle} \textcolor{purple}{\faTimesCircle}$\;\;\;\;\;\;\;\;$Explain Error Type: \textcolor{green}{\usym{1F5F5}} \textcolor{blue}{\usym{1F5F5}} \textcolor{brown}{\usym{1F5F5}} \textcolor{magenta}{\usym{1F5F5}}  \textcolor{violet}{\usym{1F5F9}}
            }\\
            \hline
            \rotatebox{90}{\makecell[c]{ChatGPT\\of 1-shot}}& \makecell[{{p{16cm}}}]{
            Hypothesis: The reason why Joy left the donut store with 13 instead of 12 donuts may be due to a \textcolor{red}{generous gesture from the store clerk, a promotional offer that allowed her to receive an extra donut for free, a mistake made by the store clerk, or Joy's persuasive negotiation or effective communication skills}.\\
            Explanation: The fact that Joy left the donut store with 13 instead of 12 donuts suggests that there may have been an unexpected addition or alteration to her order. One possibility is that the store clerk may have generously offered Joy an additional donut as a kind gesture or to show appreciation for her business. Another possibility is that there may have been a promotional offer where customers who purchased a dozen donuts received an additional donut for free. Alternatively, a mistake may have been made by the store clerk when fulfilling Joy's order, resulting in an extra donut being added to her purchase. Finally, Joy may have effectively persuaded or negotiated with the store clerk to receive an extra donut, perhaps by demonstrating loyal patronage or simply asking for an additional donut. These factors may have contributed to Joy's pleasant surprise at leaving the donut store with an extra donut, and highlight the importance of customer service, promotions, careful order fulfillment, and communication skills in the food service industry. Therefore, the hypothesis proposes that the reason why Joy left the donut store with 13 instead of 12 donuts may be due to a generous gesture from the store clerk, a promotional offer that allowed her to receive an extra donut for free, a mistake made by the store clerk, or Joy's persuasive negotiation or effective communication skills.\\
            Answer and Explain: \textcolor{red}{\faCheckCircle} \textcolor{cyan}{\faCheckCircle} \textcolor{teal}{\faCheckCircle} \textcolor{purple}{\faTimesCircle}
            }\\
            \hline
            \rotatebox{90}{\makecell[c]{ChatGPT\\of 3-shot}}& \makecell[{{p{16cm}}}]{
            Hypothesis: \textcolor{red}{The donut shop may have made an error and accidentally given Joy an extra donut}, which she was pleased to receive. Alternatively, \textcolor{red}{the donut store may have had a special promotion where customers receive an extra donut for free}, which Joy was not aware of but happily benefited from. This scenario highlights the potential for unexpected surprises and bonuses in our daily lives.\\
            Answer and Explain: \textcolor{red}{\faCheckCircle} \textcolor{cyan}{\faCheckCircle} \textcolor{teal}{\faCheckCircle} \textcolor{purple}{\faTimesCircle}
            }\\
            \hline
            \rotatebox{90}{\makecell[c]{Davinci-003\\of 0-shot}}& \makecell[{{p{16cm}}}]{
            Hypothesis: \textcolor{red}{The donut store owner gave Joy an extra donut for free}. \\
            Explanation: The extra donut could indicate that the store owner gave Joy an extra donut for free as a gesture of goodwill or appreciation. This could be due to Joy being loyal customer at the store, or due to her being a child and the store owner having sympathy for her.\\
            Answer and Explain: \textcolor{red}{\faCheckCircle} \textcolor{cyan}{\faCheckCircle} \textcolor{teal}{\faCheckCircle} \textcolor{purple}{\faTimesCircle}
            }\\
            \hline
            \rotatebox{90}{\makecell[c]{Bard\\of 0-shot}}& \makecell[{{p{16cm}}}]{
            Sure, here are some possible hypotheses for why Joy was happy when she left the donut store with 13 donuts instead of 12:\\
            \textcolor{red}{The donut store was giving away a free donut with every dozen purchased}.\\
            The donut store made a mistake and gave Joy an extra donut.\\
            Joy was able to negotiate with the donut store owner to get an extra donut.\\
            Joy is a very good friend or family member and the donut store owner wanted to give her a little extra something.\\
            Here are some simple explanations for each hypothesis:\\
            The donut store was giving away a free donut with every dozen purchased because they were trying to promote their business or because they had too many donuts on hand.\\
            The donut store made a mistake and gave Joy an extra donut because they were busy or because they were not paying attention.\\
            Joy was able to negotiate with the donut store owner to get an extra donut because she was polite and persuasive.\\
            Joy is a very good friend or family member and the donut store owner wanted to give her a little extra something because they appreciated her business or because they were close friends or family.\\
            It is impossible to know for sure which hypothesis is correct, but all of them are possible explanations for why Joy was happy when she left the donut store with 13 donuts instead of 12.\\
            Answer and Explain: \textcolor{red}{\faCheckCircle} \textcolor{cyan}{\faCheckCircle} \textcolor{teal}{\faCheckCircle} \textcolor{purple}{\faTimesCircle}
            }\\
		\bottomrule
	\end{longtable}
}
\twocolumn

\onecolumn
{\small
\begin{longtable}{m{0.5cm}|l}
\caption{A reasoning case in AbductiveRules. }\label{tab:long}\\
		\toprule
            \rotatebox{90}{\makecell[c]{Question, answer\\and explanation}} & \makecell[{{p{16.3cm}}}]{Context: 1 Gary is strong. 2 Things that are tiny, are small, and are short, are thin. 3 All people that are quiet, are poor, and are smart, are clever. 4 Gary is big. 5 If something is short, is tiny, and is nice, it is also thin. 6 If a person is bad, is sad, and is strong, that person is rough. 7 Things that are strong, are huge, and are big, are high. 8 People that are bad, are short, and are strong, are also clever. 9 Dave is poor. 10 Dave is imperfect. 11 Charlie is quiet. 12 Gary is heavy. 13 Anne is tiny. 14 All things that are huge, are small, and are big, are also high. 15 All things that are quiet, are huge, and are poor, are also thin. 16 Anne is little. 17 Charlie is nice. 18 Dave is bad. 19 If a person is tiny, is nice, and is sad, that person is high. 20 If a person is quiet, is nice, and is smart, that person is clever. 21 All things that are small, are big, and are smart, are rough. 22 Anne is small. 23 People that are bad, are poor, and are sad, are also rough.\\
            Question: Please generate a short explanation for the given context and observation. The observation is: Dave is rough.\\
            Answer: \textcolor{red}{Dave is sad.}\\
            Explanation: 9, 18$\xrightarrow{23}$ Dave is sad.
            }\\
            \hline
            \rotatebox{90}{\makecell[c]{ChatGPT\\of 0-shot}}& \makecell[{{p{16cm}}}]{
            Based on the given context, individuals who possess traits such \textcolor{purple}{as badness, sadness, and strength are typically perceived as rough}. Dave has been described as poor and imperfect which would suggest a degree of hardship and inadequacy, and the observation confirms that he is perceived as rough. Therefore, it can be inferred that Dave's perceived roughness is a result of his \textcolor{red}{perceived badness, sadness, and poor state} in accordance with the given context.\\
            Answer and Explain: \textcolor{red}{\faTimesCircle} \textcolor{cyan}{\faTimesCircle} \textcolor{teal}{\faTimesCircle} \textcolor{purple}{\faCheckCircle}$\;\;\;\;\;\;\;\;$Explain Error Type: \textcolor{green}{\usym{1F5F9}} \textcolor{blue}{\usym{1F5F5}} \textcolor{brown}{\usym{1F5F5}} \textcolor{magenta}{\usym{1F5F9}}  \textcolor{violet}{\usym{1F5F5}}
            }\\
            \hline
            \rotatebox{90}{\makecell[c]{ChatGPT\\of 1-shot}}& \makecell[{{p{16cm}}}]{
            The context indicates that being \textcolor{purple}{bad, sad, and strong is associated with being rough}, and being poor and imperfect are also negative qualities. The observation that Dave is rough suggests that he may possess these negative traits or exhibit aggressive behavior.\\
            Answer and Explain: \textcolor{red}{\faTimesCircle} \textcolor{cyan}{\faTimesCircle} \textcolor{teal}{\faTimesCircle} \textcolor{purple}{\faCheckCircle}$\;\;\;\;\;\;\;\;$Explain Error Type: \textcolor{green}{\usym{1F5F9}} \textcolor{blue}{\usym{1F5F5}} \textcolor{brown}{\usym{1F5F5}} \textcolor{magenta}{\usym{1F5F9}}  \textcolor{violet}{\usym{1F5F5}}
            }\\
            \hline
            \rotatebox{90}{\makecell[c]{ChatGPT\\of 3-shot}}& \makecell[{{p{16cm}}}]{
            The given observation contradicts the context, as it states that Dave is rough whereas the context indicates that \textcolor{purple}{he is poor and imperfect}. Therefore, there is \textcolor{red}{no valid explanation} for this observation within the given context.\\
            Answer and Explain: \textcolor{red}{\faTimesCircle} \textcolor{cyan}{\faTimesCircle} \textcolor{teal}{\faTimesCircle} \textcolor{purple}{\faCheckCircle}$\;\;\;\;\;\;\;\;$Explain Error Type: \textcolor{green}{\usym{1F5F9}} \textcolor{blue}{\usym{1F5F5}} \textcolor{brown}{\usym{1F5F9}} \textcolor{magenta}{\usym{1F5F5}}  \textcolor{violet}{\usym{1F5F5}}
            }\\
            \hline
            \rotatebox{90}{\makecell[c]{Davinci-003\\of 0-shot}}& \makecell[{{p{16cm}}}]{
            Dave is rough because \textcolor{red}{he is bad, poor, and strong}, which meets the criteria of \textcolor{purple}{people that are bad, poor, and strong being rough}.\\
            Answer and Explain: \textcolor{red}{\faTimesCircle} \textcolor{cyan}{\faTimesCircle} \textcolor{teal}{\faTimesCircle} \textcolor{purple}{\faCheckCircle}$\;\;\;\;\;\;\;\;$Explain Error Type: \textcolor{green}{\usym{1F5F9}} \textcolor{blue}{\usym{1F5F5}} \textcolor{brown}{\usym{1F5F9}} \textcolor{magenta}{\usym{1F5F5}}  \textcolor{violet}{\usym{1F5F5}}
            }\\
            \hline
            \rotatebox{90}{\makecell[c]{Bard\\of 0-shot}}& \makecell[{{p{16cm}}}]{
            Sure, here is a short explanation for the given context and observation:\\
            Dave is rough because \textcolor{red}{he is bad, sad, and strong}. He is also poor and imperfect. This suggests that he is also quiet and clever. However, he is also tiny and small. This suggests that he is also weak.\\
            Here is a more detailed explanation:\\
            Dave is bad. This is evident from the fact that he is described as ``bad" and ``rough".\\
            Dave is sad. This is evident from the fact that he is described as \textcolor{blue}{``sad"} and ``imperfect".\\
            \textcolor{purple}{Dave is strong}. This is evident from the fact that he is described as \textcolor{blue}{``strong"} and \textcolor{blue}{``huge"}.\\
            Dave is poor. This is evident from the fact that he is described as ``poor" and ``imperfect".\\
            Dave is imperfect. This is evident from the fact that he is described as \textcolor{blue}{``sad"} and ``imperfect".\\
            \textcolor{purple}{Dave is quiet}. This is evident from the fact that he is described as \textcolor{blue}{``quiet"} and \textcolor{blue}{``clever"}.\\
            \textcolor{purple}{Dave is clever}. This is evident from the fact that he is described as \textcolor{blue}{``clever"} and \textcolor{blue}{``strong"}\\
            \textcolor{purple}{Dave is tiny}. This is evident from the fact that he is described as \textcolor{blue}{``tiny"} and \textcolor{blue}{``small"}.\\
            \textcolor{purple}{Dave is small}. This is evident from the fact that he is described as \textcolor{blue}{``small"} and \textcolor{blue}{``weak"}.\\
            Dave's badness, sadness, strength, poverty, imperfection, quietness, cleverness, tininess, and smallness suggest that he is a complex and contradictory person. He is capable of great things, but he is also capable of great evil. He is a force to be reckoned with, and he will not be easily defeated.\\
            Answer and Explain: \textcolor{red}{\faTimesCircle} \textcolor{cyan}{\faTimesCircle} \textcolor{teal}{\faTimesCircle} \textcolor{purple}{\faCheckCircle}$\;\;\;\;\;\;\;\;$Explain Error Type: \textcolor{green}{\usym{1F5F9}} \textcolor{blue}{\usym{1F5F9}} \textcolor{brown}{\usym{1F5F5}} \textcolor{magenta}{\usym{1F5F9}}  \textcolor{violet}{\usym{1F5F5}}
            }\\
		\bottomrule
	\end{longtable}
}
\twocolumn

\onecolumn
{\small
\begin{longtable}{m{0.5cm}|l}
\caption{A reasoning case in D*-Ab. }\label{tab:long}\\
		\toprule
            \rotatebox{90}{\makecell[c]{Question, answer\\and explanation}} & \makecell[{{p{16.3cm}}}]{Context: Fact1: The bear chases the cat. Fact2: The bear is big. Fact3: The bear is cold. Fact4: The bear likes the squirrel. Fact5: The bear likes the tiger. Fact6: The bear sees the squirrel. Fact7: The cat is big. Fact8: The cat is cold. Fact9: The cat is young. Fact10: The cat likes the bear. Fact11: The cat likes the tiger. Fact12: The cat sees the bear. Fact13: The cat sees the tiger. Fact14: The squirrel is young. Fact15: The tiger chases the bear. Fact16: The tiger likes the squirrel. Rule1: If the tiger is young then the tiger likes the squirrel. Rule2: If something is round then it likes the bear. Rule3: If something likes the bear then it is cold. Rule4: If the tiger likes the bear and the bear sees the squirrel then the squirrel is round. Rule5: If something likes the squirrel then it is round. Rule6: If something chases the bear and the bear likes the cat then the cat likes the squirrel.\\
            Question: Please generate a short missing fact for the given context and fact. And give simple explanations. The fact is: The cat likes the squirrel.\\
            Answer: \textcolor{red}{The bear likes the cat.}\\
            Explanation: Fact15$\xrightarrow{Rule6}$ The bear likes the cat.
            }\\
            \hline
            \rotatebox{90}{\makecell[c]{ChatGPT\\of 0-shot}}& \makecell[{{p{16cm}}}]{
            Missing fact: \textcolor{red}{The squirrel is round}.\\
            Explanation: According to the given rules, \textcolor{purple}{if the tiger likes the bear and the bear sees the squirrel, then the squirrel is round}. As the \textcolor{purple}{tiger likes the squirrel} in the given context, we can conclude that the tiger likes the bear, and since the bear likes the squirrel, we can infer that the squirrel is round.\\
            Answer and Explain: \textcolor{red}{\faTimesCircle} \textcolor{cyan}{\faTimesCircle} \textcolor{teal}{\faTimesCircle} \textcolor{purple}{\faCheckCircle}$\;\;\;\;\;\;\;\;$Explain Error Type: \textcolor{green}{\usym{1F5F9}} \textcolor{blue}{\usym{1F5F5}} \textcolor{brown}{\usym{1F5F5}} \textcolor{magenta}{\usym{1F5F9}}  \textcolor{violet}{\usym{1F5F5}}
            }\\
            \hline
            \rotatebox{90}{\makecell[c]{ChatGPT\\of 1-shot}}& \makecell[{{p{16cm}}}]{
            The fact that the cat likes the squirrel suggests a potential behavioral or social relationship between the two animals, such as companionship or mutual interest in resources. The missing fact that \textcolor{red}{Anne is young} could imply a new or additional character to the context.\\
            Answer and Explain: \textcolor{red}{\faTimesCircle} \textcolor{cyan}{\faTimesCircle} \textcolor{teal}{\faTimesCircle} \textcolor{purple}{\faTimesCircle}$\;\;\;\;\;\;\;\;$Explain Error Type: \textcolor{green}{\usym{1F5F9}} \textcolor{blue}{\usym{1F5F5}} \textcolor{brown}{\usym{1F5F5}} \textcolor{magenta}{\usym{1F5F9}}  \textcolor{violet}{\usym{1F5F5}}
            }\\
            \hline
            \rotatebox{90}{\makecell[c]{ChatGPT\\of 3-shot}}& \makecell[{{p{16cm}}}]{
            According to the context, \textcolor{purple}{the bear likes the squirrel, and the cat likes the bear}, which could suggest a predator-prey relationship between the cat and the squirrel. Moreover, given the fact that the cat likes the squirrel, it is possible that the \textcolor{red}{cat may try to protect the squirrel from the bear's chase}. Additionally, the context suggests that.\\
            Answer and Explain: \textcolor{red}{\faTimesCircle} \textcolor{cyan}{\faTimesCircle} \textcolor{teal}{\faTimesCircle} \textcolor{purple}{\faCheckCircle}$\;\;\;\;\;\;\;\;$Explain Error Type: \textcolor{green}{\usym{1F5F9}} \textcolor{blue}{\usym{1F5F5}} \textcolor{brown}{\usym{1F5F5}} \textcolor{magenta}{\usym{1F5F9}}  \textcolor{violet}{\usym{1F5F5}}
            }\\
            \hline
            \rotatebox{90}{\makecell[c]{Davinci-003\\of 0-shot}}& \makecell[{{p{16cm}}}]{
            The missing fact is: \textcolor{red}{The tiger is young}. Explanation: The rule1 states that \textcolor{purple}{if the tiger is young then the tiger likes the squirrel. The fact16 states that the tiger likes the squirrel}, so the tiger must also be young.\\
            Answer and Explain: \textcolor{red}{\faTimesCircle} \textcolor{cyan}{\faTimesCircle} \textcolor{teal}{\faTimesCircle} \textcolor{purple}{\faCheckCircle}$\;\;\;\;\;\;\;\;$Explain Error Type: \textcolor{green}{\usym{1F5F9}} \textcolor{blue}{\usym{1F5F5}} \textcolor{brown}{\usym{1F5F5}} \textcolor{magenta}{\usym{1F5F9}}  \textcolor{violet}{\usym{1F5F5}}
            }\\
            \hline
            \rotatebox{90}{\makecell[c]{Bard\\of 0-shot}}& \makecell[{{p{16cm}}}]{
            Sure, here is a short missing fact for the given context and fact:\\
            \textcolor{red}{The squirrel sees the cat}.\\
            This is the missing fact because it is the only fact that can explain all of the given facts and rules. For example, \textcolor{purple}{Rule 4 says that if the tiger likes the bear and the bear sees the squirrel then the squirrel is round}. However, we know that the squirrel is round, so the squirrel must also see the bear.\\Here are some simple explanations for the missing fact:\\
            \textcolor{blue}{The squirrel is friends with the cat}.\\
            \textcolor{blue}{The squirrel is related to the cat}.\\
            \textcolor{blue}{The squirrel is a food source for the cat}.\\
            I hope this helps! Let me know if you have any other questions.\\
            However, I think there is a mistake in the given context. The fact that ``The cat likes the squirrel" contradicts the fact that "The cat chases the squirrel". If the cat likes the squirrel, then it would not chase the squirrel. Therefore, one of these facts must be incorrect.\\
            Answer and Explain: \textcolor{red}{\faTimesCircle} \textcolor{cyan}{\faTimesCircle} \textcolor{teal}{\faTimesCircle} \textcolor{purple}{\faCheckCircle}$\;\;\;\;\;\;\;\;$Explain Error Type: \textcolor{green}{\usym{1F5F9}} \textcolor{blue}{\usym{1F5F9}} \textcolor{brown}{\usym{1F5F5}} \textcolor{magenta}{\usym{1F5F9}}  \textcolor{violet}{\usym{1F5F5}}
            }\\
		\bottomrule
	\end{longtable}
}
\twocolumn

\onecolumn
{\small
\begin{longtable}{m{0.5cm}|l}
\caption{A reasoning case in ReClor. }\label{tab:long}\\
		\toprule
            \rotatebox{90}{\makecell[c]{Question, answer\\and explanation}} & \makecell[{{p{16.3cm}}}]{Context: A science class stored one selection of various fruits at 30 degrees Celsius, a similar selection in similar conditions at 20 degrees, and another similar selection in similar conditions at 10 degrees. Because the fruits stored at 20 degrees stayed fresh longer than those stored at 30 degrees, and those stored at 10 degrees stayed fresh longest, the class concluded that the cooler the temperature at which these varieties of fruits are stored, the longer they will stay fresh.\\
            Question: The class's reasoning is flawed in that the class A. neglected to offer any explanation for the results it discovered B. assumed without proof that its thermometer was reliable C. generalized too readily from the fruits it tested to fruits it did not test D. too readily extrapolated from a narrow range of temperatures to the entire range of temperatures.\\
            Answer: \textcolor{red}{D}
            }\\
            \hline
            \rotatebox{90}{\makecell[c]{ChatGPT\\of 0-shot}}& \makecell[{{p{16cm}}}]{
            The correct answer is \textcolor{red}{D}. Option A is not relevant to the class's reasoning, as it still made an attempt to draw a conclusion from its results. Option B is a possible flaw, but it is not stated in the argument whether the thermometer was reliable or not, so it cannot be the correct answer. Option C is a possible flaw, but it is not stated whether the class attempted to generalize its results to fruits it did not test, so it cannot be the correct answer. Option D is the correct answer, as the class too readily made a generalization from a narrow range of temperatures (10 degrees, 20 degrees, and 30 degrees) to the entire range of temperatures at which fruits can be stored, which is likely to be much broader.\\
            Answer and Explain: \textcolor{red}{\faCheckCircle} \textcolor{cyan}{\faCheckCircle} \textcolor{teal}{\faCheckCircle} \textcolor{purple}{\faTimesCircle}
            }\\
            \hline
            \rotatebox{90}{\makecell[c]{ChatGPT\\of 1-shot}}& \makecell[{{p{16cm}}}]{
            The correct answer is \textcolor{red}{C}. The class's reasoning is flawed in that it generalized too readily from the fruits it tested to fruits it did not test. The experiment only tested a specific selection of fruits and did not test all varieties of fruits, so it is not necessarily true that cooler temperatures will always keep fruits fresh longer. Option A is incorrect because the class did offer an explanation for the results they discovered. Option B is not relevant to the flaw in the class's reasoning. Option D is also not relevant, as the class did not extrapolate from a narrow range of temperatures to the entire range of temperatures.\\
            Answer and Explain: \textcolor{red}{\faTimesCircle} \textcolor{cyan}{\faTimesCircle} \textcolor{teal}{\faTimesCircle} \textcolor{purple}{\faTimesCircle}$\;\;\;\;\;\;\;\;$Explain Error Type: \textcolor{green}{\usym{1F5F5}} \textcolor{blue}{\usym{1F5F5}} \textcolor{brown}{\usym{1F5F5}} \textcolor{magenta}{\usym{1F5F9}}  \textcolor{violet}{\usym{1F5F5}}
            }\\
            \hline
            \rotatebox{90}{\makecell[c]{ChatGPT\\of 3-shot}}& \makecell[{{p{16cm}}}]{
            The correct answer is \textcolor{red}{D}. The class's reasoning is flawed in that they too readily extrapolated from a narrow range of temperatures to the entire range of temperatures. The class only tested three temperatures, and did not consider the possibility of a threshold beyond which colder temperatures may not be beneficial for all fruits. Therefore, it is too hasty to conclude that \"the cooler the temperature, the longer the fruits will stay fresh\" based on such a limited range of temperatures.\\
            Answer and Explain: \textcolor{red}{\faCheckCircle} \textcolor{cyan}{\faCheckCircle} \textcolor{teal}{\faTimesCircle} \textcolor{purple}{\faTimesCircle}
            }\\
            \hline
            \rotatebox{90}{\makecell[c]{Davinci-003\\of 0-shot}}& \makecell[{{p{16cm}}}]{
            The correct answer is \textcolor{red}{C}. The class's reasoning is flawed in that the class generalized too readily from the fruits it tested to fruits it did not test. The class was only testing \textcolor{blue}{three types of fruit} at three different temperatures, which does not truly demonstrate how all different varieties of fruits can react differently to varying temperatures. Thus, the class should not make a hasty generalization about how all fruits would respond to different temperatures.\\
            Answer and Explain: \textcolor{red}{\faTimesCircle} \textcolor{cyan}{\faTimesCircle} \textcolor{teal}{\faTimesCircle} \textcolor{purple}{\faTimesCircle}$\;\;\;\;\;\;\;\;$Explain Error Type: \textcolor{green}{\usym{1F5F5}} \textcolor{blue}{\usym{1F5F9}} \textcolor{brown}{\usym{1F5F5}} \textcolor{magenta}{\usym{1F5F9}}  \textcolor{violet}{\usym{1F5F5}}
            }\\
            \hline
            \rotatebox{90}{\makecell[c]{Bard\\of 0-shot}}& \makecell[{{p{16cm}}}]{
            The correct answer is \textcolor{red}{D}. Too readily extrapolated from a narrow range of temperatures to the entire range of temperatures.\\
            The class only tested fruits at three temperatures: 30 degrees Celsius, 20 degrees Celsius, and 10 degrees Celsius. They concluded that the cooler the temperature, the longer the fruits will stay fresh. However, this conclusion is based on a very small sample size. \textcolor{purple}{It is possible that the relationship between temperature and freshness is not linear. For example, it is possible that fruits will stay fresh for the same amount of time at 10 degrees Celsius and 5 degrees Celsius.} The class would need to test fruits at a wider range of temperatures to confirm their conclusion.\\
            The other options are not correct. Option A is incorrect because the class did offer an explanation for their results: they said that the cooler the temperature, the longer the fruits will stay fresh. Option B is incorrect because the class did not assume that their thermometer was unreliable. Option C is incorrect because the class did not generalize to fruits they did not test. They only tested fruits at three temperatures, and they only made a conclusion about those three temperatures.\\
            Answer and Explain: \textcolor{red}{\faCheckCircle} \textcolor{cyan}{\faCheckCircle} \textcolor{teal}{\faCheckCircle} \textcolor{purple}{\faCheckCircle}
            }\\
		\bottomrule
	\end{longtable}
}
\twocolumn

\onecolumn
{\small
\begin{longtable}{m{0.5cm}|l}
\caption{A reasoning case in LogiQA. }\label{tab:long}\\
		\toprule
            \rotatebox{90}{\makecell[c]{Question, answer\\and explanation}} & \makecell[{{p{16.3cm}}}]{Context: People often use microwave ovens to heat food.Some people think that microwave oven heating changes the molecular structure of food, producing molecules that the human body cannot recognize.These strange molecular formulas are unacceptable to the human body, and some are toxic and may even cause cancer. Therefore, people or animals who often eat microwave foods will have serious physiological changes in their bodies, which will cause serious health problems.\\
            Question: Which of the following can best question the above view? A. Microwave heating will not cause more nutrient loss than other cooking methods B. China's microwave oven production standards are consistent with international standards and EU standards C. It is also common for developed countries to use microwave ovens D. The microwave only heats the water molecules in the food, the food does not change chemically.\\
            Answer: \textcolor{red}{D}
            }\\
            \hline
            \rotatebox{90}{\makecell[c]{ChatGPT\\of 0-shot}}& \makecell[{{p{16cm}}}]{
            The answer is \textcolor{red}{D}. The given argument suggests that microwave oven heating changes the molecular structure of food, producing molecules that the human body cannot recognize and causing serious health problems. Option D provides an alternative explanation for the process of microwave heating, which states that the microwave only heats the water molecules in the food, and the food does not change chemically. Therefore, option D best questions the view presented in the argument. Option A is not relevant as it talks about nutrient loss rather than the effects of microwave heating on food. Option B and option C are not relevant as they provide information about the production standards of microwave ovens and their usage in different countries, but do not impact the argument presented in the context.\\
            Answer and Explain: \textcolor{red}{\faCheckCircle} \textcolor{cyan}{\faCheckCircle} \textcolor{teal}{\faCheckCircle} \textcolor{purple}{\faTimesCircle}
            }\\
            \hline
            \rotatebox{90}{\makecell[c]{ChatGPT\\of 1-shot}}& \makecell[{{p{16cm}}}]{
            The correct choice is: \textcolor{red}{D}. The argument suggests that the use of microwave ovens to heat food changes the molecular structure of the food, producing molecules that the human body cannot recognize, which can lead to serious health problems. Option D challenges this argument by stating that the microwave only heats the water molecules in the food and does not change the chemical structure of the food. This suggests that the argument may be based on a misunderstanding of how microwaves work. Option A, B, and C are not relevant to the argument or do not challenge the premise that microwaves can cause health problems.\\
            Answer and Explain: \textcolor{red}{\faCheckCircle} \textcolor{cyan}{\faCheckCircle} \textcolor{teal}{\faCheckCircle} \textcolor{purple}{\faTimesCircle}
            }\\
            \hline
            \rotatebox{90}{\makecell[c]{ChatGPT\\of 3-shot}}& \makecell[{{p{16cm}}}]{
            The correct choice is: \textcolor{red}{D}. The microwave only heats the water molecules in the food, the food does not change chemically, can best question the above view. It suggests that the claim about microwave ovens changing the molecular structure of food is not accurate and provides a scientific explanation for how microwave heating works, which does not involve the production of strange or toxic molecules.\\
            Answer and Explain: \textcolor{red}{\faCheckCircle} \textcolor{cyan}{\faCheckCircle} \textcolor{teal}{\faTimesCircle} \textcolor{purple}{\faTimesCircle}
            }\\
            \hline
            \rotatebox{90}{\makecell[c]{Davinci-003\\of 0-shot}}& \makecell[{{p{16cm}}}]{
            The best answer is \textcolor{red}{B}. China's microwave oven production standards are consistent with international standards and EU standards. This answer questions the view presented in the context that microwave oven heating changes the molecular structure of food, since it implies that safety regulations surrounding the production of microwave ovens would never be put in place if this were true. Therefore, this disproves the view that microwave oven heating produces molecules that the human body cannot recognize, which could be toxic and may even cause cancer.\\
            Answer and Explain: \textcolor{red}{\faTimesCircle} \textcolor{cyan}{\faTimesCircle} \textcolor{teal}{\faTimesCircle} \textcolor{purple}{\faTimesCircle}$\;\;\;\;\;\;\;\;$Explain Error Type: \textcolor{green}{\usym{1F5F5}} \textcolor{blue}{\usym{1F5F5}} \textcolor{brown}{\usym{1F5F5}} \textcolor{magenta}{\usym{1F5F9}}  \textcolor{violet}{\usym{1F5F5}}
            }\\
            \hline
            \rotatebox{90}{\makecell[c]{Bard\\of 0-shot}}& \makecell[{{p{16cm}}}]{
            The correct answer is \textcolor{red}{D}. The microwave only heats the water molecules in the food, the food does not change chemically.\\
            The context states that some people think that microwave oven heating changes the molecular structure of food, producing molecules that the human body cannot recognize. However, this is not true. Microwaves only heat the water molecules in food, and the food does not change chemically. This is why microwaved food is safe to eat.\\
            The other options are not necessarily correct. Option A is incorrect, \textcolor{blue}{as there is some evidence that microwave heating can cause nutrient loss.} Option B is irrelevant, as it does not address the question of whether or not microwave oven heating changes the molecular structure of food. Option C is also irrelevant, as it does not address the question of whether or not microwaved food is safe to eat.\\
            \textcolor{purple}{Here are some additional information about microwaves and their effects on food:}\\
            \textcolor{purple}{Microwaves are a form of non-ionizing radiation, which means that they do not have enough energy to damage DNA.}\\ 
            \textcolor{purple}{Microwaves heat food by causing water molecules to vibrate. This vibration produces heat, which cooks the food.}\\
            \textcolor{purple}{Microwaves do not change the chemical structure of food. This means that microwaved food is safe to eat.}\\ 
            \textcolor{purple}{Some studies have shown that microwave heating can cause nutrient loss. However, these studies have been small and the results have been mixed. More research is needed to determine the extent of nutrient loss from microwaved food.}\\
            Overall, there is no evidence that microwave oven heating is harmful to human health. Microwaved food is safe to eat.\\
            Answer and Explain: \textcolor{red}{\faCheckCircle} \textcolor{cyan}{\faTimesCircle} \textcolor{teal}{\faCheckCircle} \textcolor{purple}{\faCheckCircle}$\;\;\;\;\;\;\;\;$Explain Error Type: \textcolor{green}{\usym{1F5F5}} \textcolor{blue}{\usym{1F5F9}} \textcolor{brown}{\usym{1F5F5}} \textcolor{magenta}{\usym{1F5F9}}  \textcolor{violet}{\usym{1F5F5}}
            }\\
		\bottomrule
	\end{longtable}
}
\twocolumn

\onecolumn
{\small
\begin{longtable}{m{0.5cm}|l}
\caption{A reasoning case in LogiQA 2.0. }\label{tab:long}\\
		\toprule
            \rotatebox{90}{\makecell[c]{Question, answer\\and explanation}} & \makecell[{{p{16.3cm}}}]{Context: In a track and field qualifier, Zhang Qiang surpassed Li Jin, and Song Zhi's results were better than Wang Ping, but not as good as Ma Zheng.\\
            Question: Which of the following options can be inferred? A. Ma Zheng's grades are better than Zhang Qiang. B. Li Jin's results surpass Wang Ping. C. Zhang Qiang's performance is better than Song Zhi. D. Among the five people, Wang Ping ranked third at most.\\
            Answer: \textcolor{red}{D} 
            }\\
            \hline
            \rotatebox{90}{\makecell[c]{ChatGPT\\of 0-shot}}& \makecell[{{p{16cm}}}]{
            If Zhang Qiang surpassed Li Jin and Song Zhi's results were better than Wang Ping, but not as good as Ma Zheng, it can be inferred that \textcolor{red}{Zhang Qiang's performance is better than Song Zhi}. While the relative positions of Zhang Qiang, Li Jin, Wang Ping, and Ma Zheng are known, there is no information to suggest that Ma Zheng's grades are better than Zhang Qiang (option A) or that Li Jin's results surpass Wang Ping (option B). Option D introduces a ranking that is not directly supported by the information in the passage.\\
            Answer and Explain: \textcolor{red}{\faTimesCircle} \textcolor{cyan}{\faTimesCircle} \textcolor{teal}{\faTimesCircle} \textcolor{purple}{\faTimesCircle}$\;\;\;\;\;\;\;\;$Explain Error Type: \textcolor{green}{\usym{1F5F9}} \textcolor{blue}{\usym{1F5F5}} \textcolor{brown}{\usym{1F5F5}} \textcolor{magenta}{\usym{1F5F9}}  \textcolor{violet}{\usym{1F5F5}}
            }\\
            \hline
            \rotatebox{90}{\makecell[c]{ChatGPT\\of 1-shot}}& \makecell[{{p{16cm}}}]{
            The correct answer is \textcolor{red}{D}. From the given information, we know that Zhang Qiang's result was better than Li Jin's, and Song Zhi's result was better than Wang Ping's but not as good as Ma Zheng's. Therefore, we can infer that Ma Zheng's result was better than all the other athletes, including Zhang Qiang, and Li Jin's result was worse than both Zhang Qiang's and Wang Ping's. However, we cannot determine the exact ranking of Wang Ping among the five athletes with the given information, so option D is the correct choice as it cannot be inferred from the given context.\\
            Answer and Explain: \textcolor{red}{\faCheckCircle} \textcolor{cyan}{\faTimesCircle} \textcolor{teal}{\faTimesCircle} \textcolor{purple}{\faTimesCircle}$\;\;\;\;\;\;\;\;$Explain Error Type: \textcolor{green}{\usym{1F5F9}} \textcolor{blue}{\usym{1F5F5}} \textcolor{brown}{\usym{1F5F5}} \textcolor{magenta}{\usym{1F5F9}}  \textcolor{violet}{\usym{1F5F5}}
            }\\
            \hline
            \rotatebox{90}{\makecell[c]{ChatGPT\\of 3-shot}}& \makecell[{{p{16cm}}}]{
            The correct answer is \textcolor{red}{C}. The passage states that Zhang Qiang surpassed Li Jin and that Song Zhi's results were better than Wang Ping but not as good as Ma Zheng. Therefore, it can be inferred that Zhang Qiang's performance is better than Song Zhi's since he surpassed Li Jin. Choice A cannot be inferred because Ma Zheng's performance isn't compared to Zhang Qiang's. Choice B also cannot be inferred because there's no comparison between Li Jin's and Wang Ping's performance. Choice D cannot be inferred because there's no information to suggest that Wang Ping ranked third at most.\\
            Answer and Explain: \textcolor{red}{\faTimesCircle} \textcolor{cyan}{\faTimesCircle} \textcolor{teal}{\faTimesCircle} \textcolor{purple}{\faTimesCircle}$\;\;\;\;\;\;\;\;$Explain Error Type: \textcolor{green}{\usym{1F5F9}} \textcolor{blue}{\usym{1F5F5}} \textcolor{brown}{\usym{1F5F5}} \textcolor{magenta}{\usym{1F5F9}}  \textcolor{violet}{\usym{1F5F5}}
            }\\
            \hline
            \rotatebox{90}{\makecell[c]{Davinci-003\\of 0-shot}}& \makecell[{{p{16cm}}}]{
            The correct answer is \textcolor{red}{C}. Zhang Qiang\'s performance is better than Song Zhi.  This can be inferred from the given context, as the context stated that "Zhang Qiang surpassed Li Jin", which means Zhang Qiang is better than Li Jin, and "Song Zhi\'s results were better than Wang Ping", which means Song Zhi is better than Wang Ping, therefore Zhang Qiang is better than Song Zhi.  Therefore, option C is the correct choice.\\
            Answer and Explain: \textcolor{red}{\faTimesCircle} \textcolor{cyan}{\faTimesCircle} \textcolor{teal}{\faTimesCircle} \textcolor{purple}{\faTimesCircle}$\;\;\;\;\;\;\;\;$Explain Error Type: \textcolor{green}{\usym{1F5F9}} \textcolor{blue}{\usym{1F5F5}} \textcolor{brown}{\usym{1F5F5}} \textcolor{magenta}{\usym{1F5F9}}  \textcolor{violet}{\usym{1F5F5}}
            }\\
            \hline
            \rotatebox{90}{\makecell[c]{Bard\\of 0-shot}}& \makecell[{{p{16cm}}}]{
            The correct answer is \textcolor{red}{D}. Among the five people, Wang Ping ranked third at most.\\
            The passage states that Zhang Qiang surpassed Li Jin, and Song Zhi's results were better than Wang Ping, but not as good as Ma Zheng. This information can be organized into the following table:\\
            | Athlete | Results |\\
            | Zhang Qiang | Better than Li Jin |\\
            | Song Zhi | Better than Wang Ping |\\
            | Ma Zheng | Better than Song Zhi |\\
            From this table, we can see that Wang Ping's results are not as good as Ma Zheng's, so Wang Ping ranked third at most.\\
            The other answer choices are not supported by the passage. Answer choice A is not supported because the passage does not say that Ma Zheng's grades are better than Zhang Qiang's. Answer choice B is not supported because the passage does not say that Li Jin's results surpass Wang Ping's. Answer choice C is not supported because the passage does not say that Zhang Qiang's performance is better than Song Zhi's.\\
            Answer and Explain: \textcolor{red}{\faCheckCircle} \textcolor{cyan}{\faCheckCircle} \textcolor{teal}{\faCheckCircle} \textcolor{purple}{\faTimesCircle}
            }\\
		\bottomrule
	\end{longtable}
}
\twocolumn

\onecolumn
{\small
\begin{longtable}{m{0.5cm}|l}
\caption{A reasoning case in LogiQA2NLI. }\label{tab:long}\\
		\toprule
            \rotatebox{90}{\makecell[c]{Question, answer\\and explanation}} & \makecell[{{p{16.3cm}}}]{Context: The premis is: Researcher: People who participate in opinion surveys often give answers they believe the opinion surveyor expects to hear, and it is for this reason that some opinion surveys do not reflect the actual views of those being surveyed. However, in well-constructed surveys, the questions are worded so as to provide respondents with no indication of which answers the surveyor might expect. So if a survey is well constructed, survey respondents' desire to meet surveyors' expectations has no effect on the survey's results. The conclusion is: the answers of opinion-survey respondents can be influenced by beliefs about the surveyor's expectations even if those beliefs are unfounded.\\
            Question: Determine whether the premise and conclusion entailed or NOT entailed.\\
            Answer: \textcolor{red}{entailed}
            }\\
            \hline
            \rotatebox{90}{\makecell[c]{ChatGPT\\of 0-shot}}& \makecell[{{p{16cm}}}]{
            The conclusion \textcolor{red}{is not logically entailed} by the given premise. The premise discusses the potential for opinion survey respondents to give answers they believe the surveyor expects to hear, but notes that well-constructed surveys prevent this by wording questions in a way that does not indicate expected answers. It then concludes that if a survey is well constructed, respondents' desire to meet surveyors' expectations has no effect on the survey's results. Therefore, it does not support the conclusion that the answers of opinion-survey respondents can be influenced by beliefs about the surveyor's expectations even if those beliefs are unfounded.\\
            Answer and Explain: \textcolor{red}{\faTimesCircle} \textcolor{cyan}{\faTimesCircle} \textcolor{teal}{\faTimesCircle} \textcolor{purple}{\faTimesCircle}$\;\;\;\;\;\;\;\;$Explain Error Type: \textcolor{green}{\usym{1F5F9}} \textcolor{blue}{\usym{1F5F5}} \textcolor{brown}{\usym{1F5F5}} \textcolor{magenta}{\usym{1F5F9}}  \textcolor{violet}{\usym{1F5F5}}
            }\\
            \hline
            \rotatebox{90}{\makecell[c]{ChatGPT\\of 1-shot}}& \makecell[{{p{16cm}}}]{
            The conclusion is \textcolor{red}{not entailed} by the given premise. While the premise discusses how well-constructed surveys are worded so as to provide respondents with no indication of which answers the surveyor might expect, and how this can mitigate respondents' desires to meet surveyors' expectations, it does not provide information to support or refute the claim that the answers of opinion-survey respondents can be influenced by beliefs about the surveyor's expectations even if those beliefs are unfounded. Therefore, the conclusion cannot be reliably inferred from the given premise.\\
            Answer and Explain: \textcolor{red}{\faTimesCircle} \textcolor{cyan}{\faTimesCircle} \textcolor{teal}{\faTimesCircle} \textcolor{purple}{\faTimesCircle}$\;\;\;\;\;\;\;\;$Explain Error Type: \textcolor{green}{\usym{1F5F9}} \textcolor{blue}{\usym{1F5F5}} \textcolor{brown}{\usym{1F5F5}} \textcolor{magenta}{\usym{1F5F9}}  \textcolor{violet}{\usym{1F5F5}}
            }\\
            \hline
            \rotatebox{90}{\makecell[c]{ChatGPT\\of 3-shot}}& \makecell[{{p{16cm}}}]{
            \textcolor{red}{Not entailed}. The conclusion about the answers of opinion-survey respondents being influenced by beliefs about the surveyor's expectations cannot be logically derived from the given premise.\\
            Answer and Explain: \textcolor{red}{\faTimesCircle} \textcolor{cyan}{\faTimesCircle} \textcolor{teal}{\faTimesCircle} \textcolor{purple}{\faTimesCircle}$\;\;\;\;\;\;\;\;$Explain Error Type: \textcolor{green}{\usym{1F5F9}} \textcolor{blue}{\usym{1F5F5}} \textcolor{brown}{\usym{1F5F5}} \textcolor{magenta}{\usym{1F5F9}}  \textcolor{violet}{\usym{1F5F5}}
            }\\
            \hline
            \rotatebox{90}{\makecell[c]{Davinci-003\\of 0-shot}}& \makecell[{{p{16cm}}}]{
            \textcolor{red}{NOT entailed}. The premise is implying that because the survey is well constructed with no indication of what answers the surveyor might expect, the survey respondents' answers will not be impacted by any beliefs they have of the surveyor's expectations.  The conclusion is suggesting that even if those beliefs are unfounded, the responses will still be influenced by them.  This goes against the premise as the premise has stated that well-constructed surveys will give no indication of the surveyor's expectations, and therefore, their beliefs of the surveyor's expectations will have no influence.\\
            Answer and Explain: \textcolor{red}{\faTimesCircle} \textcolor{cyan}{\faTimesCircle} \textcolor{teal}{\faTimesCircle} \textcolor{purple}{\faTimesCircle}$\;\;\;\;\;\;\;\;$Explain Error Type: \textcolor{green}{\usym{1F5F9}} \textcolor{blue}{\usym{1F5F5}} \textcolor{brown}{\usym{1F5F5}} \textcolor{magenta}{\usym{1F5F9}}  \textcolor{violet}{\usym{1F5F5}}
            }\\
            \hline
            \rotatebox{90}{\makecell[c]{Bard\\of 0-shot}}& \makecell[{{p{16cm}}}]{
            The conclusion is \textcolor{red}{entailed} by the premise.  The premise states that people who participate in opinion surveys often give answers they believe the opinion surveyor expects to hear, even if those beliefs are unfounded.  It also states that well-constructed surveys are worded so as to provide respondents with no indication of which answers the surveyor might expect. \\
            This means that even if a survey is well-constructed, respondents may still be influenced by beliefs about the surveyor's expectations. This is because people are often motivated to please others, and they may believe that giving the answer they think the surveyor expects will make them look good.\\
            For example, a respondent may believe that a surveyor expects them to support a certain candidate in an election.  Even if the survey question is not leading, the respondent may still give the answer they think the surveyor expects, even if they do not actually support that candidate.\\
            This is why it is important for surveyors to be aware of the potential for respondents to be influenced by beliefs about their expectations.  Surveyors can try to minimize this effect by using clear and unbiased language, and by explaining to respondents that there are no right or wrong answers.\\
            Answer and Explain: \textcolor{red}{\faCheckCircle} \textcolor{cyan}{\faTimesCircle} \textcolor{teal}{\faTimesCircle} \textcolor{purple}{\faTimesCircle}$\;\;\;\;\;\;\;\;$Explain Error Type: \textcolor{green}{\usym{1F5F9}} \textcolor{blue}{\usym{1F5F5}} \textcolor{brown}{\usym{1F5F5}} \textcolor{magenta}{\usym{1F5F9}}  \textcolor{violet}{\usym{1F5F5}}
            }\\
		\bottomrule
	\end{longtable}
}
\twocolumn

\end{document}